\documentclass{article}

\usepackage[preprint]{neurips_2026}
\usepackage[utf8]{inputenc} % allow utf-8 input
\usepackage[T1]{fontenc}    % use 8-bit T1 fonts
\usepackage[colorlinks=true, citecolor=blue, linkcolor=blue, urlcolor=blue]{hyperref}
\usepackage{url}            % simple URL typesetting
\usepackage{booktabs}       % professional-quality tables
\usepackage{makecell}
\usepackage{amsfonts}       % blackboard math symbols
\usepackage{nicefrac}       % compact symbols for 1/2, etc.
\usepackage{microtype}      % microtypography
\usepackage{xcolor}         % colors

\usepackage{graphicx}

\usepackage{subcaption}

\usepackage{multirow}
\usepackage{paralist}
\usepackage{amsmath}
\usepackage{amssymb}
\usepackage{mathtools}
\usepackage{amsthm}
\usepackage[capitalize,noabbrev]{cleveref}
\usepackage{algorithm}
\usepackage{algorithmic}

\theoremstyle{plain}
\newtheorem{theorem}{Theorem}[section]

\newtheorem{lemma}[theorem]{Lemma}
\newtheorem{corollary}[theorem]{Corollary}
\theoremstyle{definition}

\newtheorem{assumption}[theorem]{Assumption}
\theoremstyle{remark}

\def \ze {\mathbf{0}}
\def \R {\mathbb{R}}
\def \K {\mathcal{K}}

\def \d {\mathbf{d}}
\def \E {\mathbb{E}}
\def \x {\mathbf{x}}

\def \y {\mathbf{y}}
\def \g {\mathbf{g}}
\def \s {\mathbf{s}}

\def \z {\mathbf{z}}

\def \uu {\mathbf{u}}

\DeclareMathOperator*{\argmin}{argmin}

\DeclareMathOperator*{\CG}{CG}

\title{Revisiting Decentralized Online Convex Optimization\\ with Compressed Communication}

\author{%
  Hao Zhou$^{1}$ \quad 
  Xiaoyu Wang$^{1}$ \quad
  Chang Yao$^{1,2}$ \quad 
  Mingli Song$^{1,2}$ \quad 
  Yuanyu Wan$^{1,2}$\thanks{Corresponding author.} \\
  $^1$ School of Software Technology, Zhejiang University, Ningbo, China \\
  $^2$ State Key Laboratory of Blockchain and Data Security, Zhejiang University, Hangzhou, China \\
  \texttt{\{juhao, xiaoyuw, brooksong, changy, wanyy\}@zju.edu.cn}
}

\begin{document}

\maketitle

\begin{abstract}
Decentralized online convex optimization (D-OCO) is a popular framework for distributed applications with streaming data. To tackle the communication bottleneck, previous studies have investigated D-OCO with compressed communication and proposed several algorithms that are variants of online gradient descent (OGD). However, for D-OCO with exact communication, the best existing algorithms are variants of follow-the-regularized-leader (FTRL). In this paper, for the first time, we propose two FTRL-type algorithms for D-OCO with compressed communication. Compared with OGD-type algorithms, our algorithms are more elegant in both algorithmic design and theoretical analysis. The key insight is that the dual update mechanism of FTRL allows us to make a simple application of the technique for average consensus with communication compression. More specifically, our first algorithm considers the full-information setting, and can match the existing regret bounds. Our second algorithm is designed for the bandit setting, and can significantly improve both the regret bounds and communication costs of existing algorithms.   
\end{abstract}

\section{Introduction}
\label{Introduction}
This paper investigates the decentralized online convex optimization (D-OCO) problem with a particular focus on algorithms that only use compressed communication. In general, D-OCO is formulated as a repeated game between $n$ learners in a network and an adversary. At each round $t$, each learner $ i \in [n]$ selects a decision $ \x_i(t) $ from a convex set $ \mathcal{K} \subseteq \mathbb{R}^d $, and then receives a convex loss function $ f_{t,i}(\cdot):\K\mapsto\mathbb{R}$ chosen by the adversary. Let $f_t(\cdot) =\sum_{j=1}^n f_{t,j}(\cdot)$ denote the global function of round $t$. For total $T$ rounds, the goal of each learner $i\in[n]$ is to minimize the regret in terms of the global function, i.e., $R_{T,i} = \sum_{t=1}^T f_t(\x_i(t)) - \min_{\x \in \K} \sum_{t=1}^T f_t(\x)$. To this end, these learners are allowed to communicate with their immediate neighbors once per round. 

In the case with exact communication, D-OCO has been extensively studied over the past decade \citep{DAOL_TKDE,hosseini2013online,wenpeng17,Wan-ICML-2020,Wan-22-JMLR,Wang-AAAI-2023,wan2024nearly,wan2025optimal}. It is well known that most algorithms for the non-distributed setting, including online gradient descent (OGD) \citep{Zinkevich2003} and follow-the-regularized-leader (FTRL) \citep{Shai07,NIPS2009_3882}, can be extended into D-OCO by combining with the gossip-based average consensus technique \citep{Xiao-Gossip04,Acc_Gossip}. Notably, by extending FTRL, \citet{wan2024nearly,wan2025optimal} have achieved nearly optimal regret bounds for D-OCO with exact communication. In contrast, only a few previous studies have considered D-OCO with compressed communication. 

To be precise, two concurrent studies \citep{tu2022distributed,cao2023decentralized} propose to combine a decentralized variant of OGD (D-OGD) \citep{DAOL_TKDE} with Choco-Gossip \citep{ICML-Gossip}---a compressed variant of the standard gossip technique \citep{Xiao-Gossip04}. Unfortunately, the proposed algorithm is unsatisfactory, because its regret has a large dependence on the spectral gap $\rho\in(0,1]$ of the gossip matrix, the compression ratio $\omega\in(0,1]$, and $n$. To address this issue, \citet{topdogd} propose an improved algorithm with $\tilde{O}(\omega^{-1/2}\rho^{-1}n\sqrt{T})$ and $\tilde{O}(\omega^{-1}\rho^{-2}n \log T)$ regret bounds for convex and strongly convex functions, respectively.\footnote{The $\tilde{O}(\cdot)$ notation hides constant factors as well as  polylogarithmic factors in $n$, but not in $T$.} This algorithm  is still a combination of D-OGD and Choco-Gossip, though it introduces a two-level blocking update mechanism and a repeated compressor to further reduce the consensus error.

\begin{table}[t]
\centering
\caption{Comparison of our bandit algorithm and the previous best one. Abbreviations: convex $\rightarrow$ cvx, strongly convex $\rightarrow$ scvx.}
\label{tab1}
% 1. 使用标准的缩小字号命令（\small 对应 9pt，是 NeurIPS 表格的推荐字号）
% \small 
% 2. 稍微缩小默认的列间距（默认是 6pt，缩小到 3pt 或 4pt）
\setlength{\tabcolsep}{4pt} 
\begin{tabular}{lccc} 
    \toprule
    % 3. 把过长的表头拆成两行（需要在导言区 \usepackage{makecell}）
    Assumption & Reference & Regret Bound & \makecell{Communication Rounds} \\
    \midrule
    
    \multirow{2}{*}[-0.5ex]{$f_{t,i}(\cdot)$: cvx} 
    & \citet{topdogd} & $\tilde{O}(\omega^{-1/4}\rho^{-1/2}nT^{3/4})$ & $O(T)$ \\
    \cmidrule(lr){2-4} 
     & Corollary \ref{cor:thm2-c} & ${O}(nT^{3/4})$ & $\tilde{O}(\omega^{-1}\rho^{-2}\sqrt{T})$ \\
    \midrule
    
    \multirow{2}{*}[-0.5ex]{$f_{t,i}(\cdot)$: scvx} 
    & \citet{topdogd} & $\tilde{O}(\omega^{-1/3}\rho^{-2/3}n T^{2/3}(\log T)^{1/3})$ & $O(T)$ \\
    \cmidrule(lr){2-4} 
     & Corollary \ref{cor:thm2-sc} & ${O}(nT^{2/3}(\log T)^{1/3})$ & $\tilde{O}(\omega^{-1}\rho^{-2}T^{1/3}(\log T)^{2/3})$ \\
    \bottomrule
\end{tabular}
\end{table}

Thus, it remains unclear whether the FTRL-type algorithm for D-OCO can also be extended into the case with compressed communication, and whether such an extension provides any benefits relative to existing algorithms. In this paper, we provide an affirmative answer to both questions. Specifically, we first propose a compressed variant of the FTRL-type algorithm, and show that it enjoys the same regret bounds as the algorithm of \citet{topdogd}. Despite no improvement in the bounds, both our algorithm and analysis are much simpler than those of \citet{topdogd}. The key insight is that for the FTRL-type algorithm, the decision is determined by a local approximation of the cumulative average gradient, i.e., the so-called dual variable. Although the mapping from the dual variable to the decision is equivalent to a projection onto the set $\K$, the consensus error of these dual variables is not affected by the projection. For this reason, it can be controlled by a simple application of Choco-Gossip.

Furthermore, we extend our first algorithm into a more challenging bandit setting, where only the loss value $f_{t,i}(\x_i(t))$ is revealed to each learner $i$. Note that previous studies \citep{tu2022distributed,topdogd} have also considered this setting. The bandit algorithm of \citet{topdogd} achieves the best existing regret bounds, i.e., $\tilde{O}(\omega^{-1/4}\rho^{-1/2}nT^{3/4})$ and $\tilde{O}(\omega^{-1/3}\rho^{-2/3}n T^{2/3}(\log T)^{1/3})$ for convex and strongly convex functions, respectively. We demonstrate that our bandit algorithm is not only simpler than that of \citet{topdogd}, but also improves their two bounds to ${O}(nT^{3/4})$ and ${O}(nT^{2/3}(\log T)^{1/3})$, respectively. Moreover, different from the algorithm of \citet{topdogd} that requires $T$ communication rounds in total, our bandit algorithm only requires $\tilde{O}(\omega^{-1}\rho^{-2}\sqrt{T})$ and $\tilde{O}(\omega^{-1}\rho^{-2}T^{1/3}(\log T)^{2/3})$ communication rounds to achieve these two bounds. A detailed comparison of our bandit algorithm and that of \citet{topdogd} is presented in Table \ref{tab1}.

\section{Related work}
\label{sec:related_work}
% In this section, we briefly review related work on the standard D-OCO, and compressed communication.
Now, we briefly review related work on the standard D-OCO, and compressed communication.

\subsection{The standard D-OCO}
The study of D-OCO dates back to the pioneering work of \citet{DAOL_TKDE}, in which D-OGD is proposed to achieve $O(n^{5/4}\rho^{-1/2}\sqrt{T})$ and $O(n^{3/2}\rho^{-1}\log T)$ regret bounds for convex and strongly convex functions, respectively. The key idea of D-OGD is to first apply a standard gossip step \citep{Xiao-Gossip04} over these local decisions, and then perform a projected gradient descent step according to only the local function. Note that the essence of the gossip step is to compute a weighted average of these local variables, which needs to be implemented via a communication round, and the weights are specified by the gossip matrix. Concurrently, \citet{hosseini2013online} propose a decentralized variant of FTRL (D-FTRL) \citep{Shai07,NIPS2009_3882}, and achieve the same $O(n^{5/4} \rho^{-1/2} \sqrt{T})$ regret bound for convex functions. Different from D-OGD, this algorithm uses the standard gossip step to update the dual variables of local decisions. 

After that, there has been a growing research interest in developing projection-free algorithms for D-OCO \citep{wenpeng17,Wan-ICML-2020,Wan-22-JMLR,Wang-AAAI-2023}. These algorithms are motivated by the fact that both D-OGD and D-FTRL require a projection to ensure the feasibility of each decision, which can be time-consuming in applications with a complex set $\K$. Although the design of projection-free algorithms is beyond the scope of this paper, \citet{Wan-22-JMLR} have also proposed a projection-free and generalized variant of D-FTRL for the bandit setting. For convex and strongly convex functions, it achieves $O(n^{5/4} \rho^{-1/2} T^{3/4})$ and $O(n^{3/2} \rho^{-1} T^{2/3}(\log T)^{1/3})$ regret bounds, respectively. More interestingly, it only requires $O(\sqrt{T})$ and $O(T^{1/3}(\log T)^{2/3})$ communication rounds, when handling these two types of functions. Such improvements mainly owe to a blocking update mechanism, i.e., dividing total $T$ rounds into several blocks and only updating the decision once per block.

Recently, \citet{wan2024nearly, wan2025optimal} propose a novel full-information algorithm called accelerated decentralized follow-the-generalized-leader (AD-FTGL), which enjoys $\tilde{O}(n\rho^{-1/4}\sqrt{T})$ and $\tilde{O}(n\rho^{-1/2}\log T)$ regret bounds for convex and strongly convex functions, respectively. This algorithm is also a variant of D-FTRL, and the critical change is to
exploit an accelerated gossip strategy \citep{Acc_Gossip}. They have provided nearly matching lower bounds to demonstrate its optimality. Moreover, \citet{wan2025optimal} have also proposed a projection-free variant of AD-FTGL with ${O}(nT^{3/4})$ and ${O}(nT^{2/3}(\log T)^{1/3})$ regret bounds for convex and strongly convex functions, while communicating $\tilde{O}(\rho^{-1/2}\sqrt{T})$ and $\tilde{O}(\rho^{-1/2}T^{1/3}(\log T)^{2/3})$ rounds, respectively. Although this variant is developed for the full-information setting, it actually can be simply extended into the bandit setting while keeping the same regret bounds and communication costs. Compared with the projection-free bandit algorithm in  \citet{Wan-22-JMLR}, the improvements on these regret bounds benefit from the accelerated gossip strategy and some sacrifices of communication costs.

\subsection{Compressed communication}
Compressed communication is a popular strategy for reducing the amount of data that has to be sent during each iteration of distributed optimization. For the centralized distributed offline optimization, it is well-known that there are many techniques for compressed communication (also known as compressors), which can be divided into quantization \citep{alistarh2017qsgd, wen2017terngrad, bernstein2018signsgd, seide_1-bit}, sparsification \citep{alistarh2018convergence, stich2018sparsified, wangni2018gradient, aji-heafield-2017-sparse, lin2018deep}, and their hybrid combination \citep{basu2019qsparse, wang2018atomo}. For the decentralized offline optimization, \citet{tang2018communication} is the first work to apply the idea of compressed communication, but only focus on unbiased compressors with high accuracy constraints. To tackle this issue, \citet{ICML-Gossip} propose a novel and unified method called Choco-Gossip for average consensus with different compressors in the decentralized setting, as well as a corresponding Choco-SGD algorithm for the decentralized offline optimization.

The study of D-OCO with compressed communication actually can be dated back to \citet{li2021decentralized}. However, they only propose a heuristic approach without any theoretical guarantee. By combining D-OGD \citep{DAOL_TKDE} with Choco-Gossip \citep{ICML-Gossip}, \citet{tu2022distributed} and \citet{cao2023decentralized} concurrently propose the first algorithm to establish rigorous regret bounds for D-OCO with compressed communication. Specifically, it can respectively achieve $O(C_1\sqrt{T})$ and $O(C_1\log T)$ regret bounds for convex and strongly convex functions, where $C_1=\max\{\omega^{-2}\rho^{-4}n^{3/2},\omega^{-4}\rho^{-8}n\}$. Additionally, \citet{tu2022distributed} have also extended this algorithm into the bandit setting, and the previous two bounds respectively degrade to $O(C_2T^{3/4})$ and $O(C_3T^{2/3}(\log T)^{1/3})$, where $C_2=\max\{\omega^{-1}\rho^{-2}n^{5/4},\omega^{-2}\rho^{-4}n\}$ and $C_3=\max\{\omega^{-2/3}\rho^{-4/3}n^{7/6},\omega^{-8/3}\rho^{-4}n\}$. 

Very recently, \citet{topdogd} further improve these two algorithms of \citet{tu2022distributed}. In the full-information setting, their algorithm enjoys $\tilde{O}(\omega^{-1/2}\rho^{-1} n \sqrt{T})$ and $\tilde{O}(\omega^{-1}\rho^{-2} n \log T)$ regret bounds for convex and strongly convex functions, respectively. In the bandit setting, their improved algorithm can achieve $\tilde{O}(\omega^{-1/4}\rho^{-1/2} n T^{3/4})$ and $\tilde{O}(\omega^{-1/3}\rho^{-2/3} n T^{2/3}(\log T)^{1/3})$ regret bounds for convex and strongly convex functions, respectively. Despite such improvements, their algorithms introduce a complicated scheme that combines a two-level blocking update mechanism with Choco-Gossip and a repeated compressor.

% Such improvements of \citet{topdogd} are owed to the combination of a two-level blocking update mechanism with Choco-Gossip and a repeated compressor.

% Note that the complicated scheme of \citet{topdogd} is mainly designed to eliminate the effect of projection on their consensus error. In this paper, we demonstrate that FTRL-type algorithms can achieve the same and even improved results in a more elegant way. 

\section{Main results}
\label{sec:main_results}
In this section, we first introduce necessary assumptions, and then present our two FTRL-type algorithms for D-OCO with compressed communication. The proofs of all theoretical results can be found in the appendix.
% introduce an FTRL-based framework with compressed communication. We first present the standard assumptions and basic tools used throughout this section. Then, we describe the proposed algorithms for the full-information and bandit settings, followed by the corresponding regret bounds.
\subsection{Assumptions}
\label{subsec:assumption}
Similar to previous studies on D-OCO with compressed communication \citep{tu2022distributed,topdogd}, we first introduce the following assumptions.
\begin{assumption}
\label{doubly stochastic}
Let $A\in \mathbb{R}^{n\times n}$ denote the gossip matrix. Define the network by an undirected graph $\mathcal{G}=([n],E)$, where $E\subseteq[n]\times[n]$ denotes the edge set. The matrix $A$ is supported on the graph $\mathcal{G}$ and doubly stochastic, i.e.,
\begin{compactitem}
\item $A_{ij}>0$ only if $(i,j)\in E$ or $i=j$;
\item $\sum_{j=1}^nA_{ij}=1,\forall i\in [n]$ and $\sum_{i=1}^nA_{ij}=1,\forall j\in [n]$.
\end{compactitem}
Moreover, $A$ is symmetric and positive semidefinite. 
\end{assumption}
\begin{assumption}
\label{assum_compress}
Let $Q(\cdot)$ : $\mathbb{R}^d \rightarrow \mathbb{R}^d$ denote a compressor, whose output can be encoded with fewer bits than the original input. 
For some $\omega \in (0, 1]$, it satisfies
\begin{equation}
\mathbb{E}_Q [\|Q(\x) - \x\|_2^2] \leq (1 - \omega)\|\x\|_2^2, \quad \forall \x \in \mathbb{R}^d, \notag
\end{equation}
where $\mathbb{E}_Q$ denotes the expectation over the possible randomness of $Q$.
\end{assumption}
\begin{assumption}
\label{r<K<R}
The convex set \( \mathcal{K} \) is full dimensional, and there exist two constants \( r, R > 0 \) such that \( r\mathcal{B}^d \subseteq \mathcal{K} \subseteq R\mathcal{B}^d \), where \( \mathcal{B}^d \) denotes the unit Euclidean ball centered at the origin in \( \mathbb{R}^d \).
\end{assumption}
\begin{assumption}
\label{G-lipschitz}
At each round $t\in[T]$, the loss function $f_{t,i}(\x)$ of each learner $i\in [n]$ is $G$-Lipschitz over $\K$, i.e., it holds that $|f_{t,i}(\x)-f_{t,i}(\mathbf{y})|\leq G\|\x-\mathbf{y}\|_2,~\forall\x,\mathbf{y}\in \K$.
\end{assumption}
\begin{assumption}
\label{scvx-assum}
At each round $t\in[T]$, the loss function $f_{t,i}(\x)$ of each learner $i\in [n]$ is $\alpha$-strongly convex over $\K$, i.e., it holds that $f_{t,i}(\mathbf{y})\geq f_{t,i}(\x)+\langle\nabla f_{t,i}(\x),\mathbf{y}-\x\rangle+\frac{\alpha}{2}\|\mathbf{y}-\x\|_2^2,~\forall\x,\mathbf{y}\in \K.$
\end{assumption}
\begin{assumption}
\label{bound:|f|<M}
At each round $t$, the loss function $f_{t,i}(\x)$ of each learner $i\in [n]$ is bounded over $\mathcal{K}$, i.e., $|f_{t,i}(\mathbf{x})| \leq M$ for any $\mathbf{x} \in \mathcal{K}$. Moreover, all loss functions are chosen beforehand, i.e., the adversary is oblivious.
\end{assumption}

\textbf{Remark.} First, from Assumption \ref{doubly stochastic}, the eigenvalues of $A$ can be denoted as $1=\lambda_1(A)>\lambda_2(A)\geq \cdots\geq \lambda_n(A)$, and the spectral gap is $\rho=1-\lambda_2(A)$. Second, as discussed in \citet{ICML-Gossip}, Assumption \ref{assum_compress} captures a broad class of both biased and unbiased compressors (see Appendix~\ref{app:unified_compressors} for detailed discussions). Third, Assumption \ref{scvx-assum} with $\alpha=0$ reduces to the case with general convex functions, which are also considered in this paper. Finally, Assumptions \ref{r<K<R} and \ref{bound:|f|<M} are mainly required by the bandit setting. One can verify that in the full-information setting, the former can be simplified to $\|\x\|_2\leq R,\forall\x\in\K$, and the latter can be removed.

\subsection{Algorithm for full-information setting}
\label{our CD-FTGL}
Before introducing our algorithms, we first briefly recall the best existing algorithm for D-OCO with exact communication, i.e., AD-FTGL \citep{wan2024nearly, wan2025optimal}. Specifically, it adopts a blocking update mechanism with block size $L$, i.e., dividing the total $T$ rounds into $T/L$ blocks,\footnote{Without loss of generality, $T/L$ is assumed to be an integer.} and only maintaining a decision $\x_i(z)$ for all rounds in each block $z$. The initial decision is simply set as $\x_i(1)=\ze$. To update the decision, it exploits an accelerated gossip strategy \citep{Acc_Gossip} to maintain a dual variable $\z_i(z)\approx \sum_{\tau=1}^{z-1}(1/n)\sum_{i=1}^{n} \mathbf{d}_i(\tau)$, where $\mathbf{d}_i(\tau) = \sum_{t \in \mathcal{T}_\tau} (\nabla f_{t,i}(\mathbf{x}_i(\tau)) - \alpha \mathbf{x}_i(\tau))$ and $\mathcal{T}_\tau = \{(\tau-1)L+1, \ldots, \tau L\}$. Based on $\z_i(z)$, it computes
\begin{equation}
\label{block_DFTGL_update}
\mathbf{x}_i(z+1) = \argmin_{\mathbf{x} \in \mathcal{K}} \langle \mathbf{z}_i(z), \mathbf{x} \rangle + \frac{(z-1)L\alpha + 2h}{2} \|\mathbf{x}\|_2^2,
\end{equation}
where $h$ is a parameter. Let $\bar{\d}(z)=(1/n)\sum_{i=1}^{n} \mathbf{\d}_i(z)$ and $\bar{\z}(z)=\sum_{\tau=1}^{z-1}\bar{\d}(\tau)$. According to their analysis, the regret of AD-FTGL can be decomposed into two parts: one about the cumulative consensus error of $\| \mathbf{z}_i(z)-\bar{\z}(z)\|_2$ and the other about the regret of a virtual update based on $\bar{\z}(z)$.

Therefore, the critical challenge for extending AD-FTGL into D-OCO with compressed communication becomes how to maintain $\mathbf{z}_i(z)$ with a small consensus error in this case. To this end, we introduce Choco-Gossip \citep{ICML-Gossip}, whose detailed procedure is outlined in Algorithm \ref{alg:choco}, and the following lemma.
\begin{lemma}
\label{lem_ezl_ez0_maintext}
Let $\beta:=\|I-A \|_2$, $\bar{\mathbf{z}}^k = (1/n) \sum_{i=1}^n \mathbf{z}_i^k$ and $e^k = \mathbb{E}[\sum_{i=1}^n \| \mathbf{z}_i^k - \bar{\mathbf{z}}^k \|_2^2]  + \mathbb{E}[ \sum_{i=1}^n \| \mathbf{z}_i^k - \hat{\mathbf{z}}_i^k \|_2^2]$, where $I$ is the identity matrix. Under Assumptions \ref{doubly stochastic} and \ref{assum_compress}, Algorithm \ref{alg:choco} with $\s_i^0=\sum_{j=1}^nA_{ij}\hat{\z}_j^0,\forall i\in[n]$ and 
\begin{equation}
\label{gamma-value}
\gamma = \frac{\rho \omega}{2(1-\omega)(\rho+2)\beta^2 + 16\rho - 8\rho\omega + \rho^2} ,
\end{equation}
 ensures  $e^k \leq (1-\gamma \rho/2 )^k e^0$ for any $k\in[K]$.
\end{lemma}
Given $n$ local vectors $\{\z_i^0\}_{i\in[n]}$ in a decentralized network, the role of Choco-Gossip is to approximately compute their average $\bar{\z}=(1/n)\sum_{i=1}^n\z_i^0$ via only compressed communication, which is known as the average consensus problem. Since it is easy to verify that $\bar{\mathbf{z}}^k=\bar{\z}$, Lemma \ref{lem_ezl_ez0_maintext} is sufficient to provide an upper bound on the consensus error. 

\textbf{Remark.} Note that Algorithm \ref{alg:choco} follows a memory-efficient implementation of Choco-Gossip, which can be found in the appendix of \citet{ICML-Gossip}. Moreover, our Lemma \ref{lem_ezl_ez0_maintext} actually is a generalized and refined version of the original guarantee on Choco-Gossip. To be precise, Theorem 2 of \citet{ICML-Gossip} only shows $e^k\leq (1-\omega\rho^2/82)^ke^0$ with the initialization $\hat{\z}_i^0=\ze$ (and $\s_i^0=\ze$ for the memory-efficient implementation), which cannot recover the classical $O((1-\rho)^k)$ consensus error bound of the standard gossip \citep{Xiao-Gossip04} even if $\omega=1$. In contrast, our lemma holds for more flexible initialization with a worst-case bound of $O((1-\omega\rho^2)^k)$ and an improved bound of $O((1-\omega\rho)^k)$ if $\omega=1$. Such initialization makes Choco-Gossip plug-and-play in our algorithms, and the improvement may be also of independent interest.

\begin{algorithm}[t]
\caption{Choco-Gossip (CG)}
\label{alg:choco}
\begin{algorithmic}[1]
\STATE \textbf{Input:} Original vectors $\{\z_i^0\}_{i\in[n]}$, initial compressed vectors $\{\hat{\z}_i^0\}_{i\in[n]}$, initial intermediate vectors $\{\mathbf{s}_i^0\}_{i\in[n]}$, step size $\gamma$, number of iterations $K$
% \STATE \textbf{Initialization:} set 
\FOR{$k = 0 \dots K-1$ \textbf{in parallel} for each $i \in [n]$}
    \STATE Compress the difference as $\mathbf{q}_i^{k} = Q(\z_i^{k} - \hat{\z}_i^{k})$
    \STATE Send $\mathbf{q}_i^{k}$ to neighbors and receive $\mathbf{q}_j^{k}$
    \STATE  $\hat{\z}_i^{k+1} = \hat{\z}_i^{k} + \mathbf{q}_i^{k}$, $\mathbf{s}_i^{k+1} = \mathbf{s}_i^{k} + \sum_{j=1}^n A_{ij} \mathbf{q}_j^{k}$
    \STATE $\z_i^{k+1} = \z_i^{k} + \gamma ( \mathbf{s}_i^{k+1} - \hat{\z}_i^{k+1} )$
\ENDFOR
\end{algorithmic}
\end{algorithm}

Now, we are ready to combine AD-FTGL with only compressed communication. For brevity, let $\CG(\cdot,\cdot,\cdot,\cdot,\cdot)$ denote Choco-Gossip in Algorithm \ref{alg:choco}. A natural idea is to applying Choco-Gossip over $\{\d_i(z-1)\}_{i\in[n]}$ during each block $z$ to generate 
\begin{equation}
\{\d^K_i(z-1)\}_{i\in[n]}
=\CG(\{\d_i(z-1)\}_{i\in[n]},\{\ze\}_{i\in[n]},\{\ze\}_{i\in[n]},\gamma,K), \notag
\end{equation}
and then setting $\z_i(z)=\sum_{\tau=1}^{z-1}\d^K_i(z-1)$. However, in this way, we need to set $K$ proportional to $O(\log T)$ for making the cumulative consensus error sufficiently small. To address this issue, we exploit Choco-Gossip in a more careful way. Specifically, we simply set $\z_{i}(1)\!=\hat{\z}_i(1)\!=\mathbf{s}_i(z)=\ze$ for any $i\in [n]$. For each block $z\geq 2$, we first set $\z_i^0(z)=\z_i(z-1)+\d_i(z-1)$, $\hat{\z}_i^0(z)=\hat{\z}_i(z-1)$, and $\s_i^0(z)=\s_i(z-1)$. Then, we apply Choco-Gossip as 
\begin{equation}
\label{eq-use-choco}
\begin{split}
&\left(\{\z^K_i(z)\}_{i\in[n]},\{\hat{\z}^K_i(z)\}_{i\in[n]},\{\s^K_i(z)\}_{i\in[n]}\right)\\
=&\CG(\{\z_i^0(z)\}_{i\in[n]},\{\hat{\z}^0_i(z)\}_{i\in[n]},\{\mathbf{s}^0_i(z)\}_{i\in[n]},\gamma,K),
\end{split}
\end{equation}
and set $\z_i(z)=\z_i^K(z)$, $\hat{\z}_i(z)=\hat{\z}^K_i(z)$, and $\s_i(z)=\s_i^K(z)$. In this way, once $\d_i(z-1)$ is incorporated into $\z_i(z)$, it will also be affected by the subsequent Choco-Gossip. Thus, $K$ can be independent of $T$.

\begin{algorithm}[t]
\caption{CD-FTGL}
\label{CD-FTGL}
\begin{algorithmic}[1]
\STATE \textbf{Input:} strongly convex factor $\alpha$, regularization parameter $h$, block size $L$, consensus step size $\gamma$
\STATE $\mathbf{x}_i(1)\!=\z_{i}(1)\!=\hat{\z}_i(1)\!=\mathbf{s}_i(1)\!=\ze, \forall i\in [n]$
\FOR{$z=1,\dots,T/L$ \textbf{in parallel} for each $i \in [n]$}
\FOR{$t=(z-1)L+1,\dots,zL$} 
\STATE Play $\x_{i}(z)$ and observe $\nabla f_{t,i}(\x_{i}(z))$
\ENDFOR
\IF{$z \geq 2$}
\STATE $\z_i^0(z)=\z_i(z-1)+\d_i(z-1)$, $\hat{\z}_i^0(z)=\hat{\z}_i(z-1)$, and $\mathbf{s}_i^0(z)=\mathbf{s}_i(z-1)$
\STATE Set $K=L$ and compute $\z^K_i(z),\hat{\z}^K_i(z),\s^K_i(z)$ by invoking Algorithm \ref{alg:choco} as in \eqref{eq-use-choco}
\STATE $\z_{i}(z)=\z_i^{K}(z)$, $\hat{\z}_i(z)=\hat{\z}_i^{K}(
z)$, $\s_i(z)=\s_i^{K}(z)$
\ENDIF
\STATE $\d_i(z)=\sum_{t\in\mathcal{T}_z}(\nabla f_{t,i}(\x_i(z))-\alpha\x_i(z))$ %, where $\mathcal{T}_z=\{(z-1)L+1,\dots,zL\}$ 
\STATE Compute 
$\x_i(z+1)$ via (\ref{block_DFTGL_update})
\ENDFOR
\end{algorithmic}
\end{algorithm}

From the above discussions, we propose our first algorithm called compressed decentralized follow-the-generalized-leader (CD-FTGL). The detailed procedure is summarized in Algorithm \ref{CD-FTGL}. 

\textbf{Remark.}
First, we want to emphasize that in step 9 of Algorithm \ref{CD-FTGL}, the invocation of Algorithm \ref{alg:choco} can be implemented in parallel to the \emph{for} loop from steps 4 to 6, which implies that both the computation and communication costs can be allocated to each round in block $z$. Second, we notice that \citet{topdogd} also exploit Choco-Gossip in a similar way, but to control the consensus error among decisions. Moreover, according to the OGD-type update of decision, they need to use a more complicated blocking mechanism and a repeated compressor to eliminate the effect of a projection operation on the consensus error. In contrast, in our algorithm, the consensus error can be controlled independent of the projection.

By combining the existing analysis of AD-FTGL \citep{wan2024nearly, wan2025optimal} and Lemma \ref{lem_ezl_ez0_maintext}, we first  establish a general guarantee on the regret of Algorithm \ref{CD-FTGL}.
\begin{theorem}
\label{thm:thm1}
    Under Assumptions \ref{doubly stochastic}, \ref{assum_compress}, \ref{r<K<R}, \ref{G-lipschitz}, and \ref{scvx-assum}, for any $i \in [n]$, Algorithm \ref{CD-FTGL} with $L=\left\lceil 2 \ln(n+2)/(\gamma \rho) \right\rceil$, where $\gamma$ is defined in \eqref{gamma-value}, ensures
    \begin{equation}
    \label{thm1}
    \E[R_{T,i}] \leq n h R^2 + \sum_{z=2}^{T/L} \frac{6 n L^2 G(G + \alpha R)}{(z-2)L\alpha + 2h} + \sum_{z=1}^{T/L} \frac{12 n L^2 G (G + 2\alpha R)}{zL\alpha + 2h}. 
    \end{equation}
\end{theorem}
By further tuning $h$, we obtain the following regret
bounds for convex and strongly convex functions.
\begin{corollary}
\label{cor:thm1-c}
Under Assumptions \ref{doubly stochastic}, \ref{assum_compress}, \ref{r<K<R}, \ref{G-lipschitz}, and \ref{scvx-assum} with \( \alpha = 0 \), for any \( i \in [n] \), Algorithm \ref{CD-FTGL} with \( h = 3\sqrt{LT}G/R \) and $L=\left\lceil 2 \ln(n+2)/(\gamma \rho) \right\rceil$ ensures 
\[
\E[R_{T,i}] \leq 6nGR\sqrt{LT}.
% = O \left( n \sqrt{\ln n} \cdot \rho^{-1} \omega^{-1/2} \cdot \sqrt{T} \right). \notag
\]
% Moreover, in the uncompressed setting (\( \omega = 1 \)), the consensus step size simplifies to \( \gamma = \frac{1}{8+\rho} \), which implies \( L = O(\rho^{-1} \ln n) \). Thus, the regret bound improves to
% \begin{equation}
% R_{T,i} = O \left( n \sqrt{\ln n} \cdot \rho^{-1/2} \cdot \sqrt{T} \right). \notag
% \end{equation}
\end{corollary}

\begin{corollary}
\label{cor:thm1-sc}
Under Assumptions \ref{doubly stochastic}, \ref{assum_compress}, \ref{r<K<R}, \ref{G-lipschitz}, and \ref{scvx-assum} with \( \alpha > 0 \), for any \( i \in [n] \), Algorithm \ref{CD-FTGL} with \( h = \alpha L \) and $L=\left\lceil 2 \ln(n+2)/(\gamma \rho) \right\rceil$ ensures
\[
\E[R_{T,i}] \leq n \alpha R^2 L + \frac{3nG(5G + 9\alpha R)L}{\alpha} \ln\left( T/L\right).
% &= O \left( n \ln n \cdot \rho^{-2} \omega^{-1} \cdot \ln T \right). \notag
\]
% Moreover, in the uncompressed setting, the regret bound improves to
% \begin{equation}
% R_{T,i} = O \left( n \ln n \cdot \rho^{-1} \cdot \ln T \right). \notag
% \end{equation}
\end{corollary}

\textbf{Remark.}~
From these two corollaries and the definition of $\gamma$ in \eqref{gamma-value}, our CD-FTGL achieves $\tilde{O}(\omega^{-1/2}\rho^{-1}n\sqrt{T})$ and $\tilde{O}(\omega^{-1}\rho^{-2} n \log T)$ regret bounds for convex and strongly convex functions, respectively. Notably, our results match the current best upper bounds achieved by \citet{topdogd}. Moreover, \citet{topdogd} also established $\Omega(\omega^{-1/2}\rho^{-1/4}n\sqrt{T})$ and $\Omega(\omega^{-1}\rho^{-1/2}n\log T)$ lower bounds for these two settings. This indicates that our dependencies on $\omega, n$ and $T$ are already nearly optimal. Furthermore, based on the discussions about the consensus error in Lemma \ref{lem_ezl_ez0_maintext}, if $\omega=1$, our two regret bounds improve to $\tilde{O}(\rho^{-1/2}n\sqrt{T})$ and $\tilde{O}(\rho^{-1} n\log T)$, which recovers the best existing results for D-OCO algorithms with standard gossip \citep{Xiao-Gossip04}.\footnote{Actually, the original analysis of both D-OGD and D-FTRL does not provide these results, which are first achieved by \citet{wan2025optimal} with a refined analysis on the consensus error.} In contrast, even if $\omega=1$, the regret bounds of \citet{topdogd} will only become $\tilde{O}(\rho^{-1}n\sqrt{T})$ and $\tilde{O}(\rho^{-2} n \log T)$.

\subsection{Algorithm for bandit setting}
\label{our CD-FTBL}
Compared with the full-information setting, the critical challenge of the bandit setting is that only the value of each local function at the played decision can be observed. Fortunately, in the literature, there exists a well-known technique called the one-point estimator \citep{OBO05} that can generate an approximate gradient by using a single loss value. It can be formalized as the following lemma, where \( \mathcal{B}^d \) and \(\mathcal{S}^d\) denotes the unit ball and sphere centered at the origin in \( \mathbb{R}^d \), respectively.
\begin{lemma}
\label{lem:epsilon-smooth}
    (Lemma 1 in \citet{OBO05}) For any function \(f(\mathbf{x}): \mathcal{K} \mapsto \mathbb{R}\) and \(\epsilon > 0\), we define its \(\epsilon\)-smoothed version as \(\hat{f}_\epsilon(\mathbf{x}) = \mathbb{E}_{\mathbf{u} \sim \mathcal{B}^d}[f(\mathbf{x} + \epsilon \mathbf{u})]\). Then, it holds that
\begin{equation}
   \nabla \hat{f}_\epsilon(\mathbf{x}) = \mathbb{E}_{\mathbf{u} \sim \mathcal{S}^d}\left[ \frac{d}{\epsilon} f(\mathbf{x} + \epsilon \mathbf{u}) \mathbf{u} \right].  \notag
\end{equation}
\end{lemma}
By combining our CD-FTGL with this technique, our bandit algorithm is outlined in Algorithm \ref{CD-FTBL}, and named as compressed decentralized follow-the-bandit-leader (CD-FTBL).

Specifically, compared with CD-FTGL, there exist some critical differences. First, to apply the one-point estimator, in step 5 of Algorithm \ref{CD-FTBL}, we play $\y_i(t)=\x_{i}(z)+\epsilon \uu_i(t)$, which consists of $\x_{i}(z)$ computed by using historical information and a random point $\epsilon \uu_i(t)$. Due to $\uu_i(t) \sim \mathcal{S}^d$ and Assumption \ref{r<K<R}, $\x_i(z)\in\mathcal{K}_\epsilon = ( 1 - \epsilon/r ) \mathcal{K} = \{ ( 1 - \epsilon/r) \mathbf{x} \mid \mathbf{x} \in \mathcal{K}\}$ is a sufficient condition for $\y_i(t)\in \K$. Therefore, the second difference is that $\x_{i}(z+1)$ in step 14 of Algorithm \ref{CD-FTBL} should be computed over the shrunk set $\mathcal{K}_\epsilon$, i.e., 
\begin{equation}
\label{block_DFTGL_update_bandit}
\mathbf{x}_i(z+1)
= \argmin_{\mathbf{x} \in \mathcal{K}_\epsilon}
\langle \mathbf{z}_i(z), \mathbf{x} \rangle
+ \frac{(z-1)L\alpha + 2h}{2} \|\mathbf{x}\|_2^2.
\end{equation}
Third, instead of querying $\nabla f_{t,i}(\y_i(t))$, in step 6 of Algorithm \ref{CD-FTBL}, we compute an approximate one as $\g_i(t)=\frac{d}{\epsilon}f_{t,i}(\y_i(t))\uu_i(t)$, which is also used to replace the gradient originally required by $\d_i(z)$ in step 13 of Algorithm \ref{CD-FTBL}. Finally, inspired by \citet{Wan-22-JMLR}, we notice that a much larger block size may be used to reduce the number of communication rounds. As a result, in step 10 of Algorithm \ref{CD-FTBL}, we no longer set $K=L$ when invoking Choco-Gossip.

\begin{algorithm}[t]
\caption{CD-FTBL}
\label{CD-FTBL}
\begin{algorithmic}[1]
\STATE \textbf{Input:} strongly convex factor $\alpha$, regularization parameter $h$, block size $L$, communication rounds $K$, consensus step size $\gamma$, exploration radius $\epsilon$
\STATE $\mathbf{x}_i(1)\!=\z_{i}(1)\!=\hat{\z}_i(1)\!=\mathbf{s}_i(1)\!=\ze, \forall i\in [n]$
\FOR{$z=1,\dots,T/L$ \textbf{in parallel} for each node $i$}
\FOR{$t=(z-1)L+1,\dots,zL$} 
\STATE Play $\y_i(t)=\x_{i}(z)+\epsilon \uu_i(t)$, where $\uu_i(t) \sim \mathcal{S}^d$
\STATE Compute $\g_i(t)=\frac{d}{\epsilon}f_{t,i}(\y_i(t))\uu_i(t)$
\ENDFOR
\IF{$z \geq 2$}
\STATE $\z_i^0(z)=\z_i(z-1)+\d_i(z-1)$, $\hat{\z}_i^0(z)=\hat{\z}_i(z-1)$, and $\mathbf{s}_i^0(z)=\mathbf{s}_i(z-1)$
\STATE Compute $\z^K_i(z),\hat{\z}^K_i(z),\s^K_i(z)$ by invoking Algorithm \ref{alg:choco} as in \eqref{eq-use-choco}
\STATE $\z_{i}(z)=\z_i^{K}(z)$, $\hat{\z}_i(z)=\hat{\z}_i^{K}(
z)$, $\s_i(z)=\s_i^{K}(z)$
\ENDIF
\STATE $\d_i(z)=\sum_{t\in\mathcal{T}_z}(\g_i(t)-\alpha\x_i(z))$
\STATE Compute 
$\x_i(z+1)$ via (\ref{block_DFTGL_update_bandit})
\ENDFOR
\end{algorithmic}
\end{algorithm}

Moreover, we have the following guarantee regarding the regret of Algorithm~\ref{CD-FTBL}.
\begin{theorem}
\label{thm:thm2}
Under Assumptions \ref{doubly stochastic}, \ref{assum_compress}, \ref{r<K<R}, \ref{G-lipschitz}, \ref{scvx-assum}, and \ref{bound:|f|<M}, for any $i \in [n]$, Algorithm \ref{CD-FTBL} with $L\geq K=\left\lceil 2 \ln(n+2)/(\gamma \rho) \right\rceil$ and $\epsilon<r$, ensures
\begin{equation}
\label{thm2}
\begin{aligned}
&\E[R_{T,i}]
\leq n h R^2 + 3\epsilon n G T + \frac{\epsilon n G R T}{r}+  \sum_{z=1}^{T/L}\frac{2 n B_1^2}{z L \alpha + 2 h}\\
&\quad+ \sum_{z=1}^{T/L} \frac{6nLG B_1}{(z-1) L \alpha + 2 h}+ \sum_{z=2}^{T/L} \frac{6nLG B_2}{(z-2)L\alpha + 2h} ,\\
\end{aligned}
\end{equation}
where $B_1=2 \sqrt{L}dM/\epsilon+2 L G+3\alpha LR$ and
$B_2= \sqrt{2L d^2M^2/\epsilon^2 +2L^2G^2  +2\alpha^2L^2R^2}$.
\end{theorem}
By further tuning $h$ and $\epsilon$, we obtain specific regret bounds for convex and strongly convex functions.
\begin{corollary}
\label{cor:thm2-c}
Under Assumptions \ref{doubly stochastic}, \ref{assum_compress}, \ref{r<K<R}, \ref{G-lipschitz}, \ref{scvx-assum} with $\alpha=0$, and \ref{bound:|f|<M}, for any $i \in [n]$, Algorithm \ref{CD-FTBL} with $K=\left\lceil 2 \ln(n+2)/(\gamma \rho) \right\rceil$, $L=\max\{K,\sqrt{T}\}$, $ h = \sqrt{dLT}M/R $, and $ \epsilon = c \sqrt{d} \, T^{-1/4} $, where $c$ is a constant such that $\epsilon < r$, ensures
\[
\mathbb{E}[R_{T,i}] = O(n T^{3/4}+n\sqrt{KT}).
\]
% Moreover, in the uncompressed setting, the regret bound improves to
% \begin{equation}
% \mathbb{E}[R_{T,i}] = O \left( n \sqrt{\ln n} \cdot \rho^{-1/2} \cdot \sqrt{d} \,T^{3/4} \right). \notag
% \end{equation}
% Furthermore, given a sufficiently large block size $ L = \sqrt{T} \geq K $, the regret decouples from $ \rho $ and $ \omega $:
% \begin{equation}
% \mathbb{E}[R_{T,i}] = O \left( n \sqrt{d}\, T^{3/4} \right). \notag
% \end{equation}
\end{corollary}

\begin{corollary}
\label{cor:thm2-sc}
Under Assumptions \ref{doubly stochastic}, \ref{assum_compress}, \ref{r<K<R}, \ref{G-lipschitz}, \ref{scvx-assum} with $\alpha>0$, and \ref{bound:|f|<M}, for any $i \in [n]$, Algorithm \ref{CD-FTBL} with $K=\left\lceil 2 \ln(n+2)/(\gamma \rho) \right\rceil$, $L=\max\{K,T^{2/3}(\ln T)^{-2/3}\}$, $ h = \alpha L $, and $ \epsilon = c d^{2/3} T^{-1/3} (\ln T)^{1/3} $, where $c$ is a constant such that $\epsilon <r$, ensures
\[
\mathbb{E}[R_{T,i}] = O\left( n T^{2/3} (\ln T)^{1/3} + n K\log T \right).
\]
% Moreover, when $\omega=1$, the regret bound improves to
% \begin{equation}
% \mathbb{E}[R_{T,i}] = O \left( n \ln n \cdot \rho^{-1} \cdot d^{2/3} T^{2/3} (\ln T)^{1/3} \right). \notag
% \end{equation}
\end{corollary}

\textbf{Remark.}~We first notice that in the above two regret bounds, only the non-dominant term depends on $\rho$ and $\omega$ via the definition of $K$. This implies that our CD-FTBL enjoys a nice ability to decouple the joint effect of the bandit feedback and the decentralized compressed communication. Moreover, as discussed in previous studies \citep{wan2025optimal,wang2025revisiting}, the total number of rounds $T$ is commonly much larger than other problem constants. By combining Corollaries \ref{cor:thm2-c} and \ref{cor:thm2-sc} with this assumption, the two regret bounds can be simplified to $ O(nT^{3/4})$ and $O( n T^{2/3} (\log T)^{1/3})$, which are much tighter than the best existing $\tilde{O}(\omega^{-1/4}\rho^{-1/2}nT^{3/4})$ and $\tilde{O}(\omega^{-1/3}\rho^{-2/3} n T^{2/3}(\log T)^{1/3})$ regret bounds for convex and strongly convex functions \citep{topdogd}. Finally, it is worth noting that the number of communication rounds of our CD-FTBL is $TK/L$. Under the same assumption, Corollaries \ref{cor:thm2-c} and \ref{cor:thm2-sc} only require $\tilde{O}(\omega^{-1}\rho^{-2}\sqrt{T})$ and $\tilde{O}(\omega^{-1}\rho^{-2}T^{1/3}(\log T)^{2/3})$ communication rounds, respectively. To the best of our knowledge, this is the first D-OCO algorithm that can reduce the number of communication rounds and the bits of each communication simultaneously.

\section{Experiments}
\label{sec:Experiments}
\begin{figure}[htbp]
    \centering
    
    % --------- ijcnn1: Loss vs Rounds ---------
    \begin{subfigure}[b]{0.42\textwidth}
        \includegraphics[width=\textwidth]{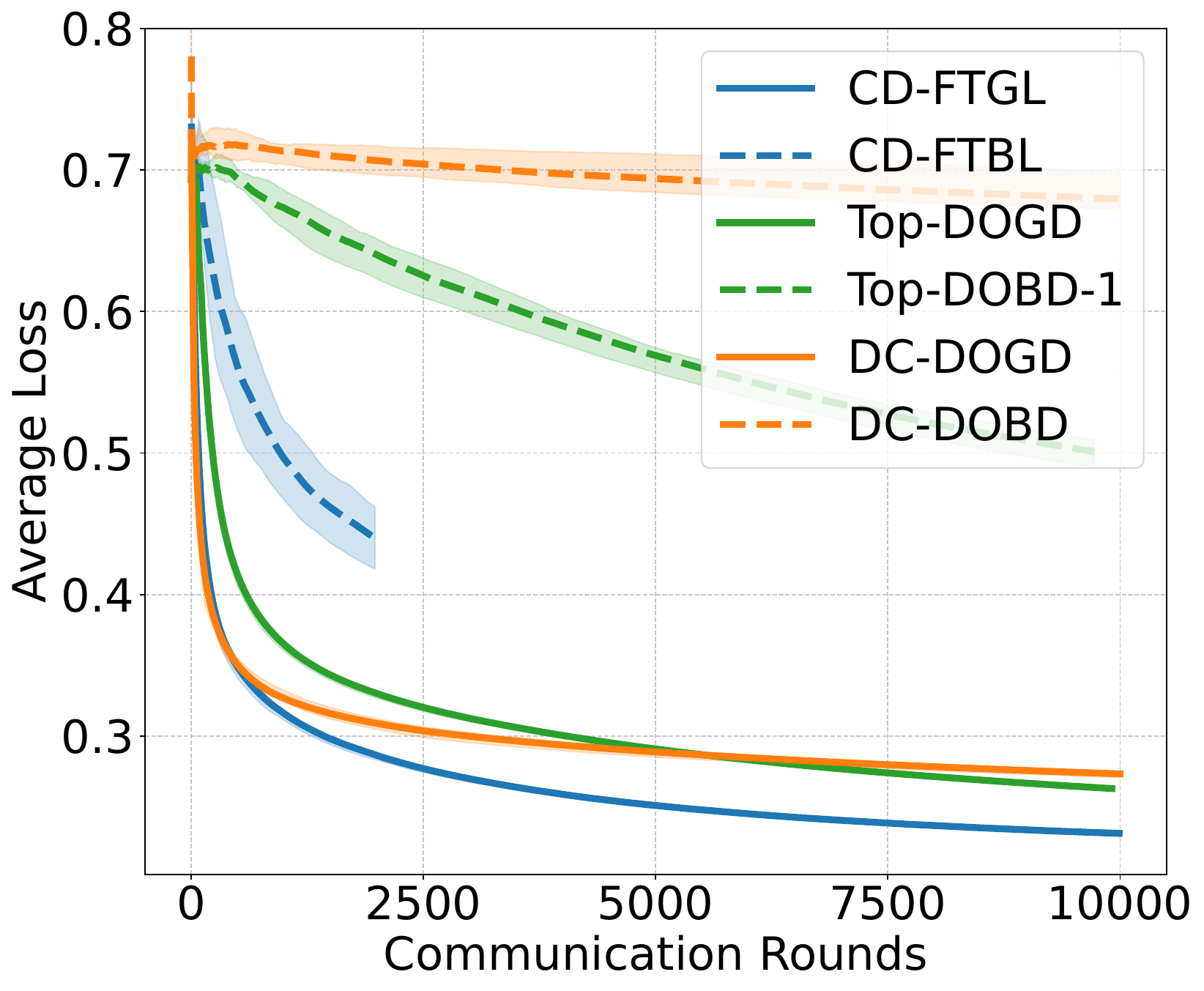}
        \caption{ijcnn1: Loss vs Rounds}
        \label{ijcnn1: rounds}
    \end{subfigure}
    \hspace{0.02\textwidth} % 缩小水平间距，让左右两张图靠得更紧
    % --------- ijcnn1: Loss vs Bits ---------
    \begin{subfigure}[b]{0.42\textwidth}
        \includegraphics[width=\textwidth]{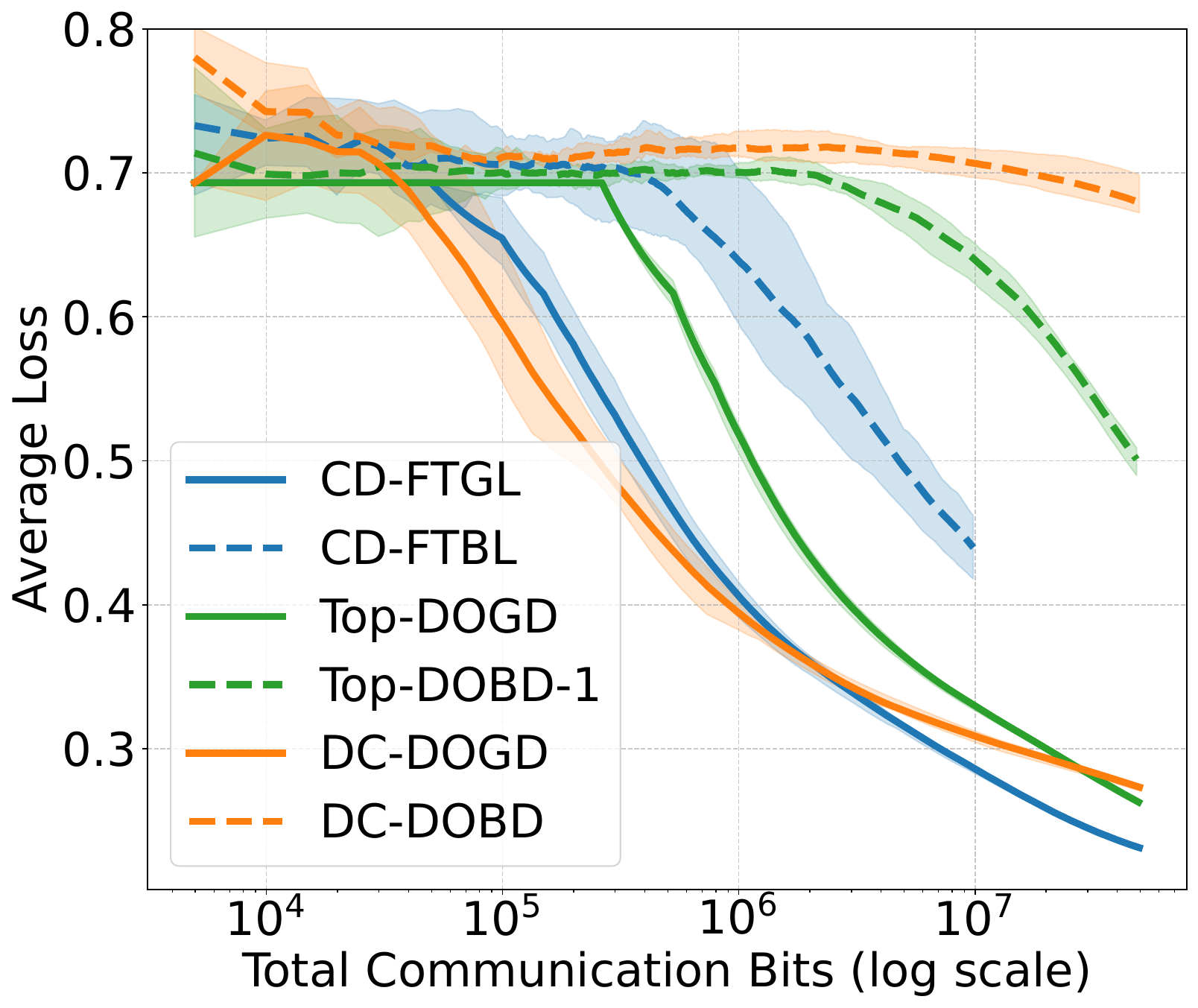}
        \caption{ijcnn1: Loss vs Bits}
        \label{ijcnn1: bits}
    \end{subfigure}
    
    \vspace{0em} % 缩小上下两排之间的垂直间距（若白边大可改为 0em 甚至 -0.5em）
    
    % --------- a9a: Loss vs Rounds ---------
    \begin{subfigure}[b]{0.42\textwidth}
        \includegraphics[width=\textwidth]{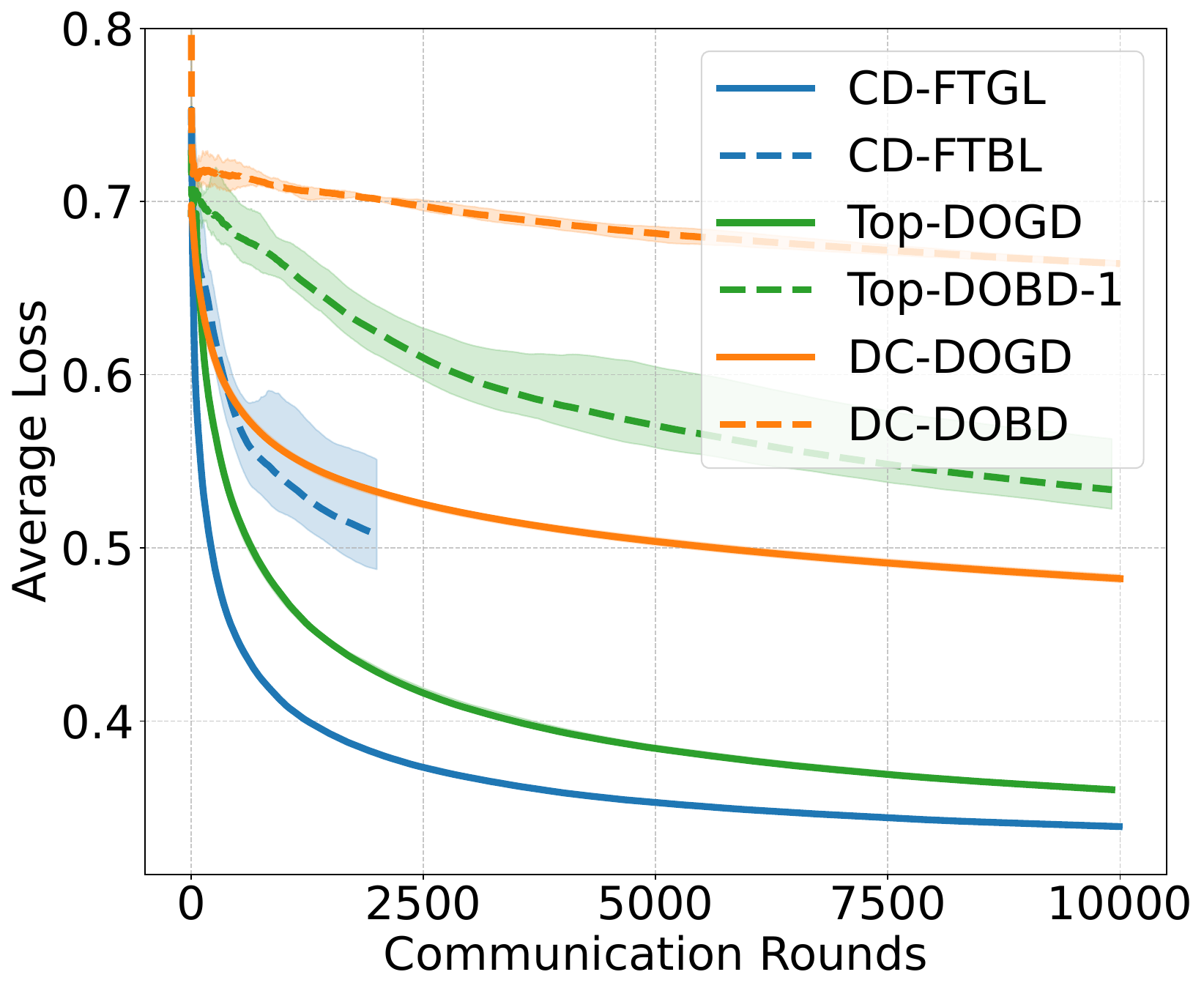}
        \caption{a9a: Loss vs Rounds}
        \label{a9a: rounds}
    \end{subfigure}
    \hspace{0.02\textwidth} % 保持与上方一致的水平间距
    % --------- a9a: Loss vs Bits ---------
    \begin{subfigure}[b]{0.42\textwidth}
        \includegraphics[width=\textwidth]{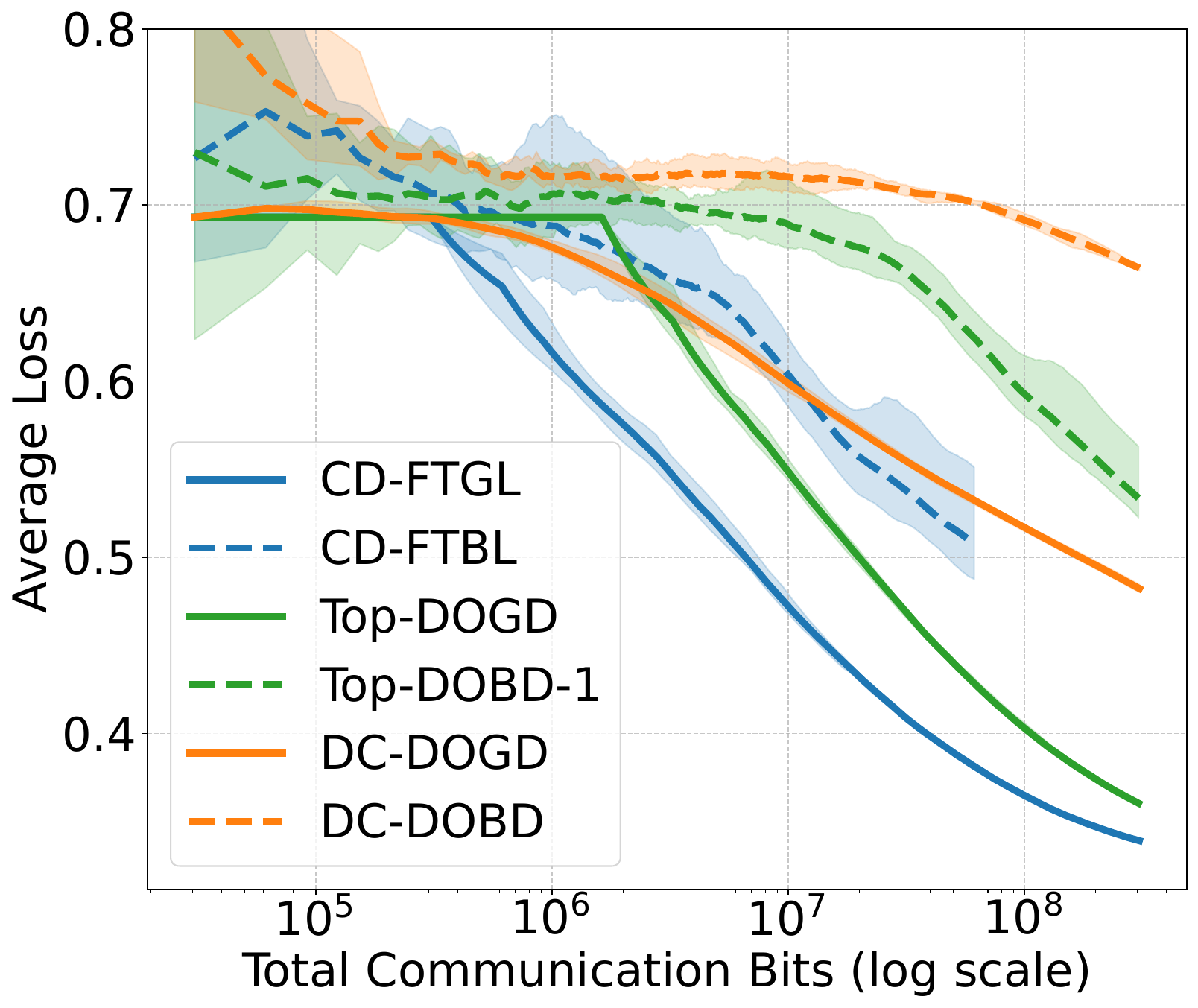}
        \caption{a9a: Loss vs Bits}
        \label{a9a: bits}
    \end{subfigure}
    
    \vspace{-0.5em} % 稍微拉近底部主标题与图片的距离
    \caption{Experimental results on a small random graph ($9$ nodes and 18 edges) with $\omega = 0.1$.}
    \label{fig:n9_omega_0.1_results}
\end{figure}

To verify the performance of our algorithms, we conduct experiments on the decentralized online logistic regression problem with two datasets: ijcnn1 and a9a from the LIBSVM repository~\citep{LIBSVM}. The loss function for learner $i$ at round $t$ is given by $f_{t, i}(\x) = \ln ( 1 + \exp( -b_{t, i} \langle \textbf{a}_{t, i}, \x \rangle ) )$, where $\textbf{a}_{t, i}\in \R^d$ is the feature vector and $b_{t, i} \in \{-1, 1\}$ is the class label. The decision set is defined as a bounded Euclidean ball $\mathcal{K} = \{ \x \in \mathbb{R}^d \mid \|\x\|_2 \leq \tau \}$, where $\tau = 10$. We adopt the average cumulative loss, defined as $AL(t,i) = \frac{1}{t n}\sum_{\tau=1}^t \sum_{j=1}^n f_{\tau, j}(\x_i(\tau))$, as the performance metric. Moreover, the original ijcnn1 and a9a datasets consist of 49990 and 32561 examples, respectively. To ensure $T=10000$, we first make a random permutation of these examples, and then allocate 10000 examples to each local learner cyclically.

The simulation involves $n = 9$ learners connected via a $\mathcal{G}(n, 2n)$ random graph, which is generated by a tool called NetworkX \citep{hagberg2008exploring}. Based on this graph topology, the gossip matrix $A$ satisfying Assumption~\ref{doubly stochastic} is constructed using the local-degree weights \citep{Xiao-Gossip04}, where $A_{ij} = 1/\max\{d_i, d_j\}$ for connected nodes $i$ and $j$, and $A_{ii} = 1 - \sum_{j \neq i} A_{ij}$. We choose Top-DOGD~\citep{topdogd} and DC-DOGD~\citep{tu2022distributed} as the full-information baselines, and Top-DOBD-1~\citep{topdogd} and DC-DOBD~\citep{tu2022distributed} as the bandit baselines. In all experiments, we use the Top-$k$ compressor to reduce communication overhead, and each algorithm is run 5 times, and the mean performance with shaded regions indicating the min-max range is reported. 

To ensure a fair comparison, we tune hyper-parameters via grid search. Specifically, the theoretical values of regularization parameter $h$ and learning rate $\eta$ are multiplied by the scaling coefficients, which are selected from $\{10^{-5}, 10^{-4}, \dots, 10^{5}\}$. The exploration radius $\epsilon$ is chosen from $\{0.1, 1, 10\}$, while the block size $L$ (and $L_1$ for Top-DOGD and Top-DOBD-1) is tuned over $\{10, 100, 1000\}$. The compression ratio is uniformly set to $\omega = 0.1$ across both datasets. Other parameters remain at their suggested theoretical values.

Figure~\ref{fig:n9_omega_0.1_results} plots the average loss against the number of communication rounds and transmitted bits for both datasets. In both the full-information and bandit settings, our algorithms achieve a lower average loss compared to their respective baselines. More significantly, our CD-FTBL requires only about 2,000 communication rounds, whereas the bandit baselines require at least 10,000 rounds. 
To further evaluate the actual communication cost, Figures~\ref{ijcnn1: bits} and \ref{a9a: bits} illustrate the average loss with respect to the total transmitted bits. In these plots, the loss curves for our algorithms visibly shift to the left compared to their baselines. This indicates that our algorithms require fewer communication bits to reach the same average loss, and thus verifies their communication efficiency again. Additional experimental results can be found in Appendix~\ref{app:experiments}.

\section{Conclusion and future work}
\label{sec:conclusion}
This paper revisits the problem of D-OCO with compressed communication. Unlike previous studies that focus on OGD-type algorithms, we propose two FTRL-type algorithms, namely CD-FTGL and CD-FTBL, which simplify the existing analysis and even achieve improved results. Specifically, our CD-FTGL is developed for the full-information setting, and can match the existing $\tilde{O}(\omega^{-1/2}\rho^{-1}n\sqrt{T})$ and $\tilde{O}(\omega^{-1}\rho^{-2}n\log T)$ regret bounds for convex and strongly convex functions, respectively. Our CD-FTBL can handle the bandit setting, and significantly improves the regret for convex and strongly convex functions to ${O}(nT^{3/4})$ and ${O}(nT^{2/3}(\log T)^{1/3})$, respectively. Interestingly, it only requires sublinear communication rounds to achieve these results.

Nonetheless, there are still some open problems. For example, as previously discussed, CD-FTGL with $\omega=1$ can recover the best existing results of full-information D-OCO algorithms with the standard gossip. However, there still exist gaps in terms of $\rho$ from the regret bounds of AD-FTGL \citep{wan2024nearly,wan2025optimal}, which is based on the accelerated gossip strategy. Thus, it is appealing to investigate whether this accelerated technique can also be combined with compressed communication. Moreover, we notice that in the full-information setting, the blocking update mechanism is not necessary for the application of the standard gossip \citep{wan2025optimal}. Thus, it is also interesting to study whether the same results achieved by our CD-FTGL can be obtained with $L=1$.

% \begin{ack}
% % [在此处填入致谢与基金支持。双盲评审阶段，ack 环境中的内容会自动被隐藏，无需手动注释]
% \end{ack}

% ===== 参考文献 =====
\bibliographystyle{plainnat} % 或使用符合你引文习惯的格式
\bibliography{nips26}

%%%%%%%%%%%%%%%%%%%%%%%%%%%%%%%%%%%%%%%%%%%%%%%%%%%%%%%%%%%%
\newpage
\appendix

% ===== 附录部分 =====
% ==========================================
% 原有图表保留区 (Network Size & Compression)
% ==========================================

\begin{figure}[htbp]
    \centering
    % --------- ijcnn1: Loss vs Rounds ---------
    \begin{subfigure}[b]{0.24\textwidth}
        \includegraphics[width=\textwidth]{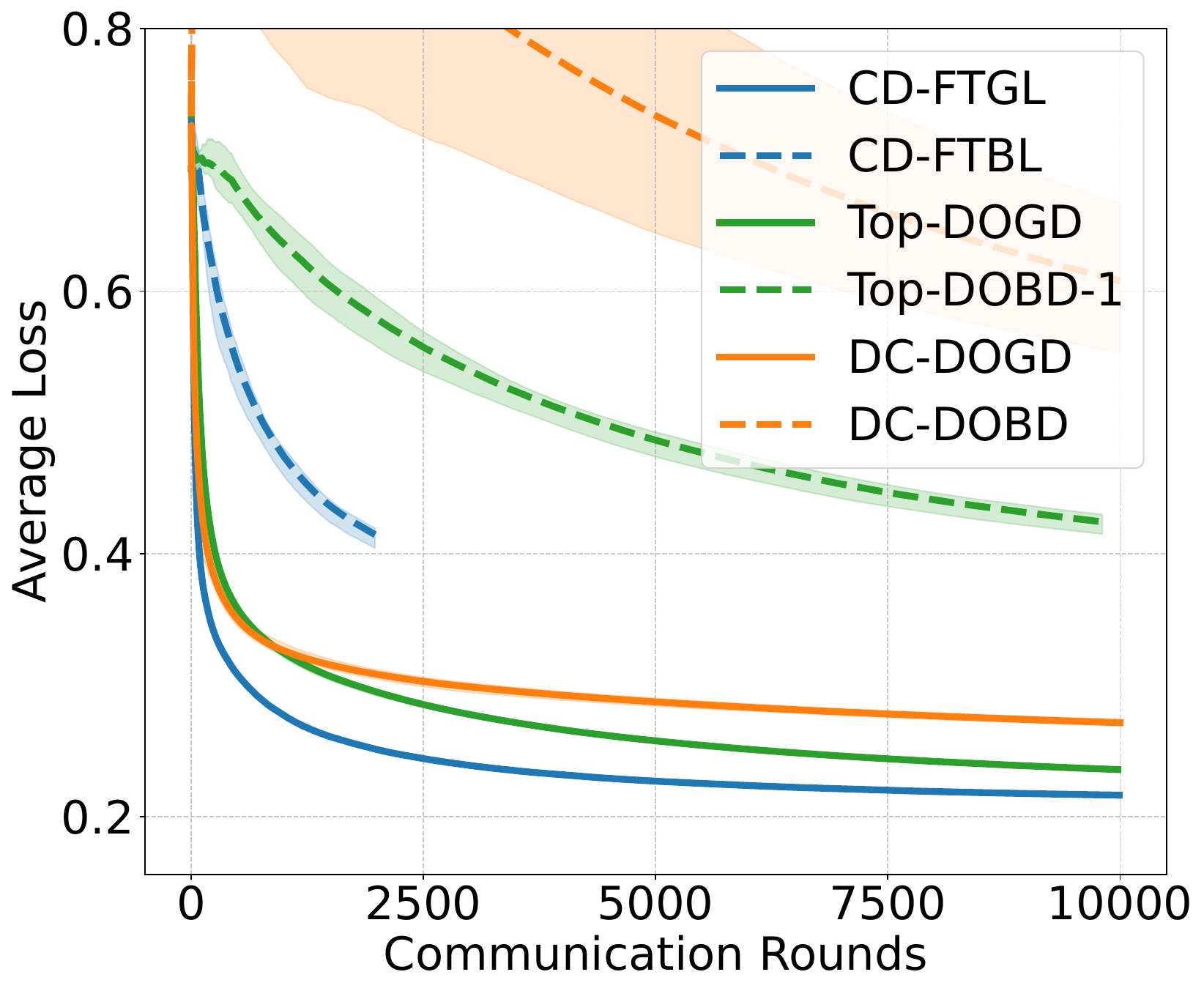}
        \caption{ijcnn1: Loss vs Rounds}
    \end{subfigure}
    \hfill
    % --------- ijcnn1: Loss vs Bits ---------
    \begin{subfigure}[b]{0.24\textwidth}
        \includegraphics[width=\textwidth]{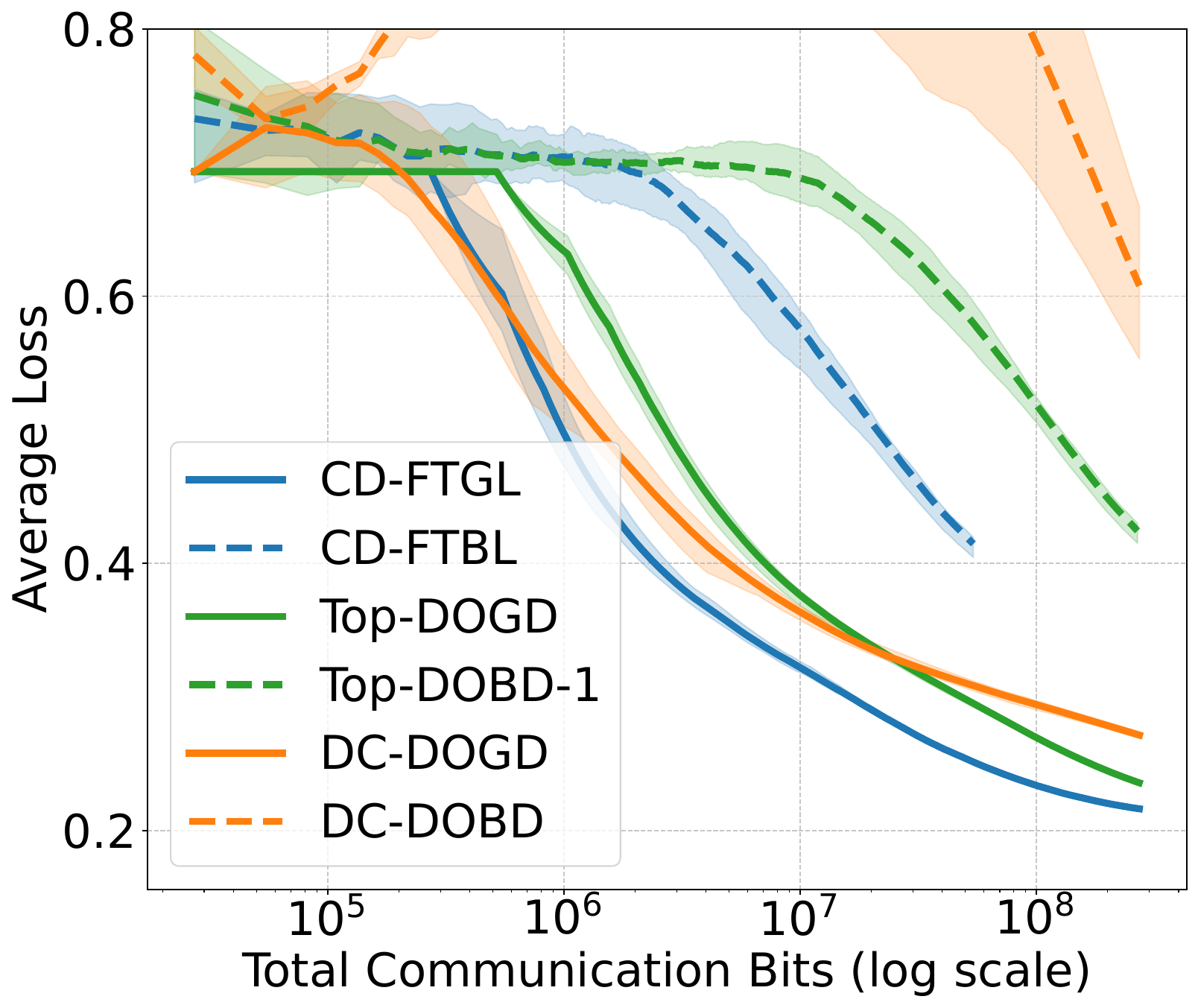}
        \caption{ijcnn1: Loss vs Bits}
    \end{subfigure}
    \hfill
    % --------- a9a: Loss vs Rounds ---------
    \begin{subfigure}[b]{0.24\textwidth}
        \includegraphics[width=\textwidth]{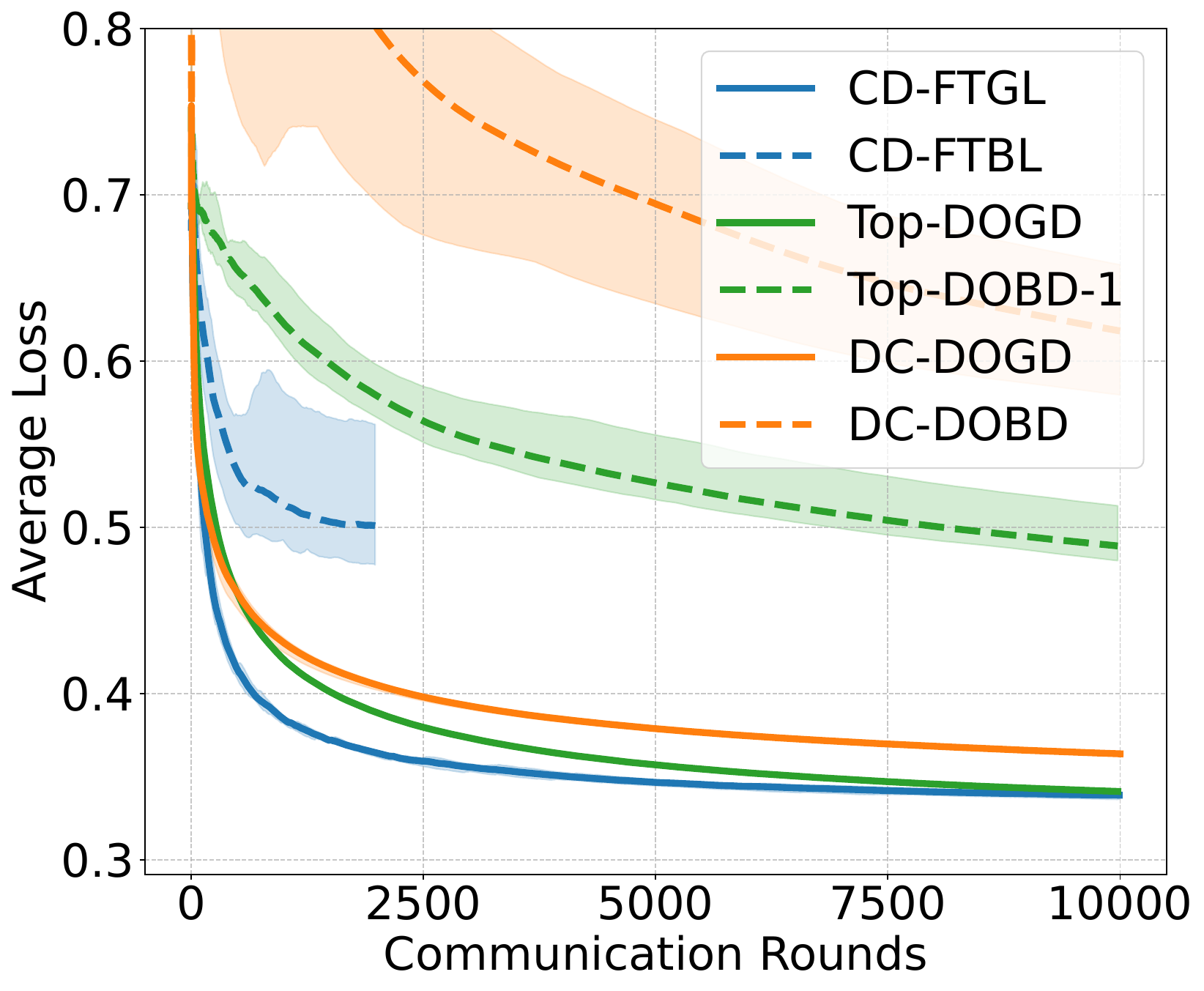}
        \caption{a9a: Loss vs Rounds}
    \end{subfigure}
    \hfill
    % --------- a9a: Loss vs Bits ---------
    \begin{subfigure}[b]{0.24\textwidth}
        \includegraphics[width=\textwidth]{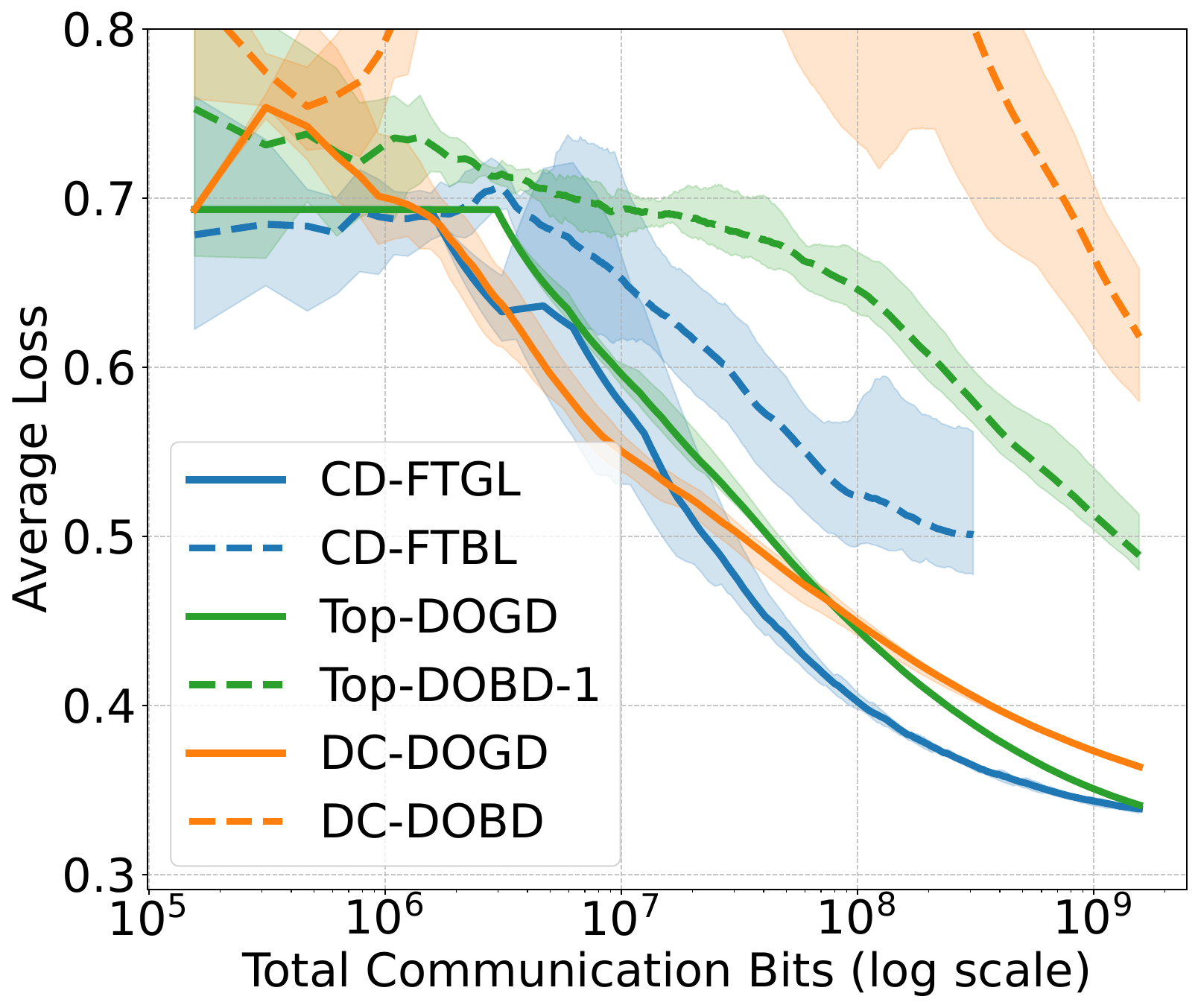}
        \caption{a9a: Loss vs Bits}
    \end{subfigure}
    \caption{Experimental results on a small random graph ($9$ nodes and 18 edges) with $\omega = 0.5$. }
    \label{fig:n9_omega_0.5_results}
\end{figure}

\begin{figure}[htbp]
    \centering
    % --------- ijcnn1: Loss vs Rounds ---------
    \begin{subfigure}[b]{0.24\textwidth}
        \includegraphics[width=\textwidth]{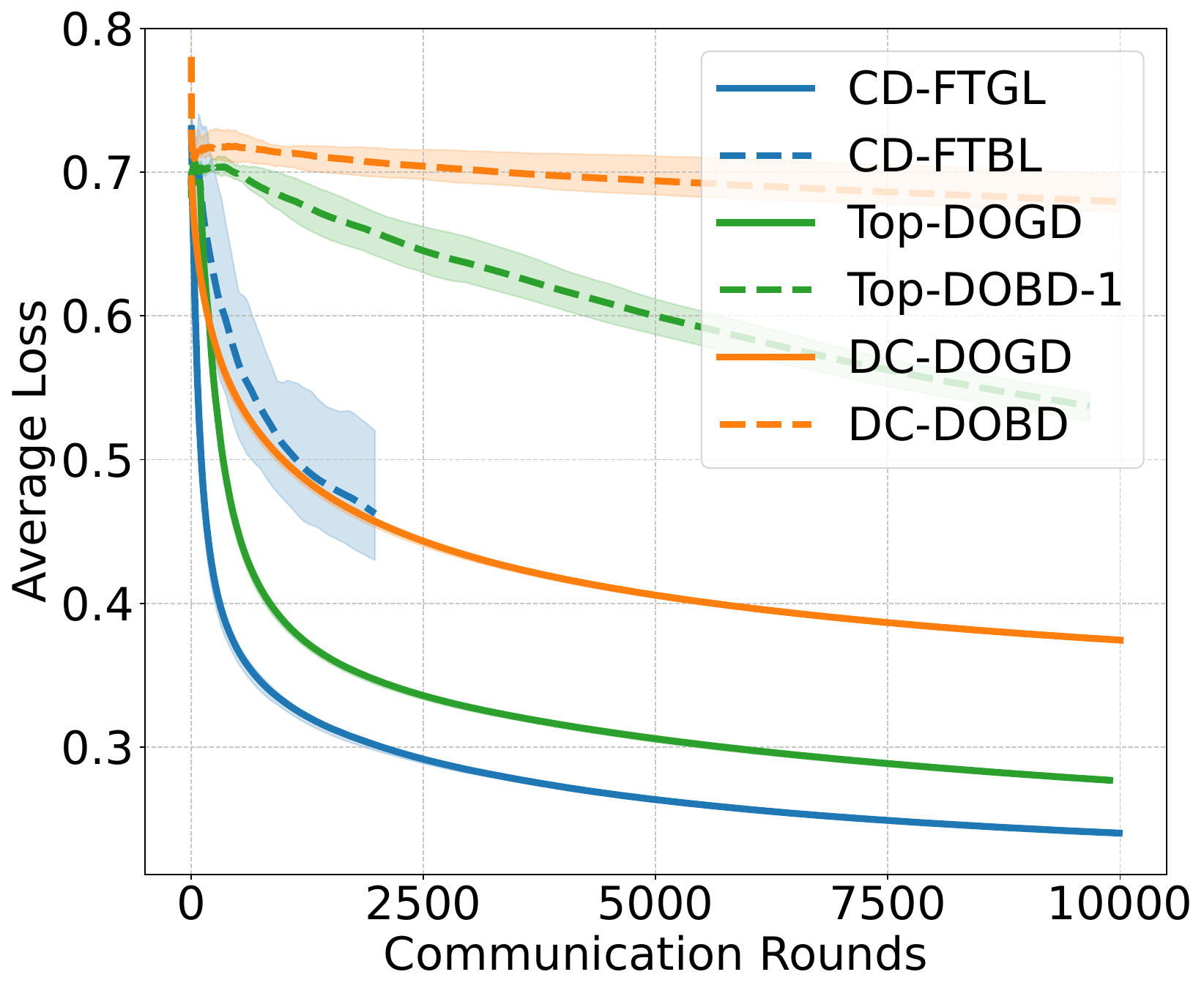}
        \caption{ijcnn1: Loss vs Rounds}
    \end{subfigure}
    \hfill
    % --------- ijcnn1: Loss vs Bits ---------
    \begin{subfigure}[b]{0.24\textwidth}
        \includegraphics[width=\textwidth]{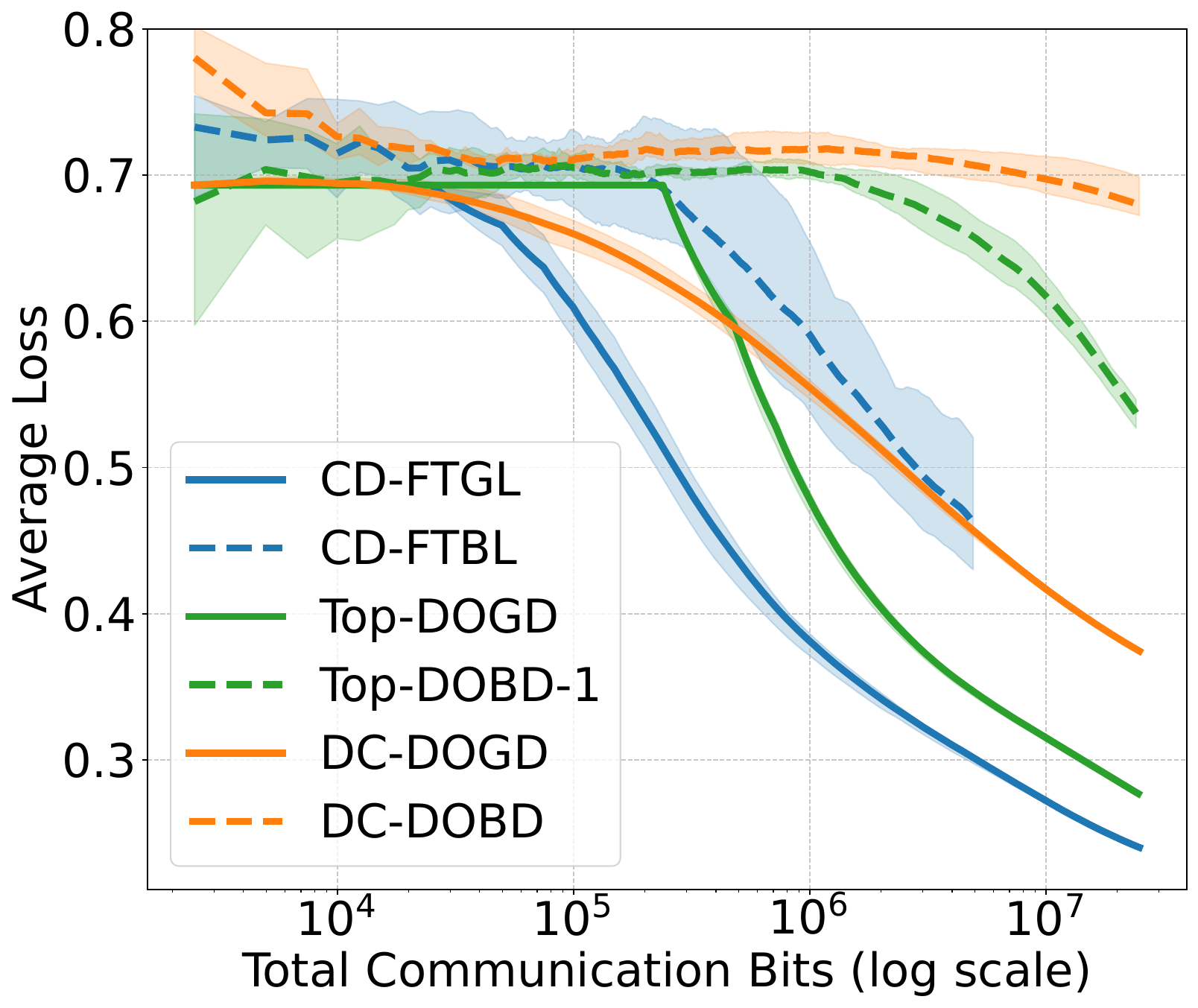}
        \caption{ijcnn1: Loss vs Bits}
    \end{subfigure}
    \hfill
    % --------- a9a: Loss vs Rounds ---------
    \begin{subfigure}[b]{0.24\textwidth}
        \includegraphics[width=\textwidth]{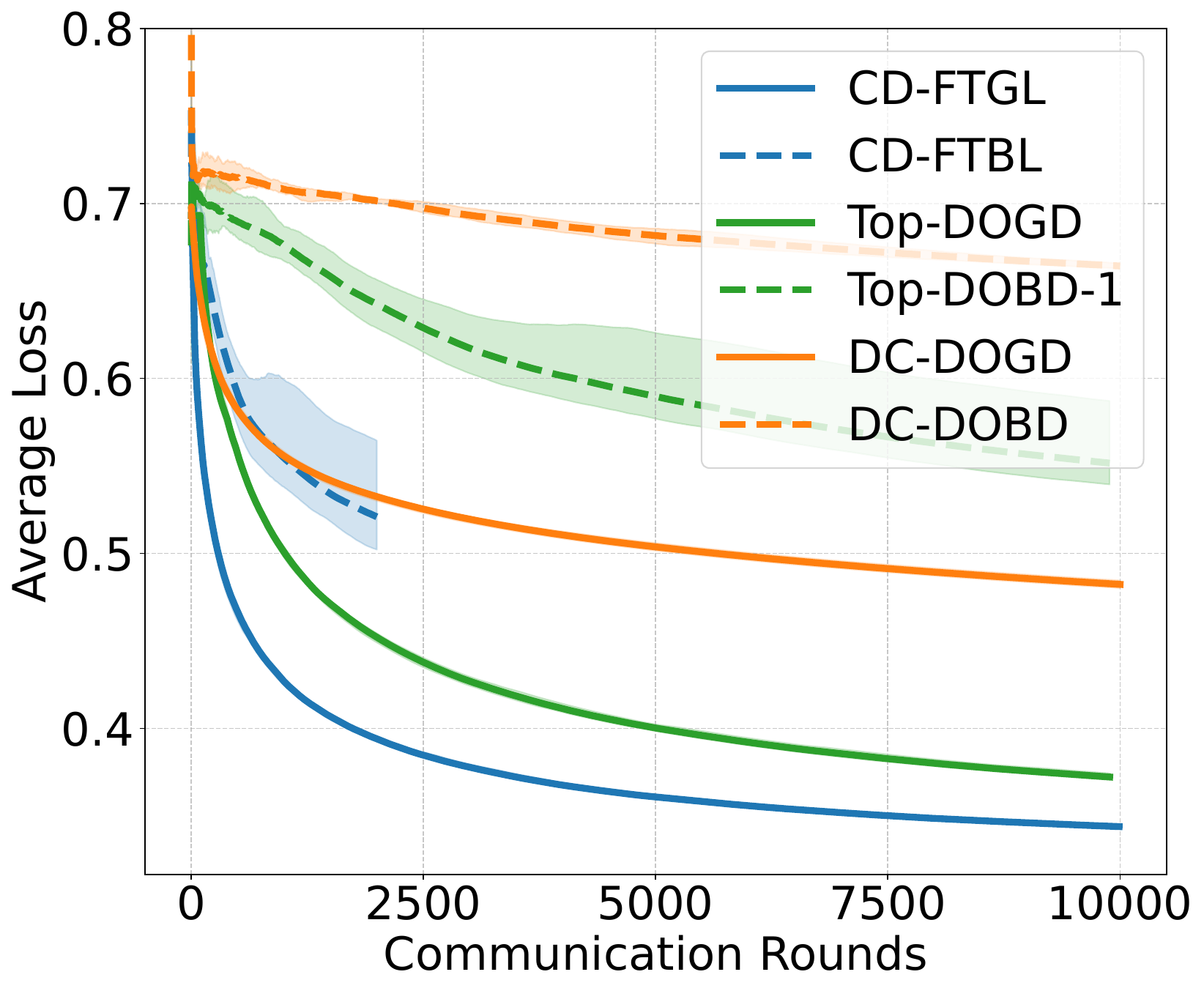}
        \caption{a9a: Loss vs Rounds}
    \end{subfigure}
    \hfill
    % --------- a9a: Loss vs Bits ---------
    \begin{subfigure}[b]{0.24\textwidth}
        \includegraphics[width=\textwidth]{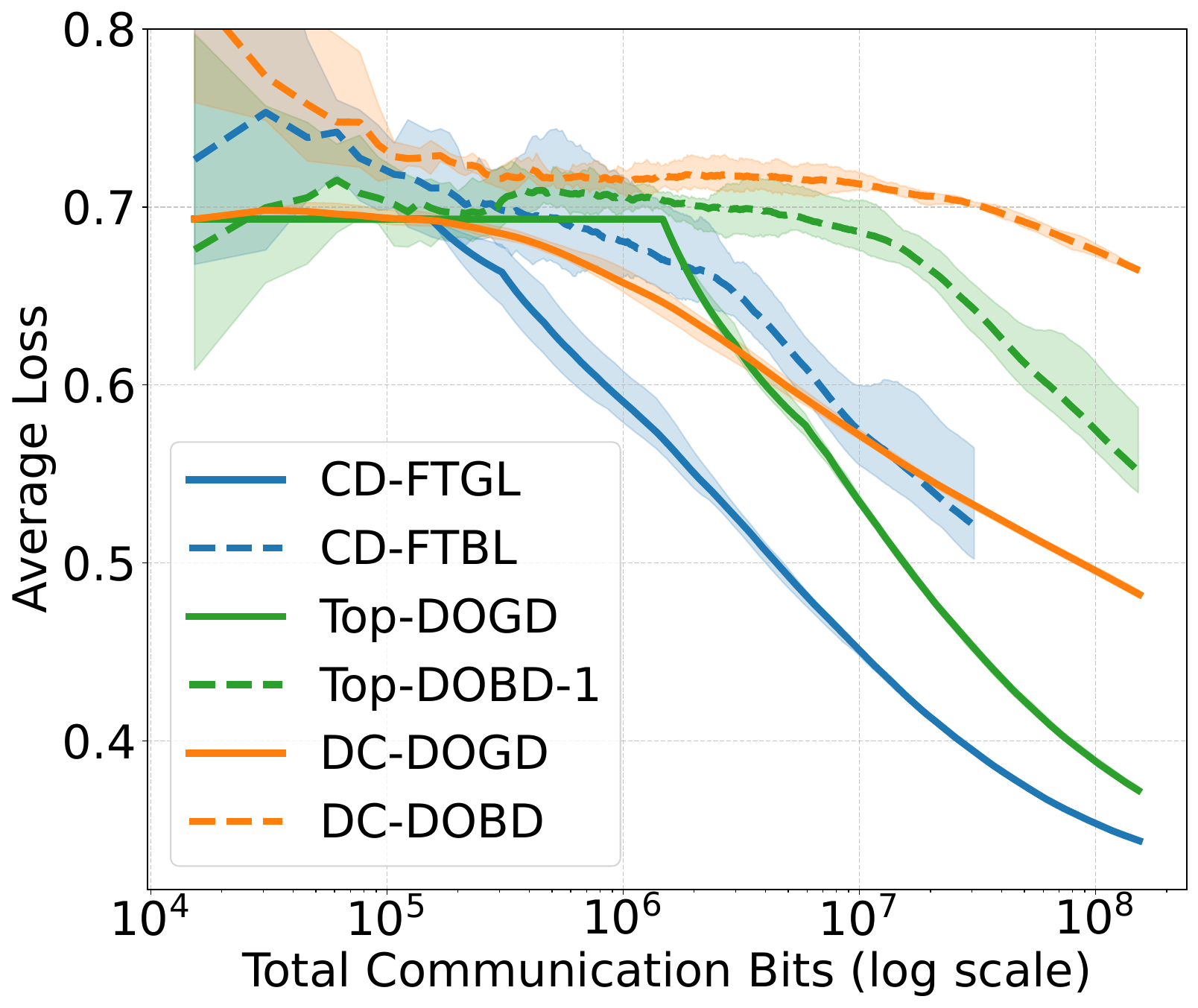}
        \caption{a9a: Loss vs Bits}
    \end{subfigure}
    \caption{Experimental results on a small random graph ($9$ nodes and 18 edges) with $\omega = 0.05$. }
    \label{fig:n9_omega_0.05_results}
\end{figure}

\begin{figure}[htbp]
    \centering
    % --------- ijcnn1: Loss vs Rounds ---------
    \begin{subfigure}[b]{0.24\textwidth}
        \includegraphics[width=\textwidth]{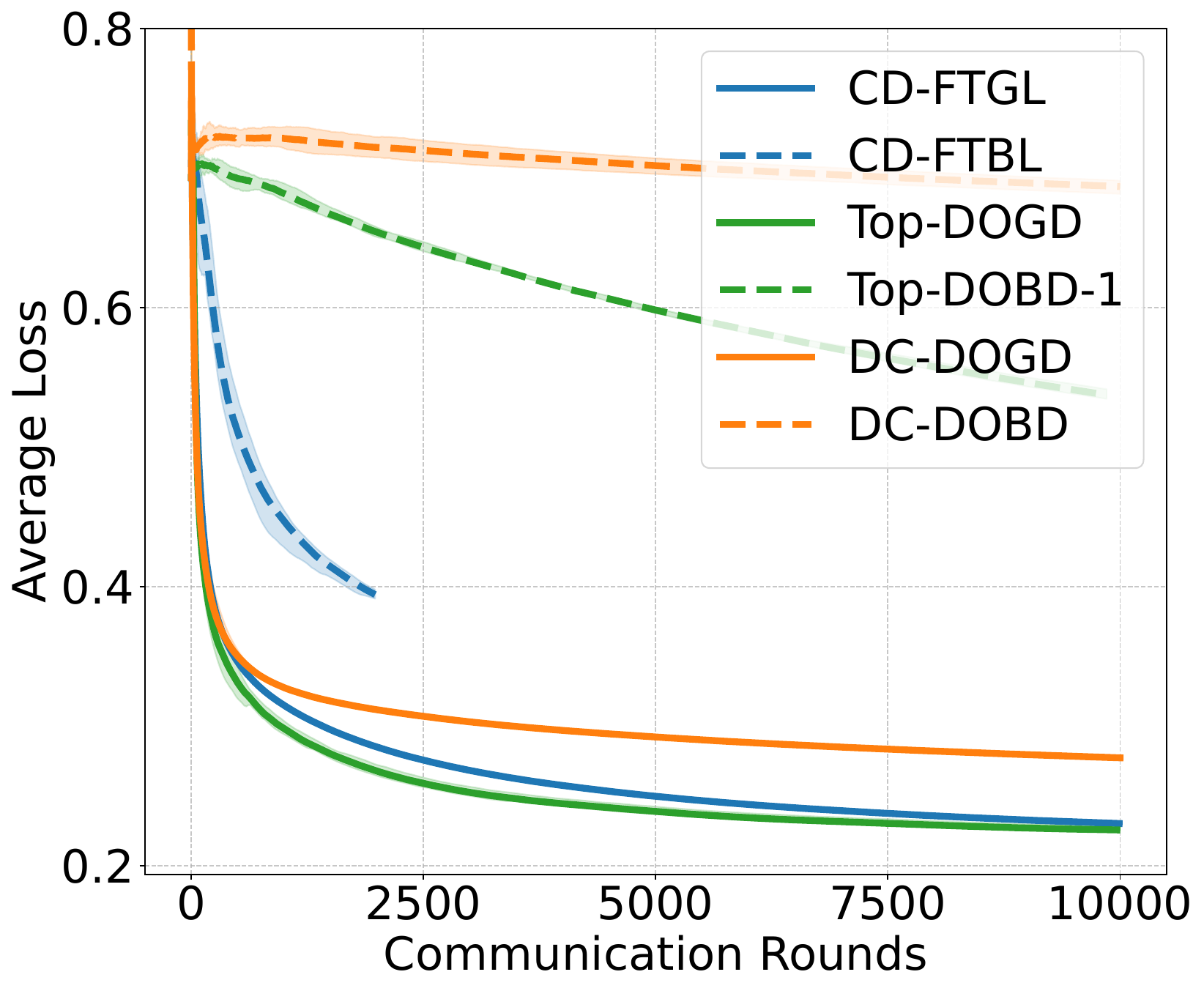}
        \caption{ijcnn1: Loss vs Rounds}
    \end{subfigure}
    \hfill
    % --------- ijcnn1: Loss vs Bits ---------
    \begin{subfigure}[b]{0.24\textwidth}
        \includegraphics[width=\textwidth]{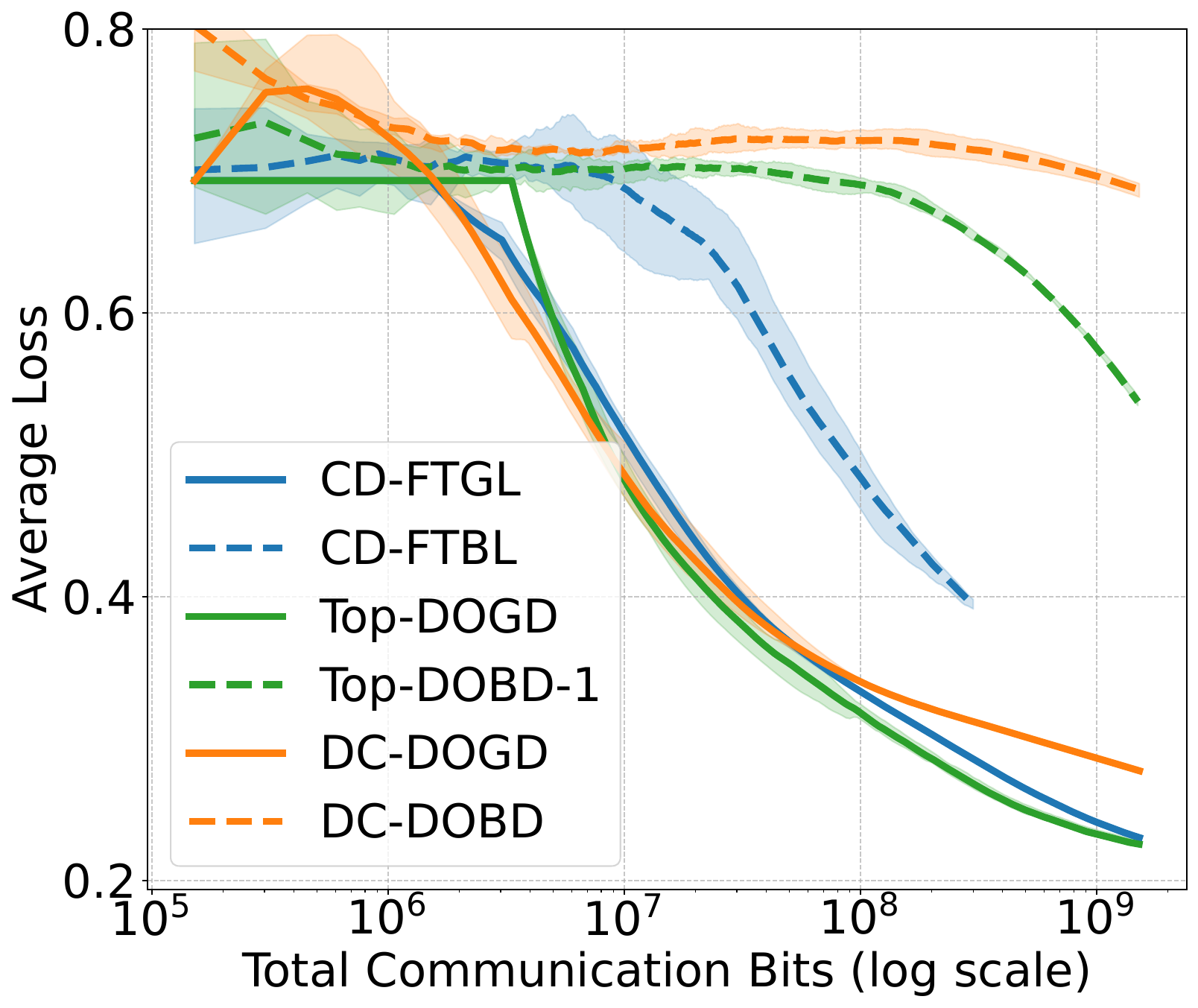}
        \caption{ijcnn1: Loss vs Bits}
    \end{subfigure}
    \hfill
    % --------- a9a: Loss vs Rounds ---------
    \begin{subfigure}[b]{0.24\textwidth}
        \includegraphics[width=\textwidth]{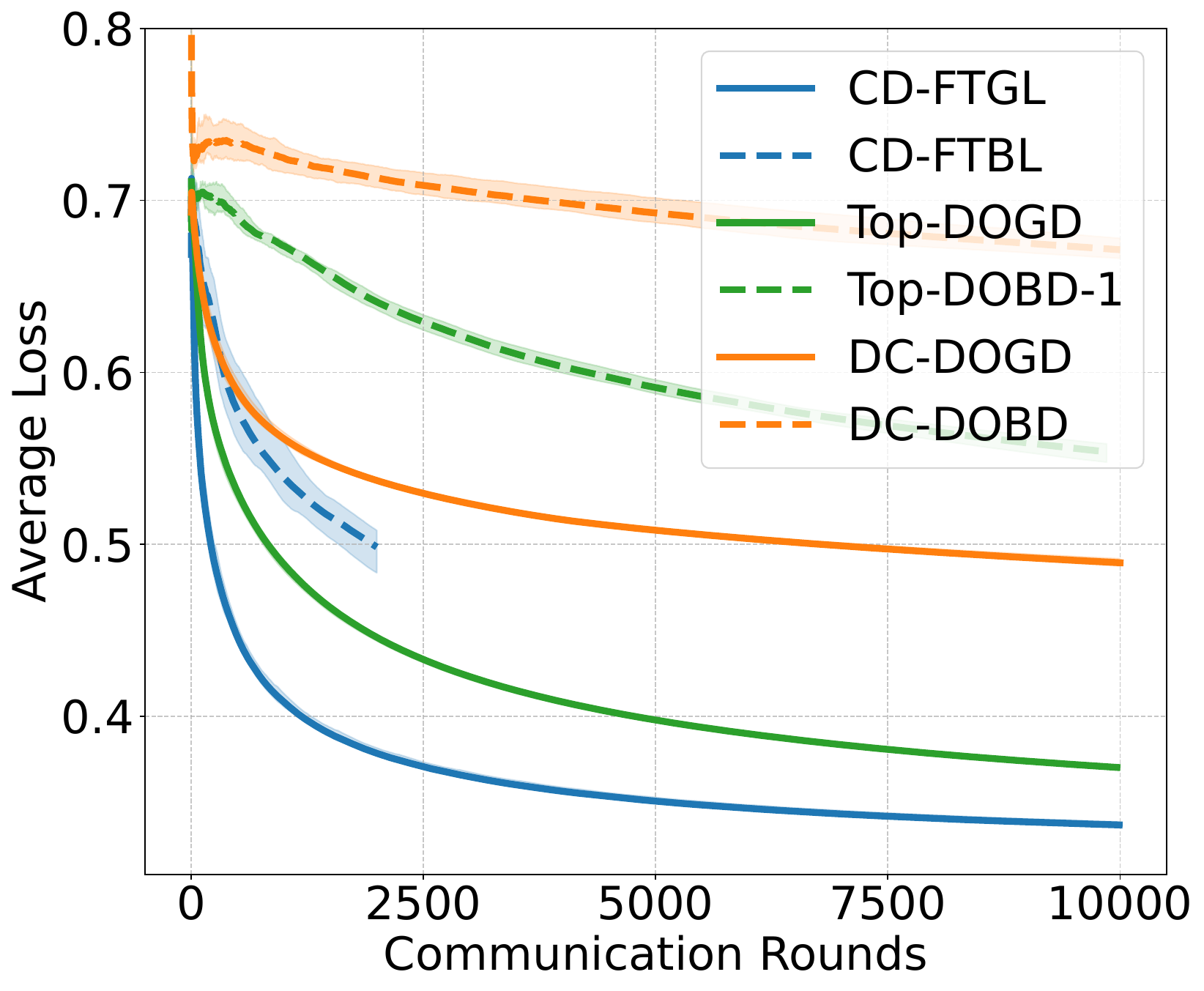}
        \caption{a9a: Loss vs Rounds}
    \end{subfigure}
    \hfill
    % --------- a9a: Loss vs Bits ---------
    \begin{subfigure}[b]{0.24\textwidth}
        \includegraphics[width=\textwidth]{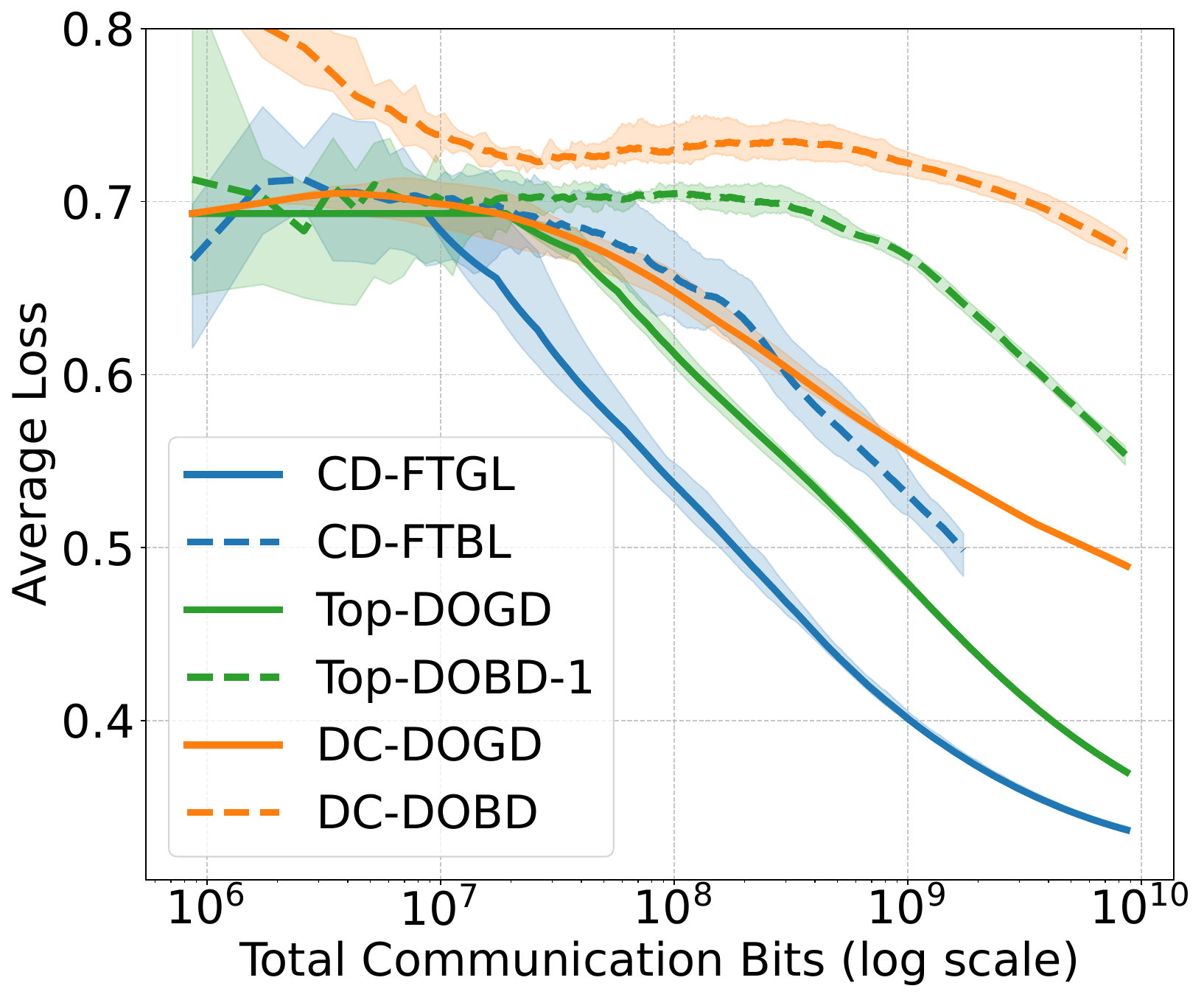}
        \caption{a9a: Loss vs Bits}
    \end{subfigure}
    \caption{Experimental results on a larger random graph ($50$ nodes and 100 edges) with $\omega = 0.5$. }
    \label{fig:n50_omega_0.5_results}
\end{figure}

\begin{figure}[htbp]
    \centering
    % --------- ijcnn1: Loss vs Rounds ---------
    \begin{subfigure}[b]{0.24\textwidth}
        \includegraphics[width=\textwidth]{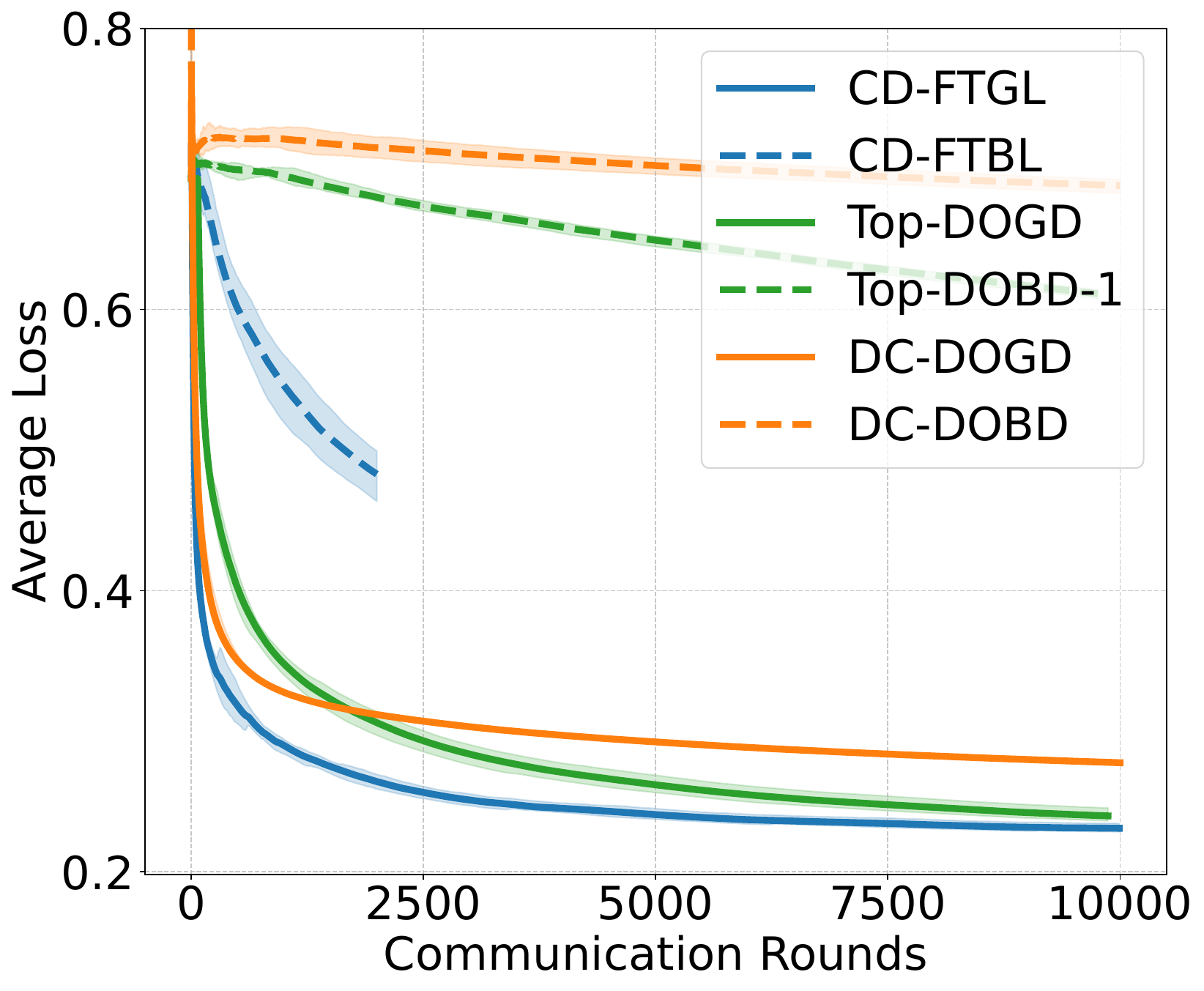}
        \caption{ijcnn1: Loss vs Rounds}
    \end{subfigure}
    \hfill
    % --------- ijcnn1: Loss vs Bits ---------
    \begin{subfigure}[b]{0.24\textwidth}
        \includegraphics[width=\textwidth]{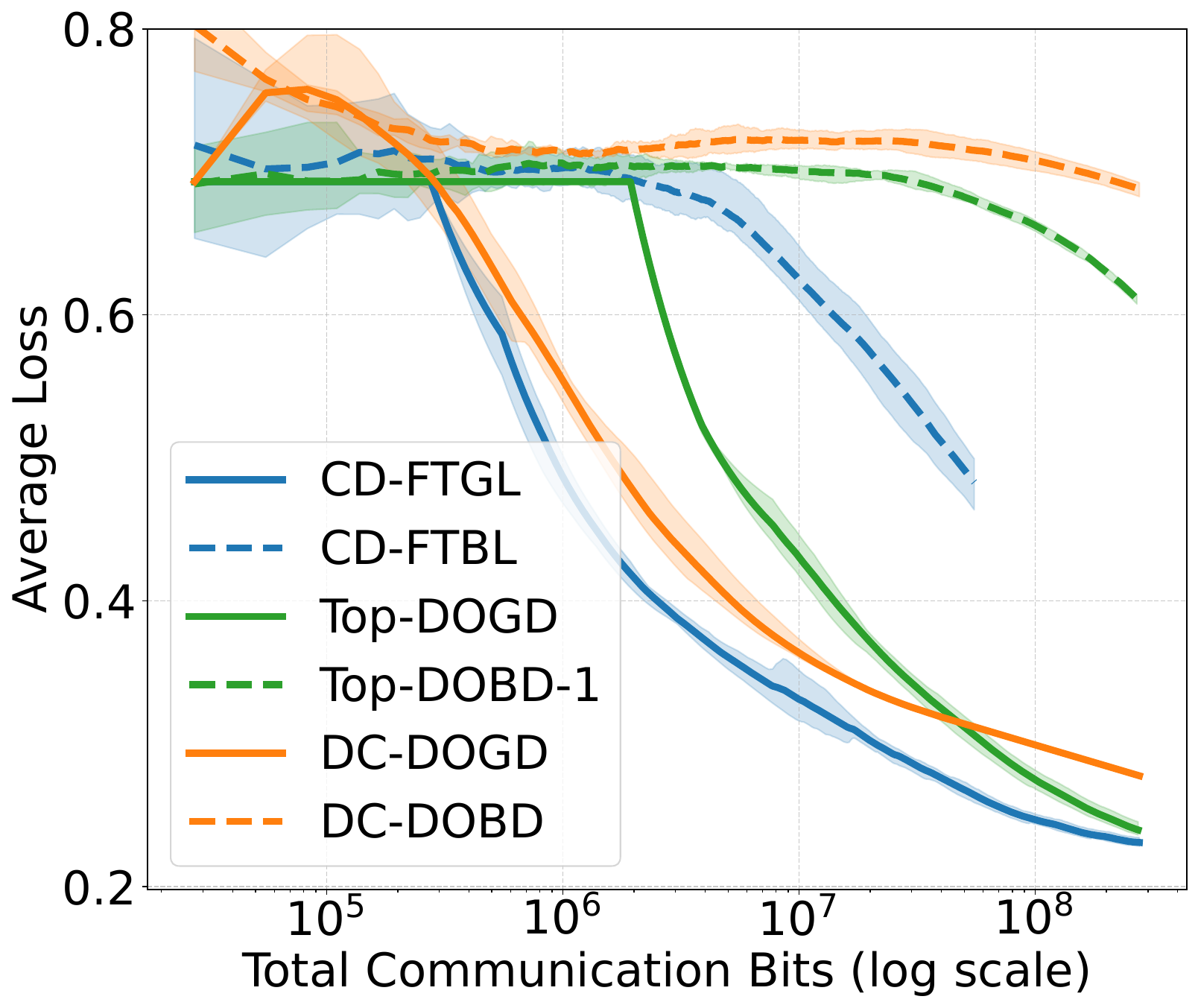}
        \caption{ijcnn1: Loss vs Bits}
    \end{subfigure}
    \hfill
    % --------- a9a: Loss vs Rounds ---------
    \begin{subfigure}[b]{0.24\textwidth}
        \includegraphics[width=\textwidth]{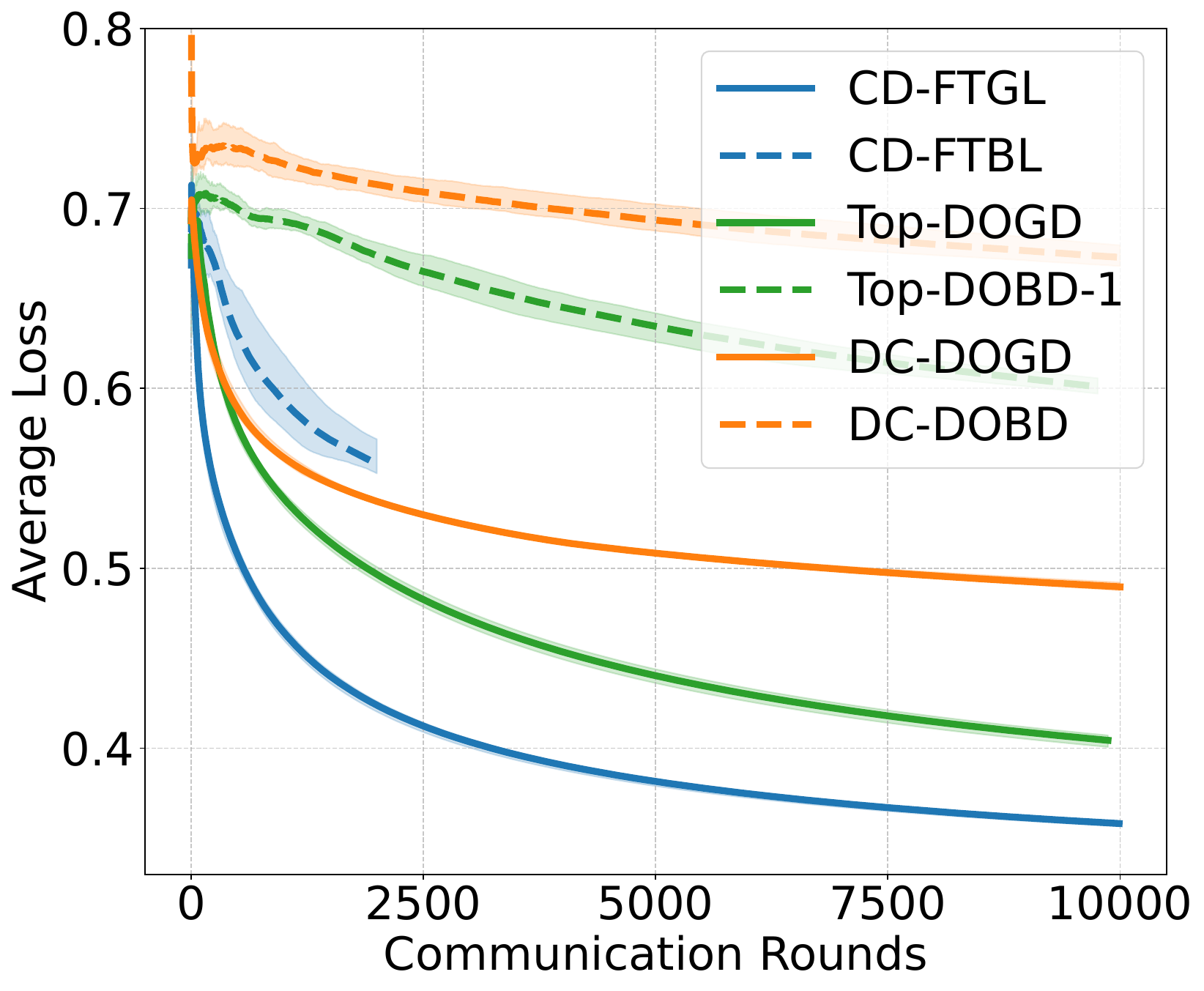}
        \caption{a9a: Loss vs Rounds}
    \end{subfigure}
    \hfill
    % --------- a9a: Loss vs Bits ---------
    \begin{subfigure}[b]{0.24\textwidth}
        \includegraphics[width=\textwidth]{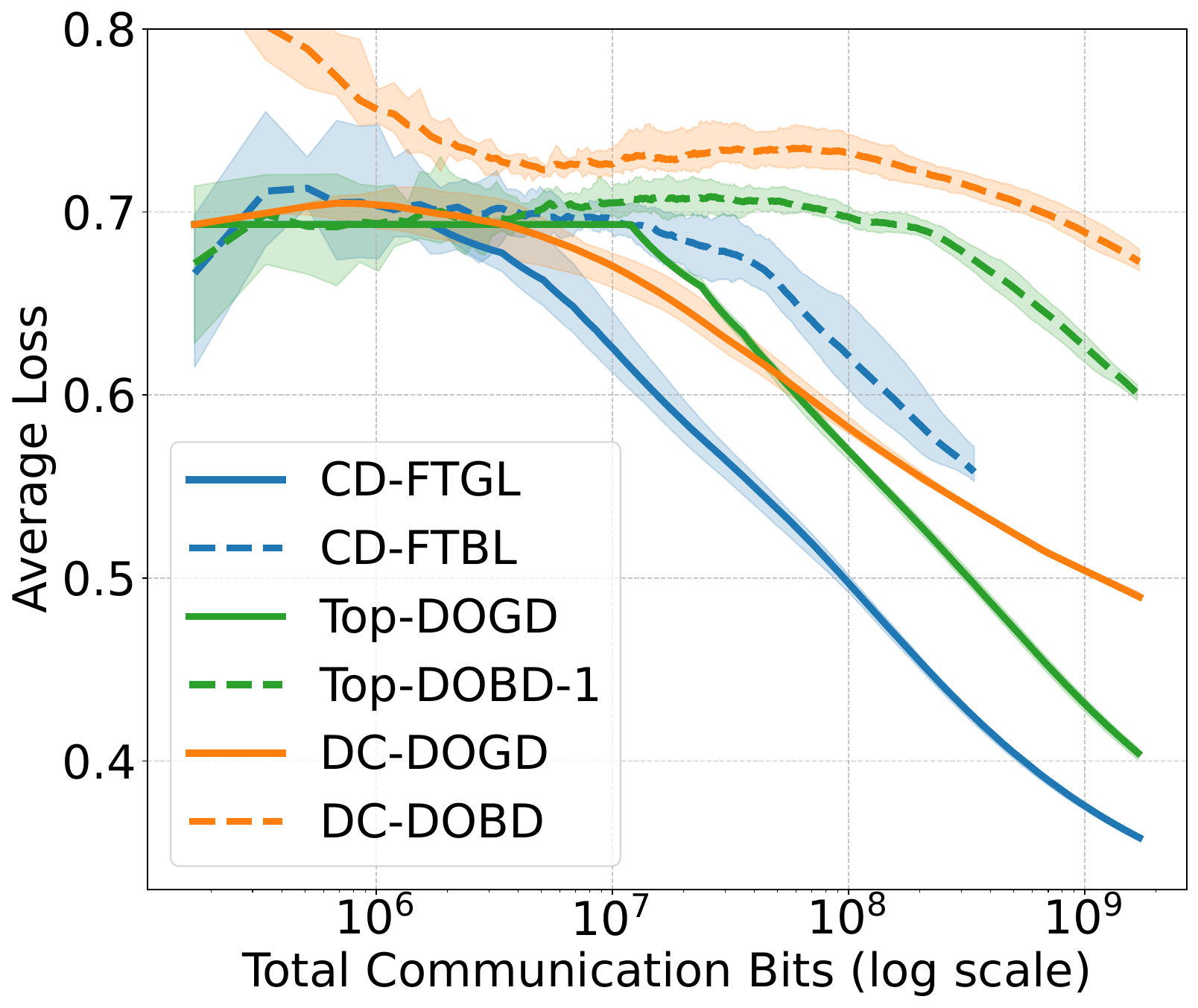}
        \caption{a9a: Loss vs Bits}
    \end{subfigure}
    \caption{Experimental results on a larger random graph ($50$ nodes and 100 edges) with $\omega = 0.1$. }
    \label{fig:n50_omega_0.1_results}
\end{figure}

\begin{figure}[htbp]
    \centering
    % --------- ijcnn1: Loss vs Rounds ---------
    \begin{subfigure}[b]{0.24\textwidth}
        \includegraphics[width=\textwidth]{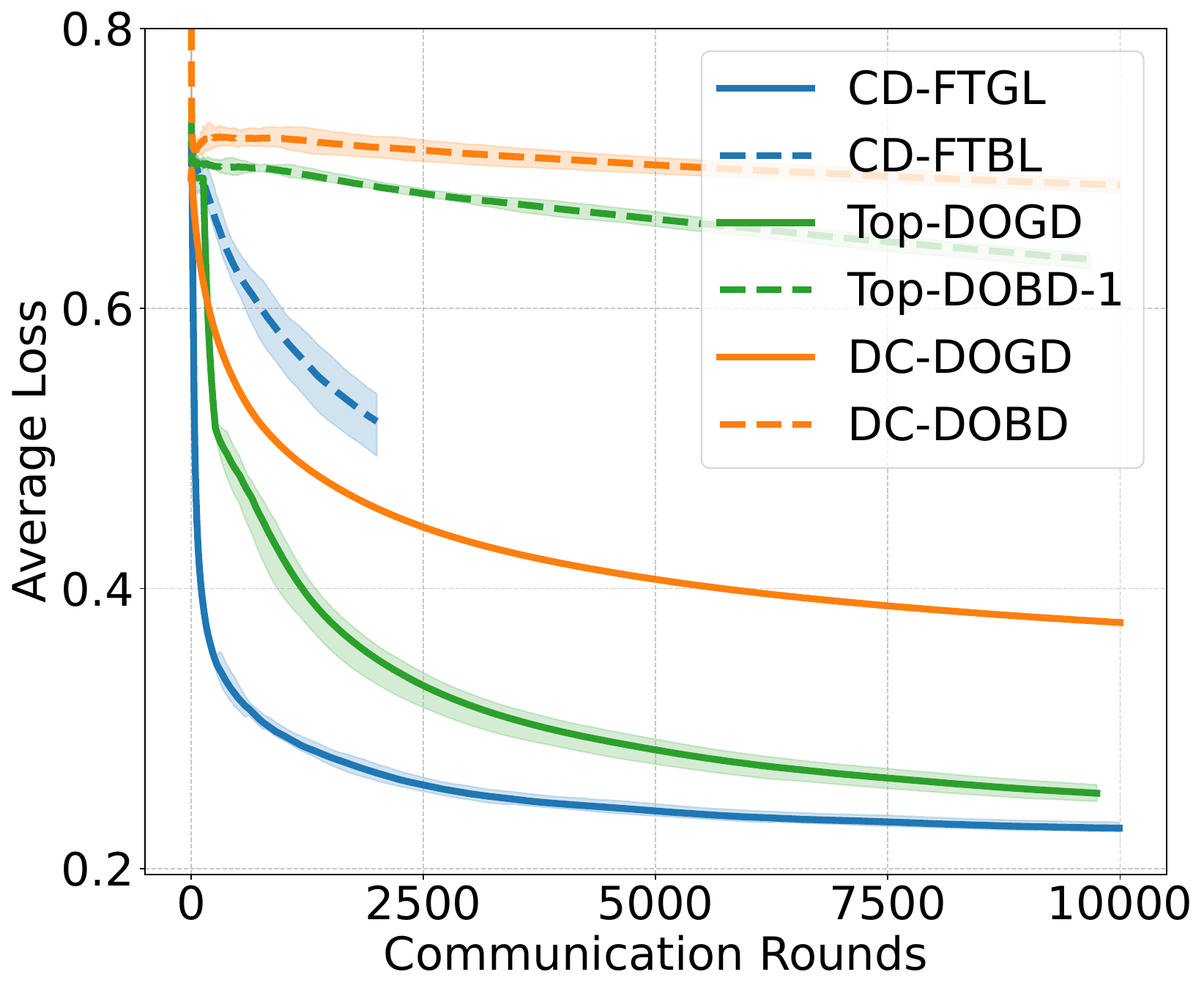}
        \caption{ijcnn1: Loss vs Rounds}
    \end{subfigure}
    \hfill
    % --------- ijcnn1: Loss vs Bits ---------
    \begin{subfigure}[b]{0.24\textwidth}
        \includegraphics[width=\textwidth]{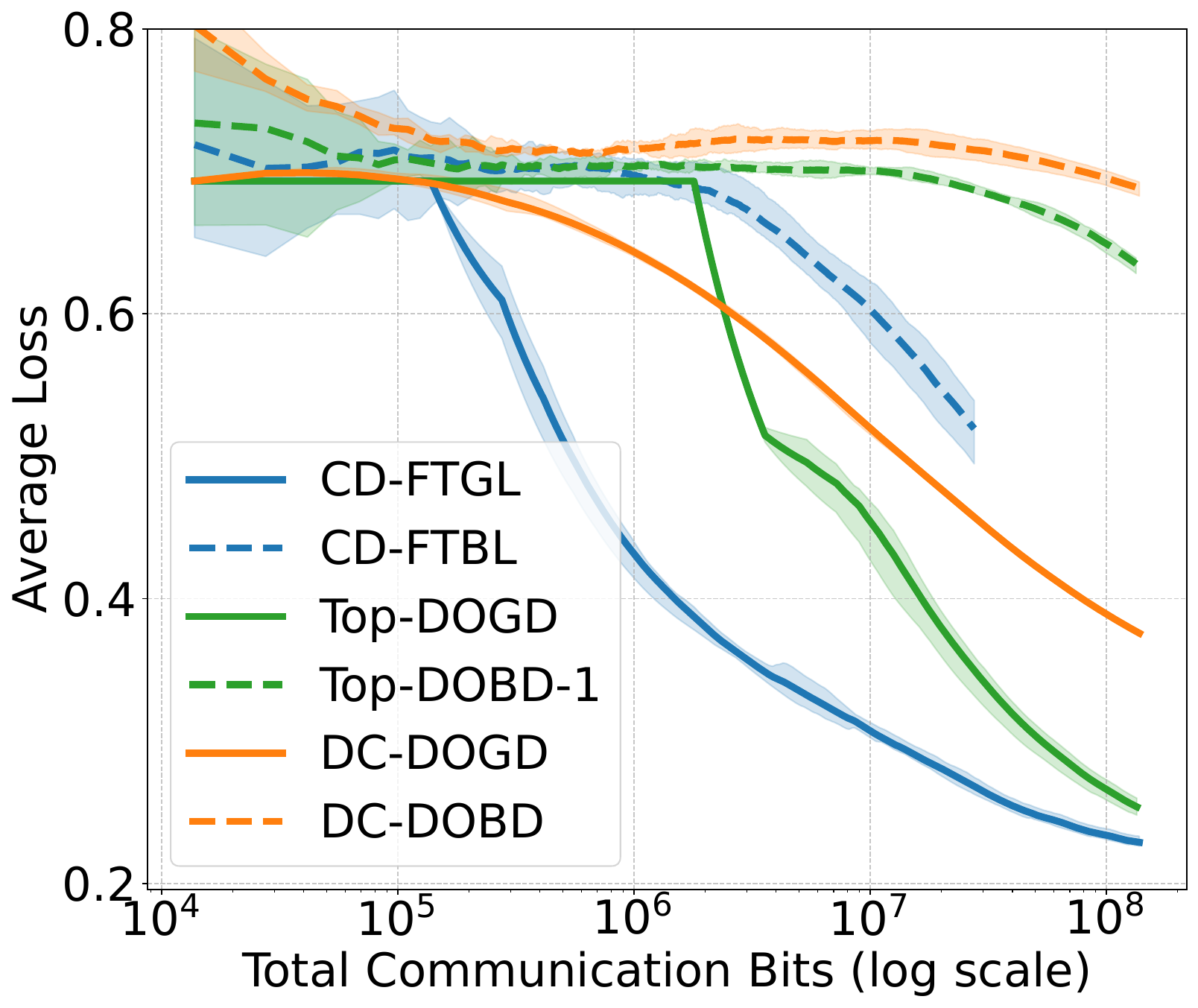}
        \caption{ijcnn1: Loss vs Bits}
    \end{subfigure}
    \hfill
    % --------- a9a: Loss vs Rounds ---------
    \begin{subfigure}[b]{0.24\textwidth}
        \includegraphics[width=\textwidth]{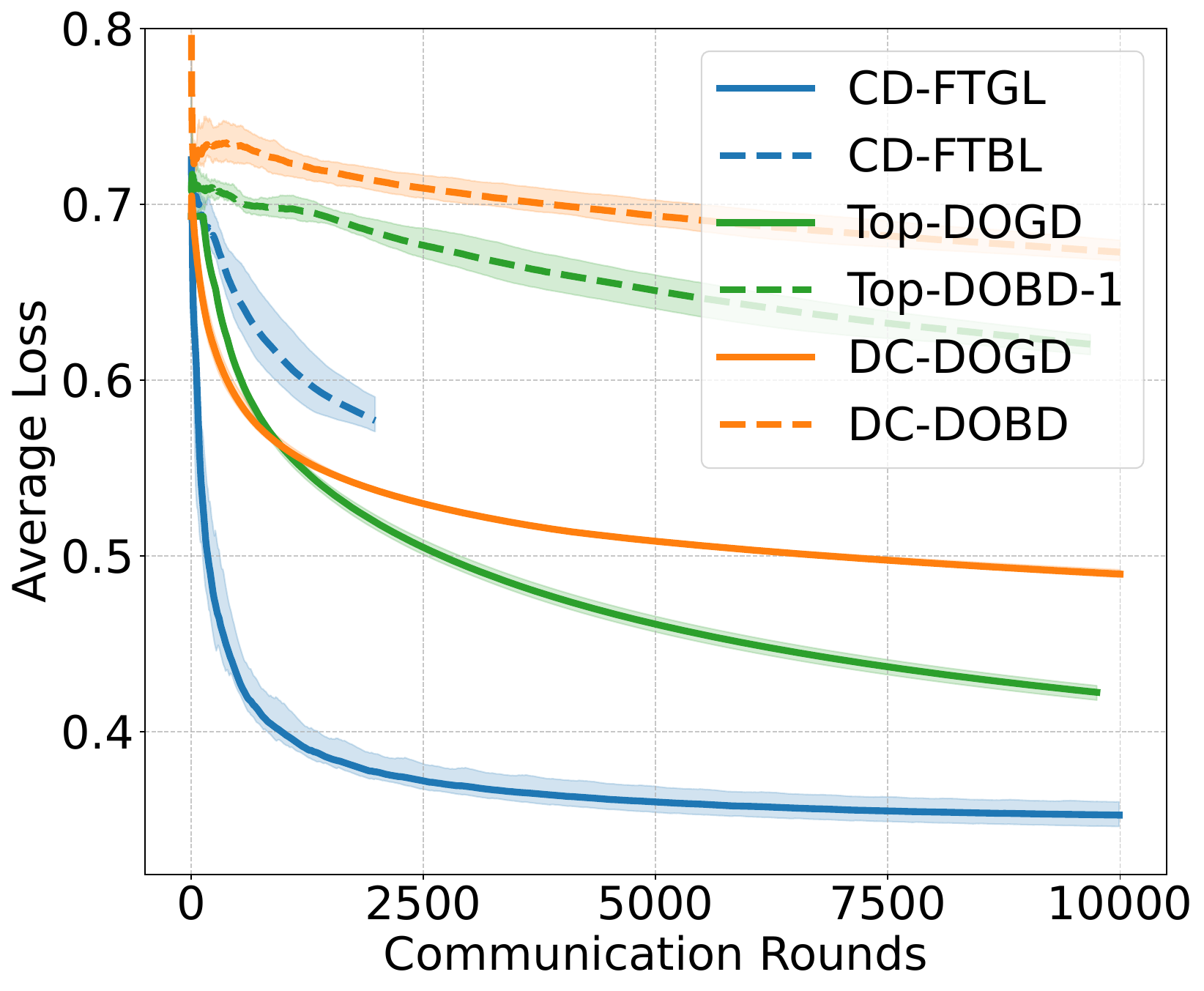}
        \caption{a9a: Loss vs Rounds}
    \end{subfigure}
    \hfill
    % --------- a9a: Loss vs Bits ---------
    \begin{subfigure}[b]{0.24\textwidth}
        \includegraphics[width=\textwidth]{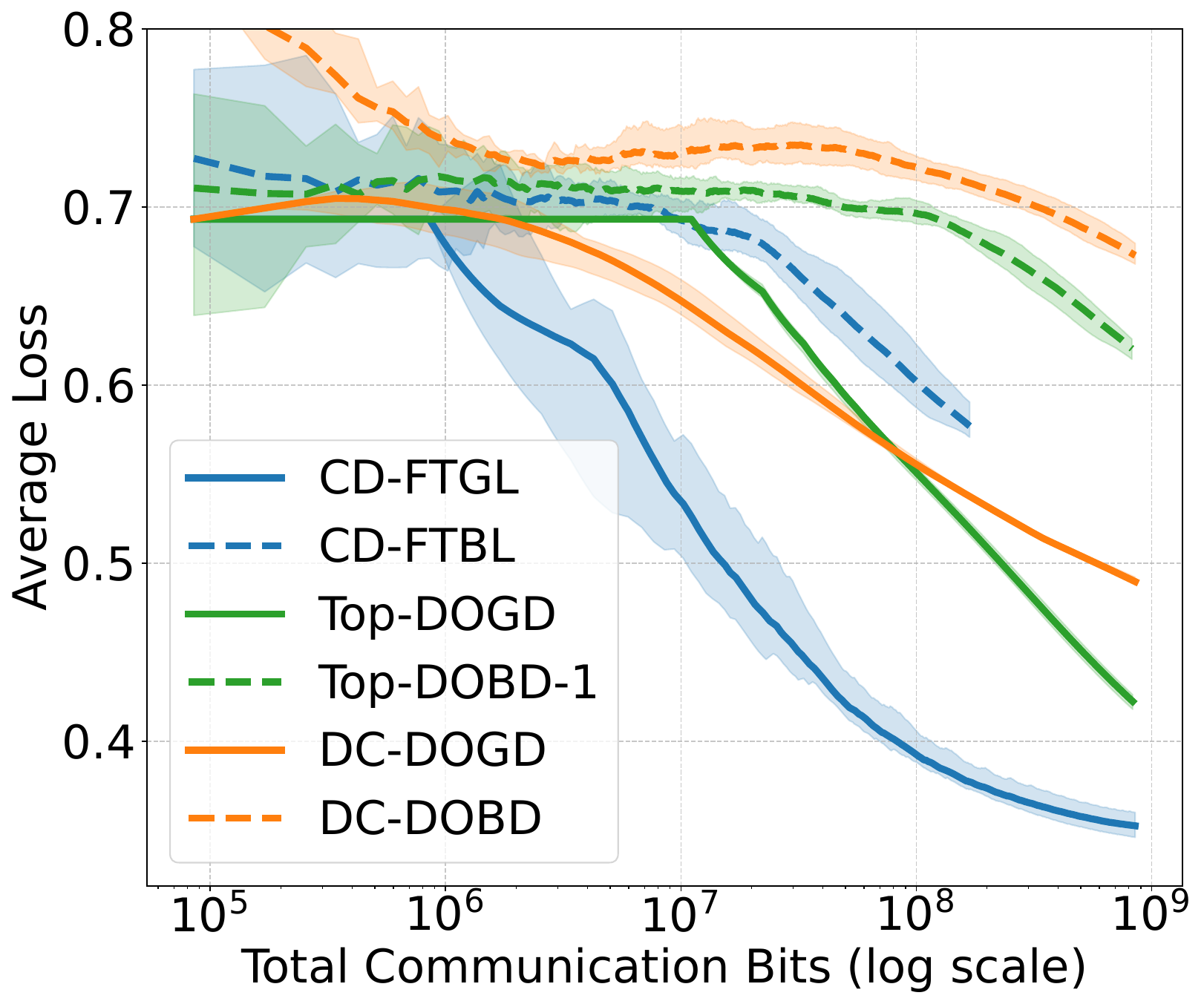}
        \caption{a9a: Loss vs Bits}
    \end{subfigure}
    \caption{Experimental results on a larger random graph ($50$ nodes and 100 edges) with $\omega = 0.05$. }
    \label{fig:n50_omega_0.05_results}
\end{figure}

% ==========================================
% 新增图表区 (Sensitivity Analysis for K)
% ==========================================

\begin{figure}[htbp]
    \centering
    \begin{subfigure}[b]{0.24\textwidth}
        \centering
        \includegraphics[width=\textwidth]{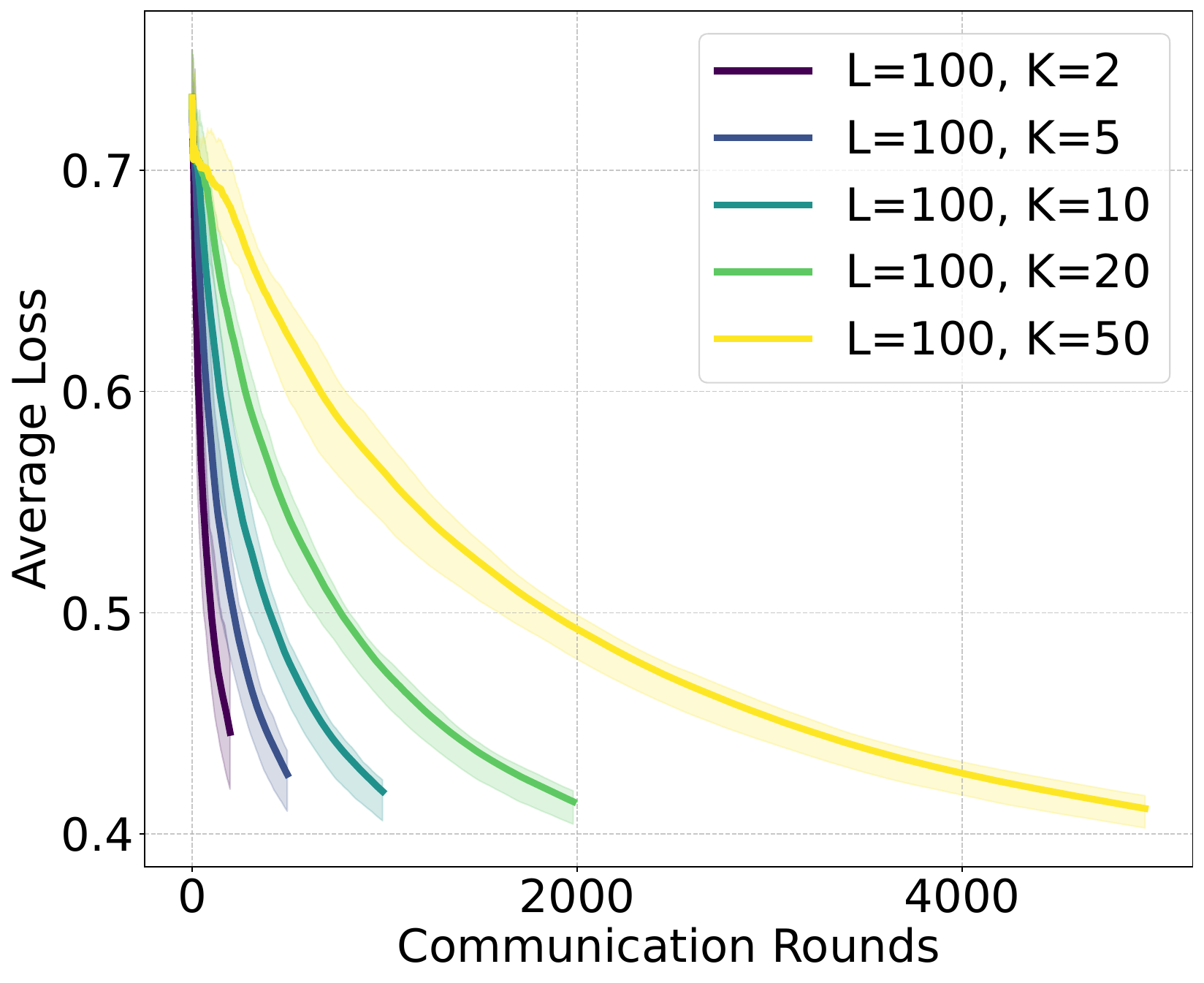}
        \caption{ijcnn1: Loss vs Rounds}
    \end{subfigure}
    \hfill
    \begin{subfigure}[b]{0.24\textwidth}
        \centering
        \includegraphics[width=\textwidth]{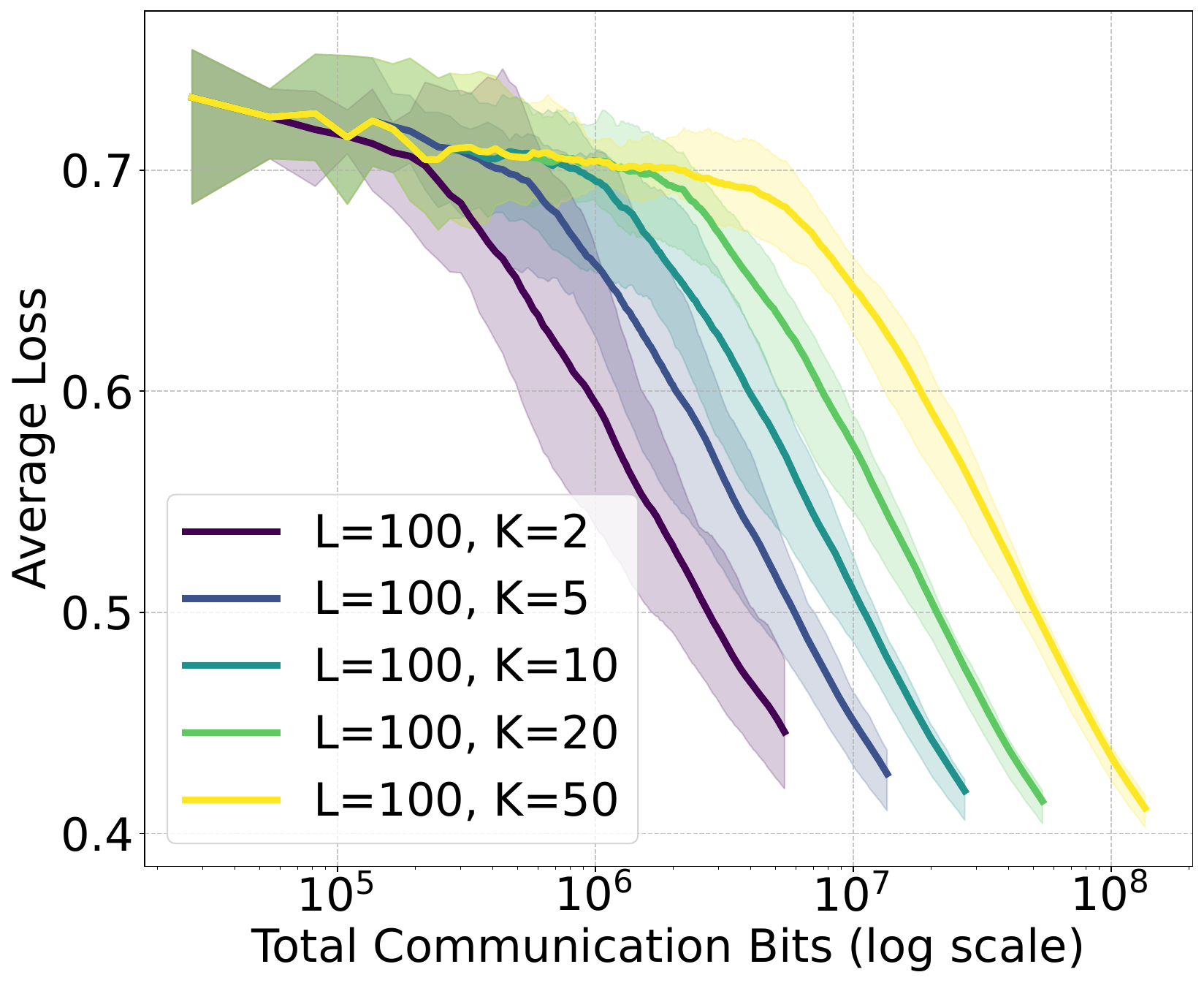}
        \caption{ijcnn1: Loss vs Bits}
    \end{subfigure}
    \hfill
    \begin{subfigure}[b]{0.24\textwidth}
        \centering
        \includegraphics[width=\textwidth]{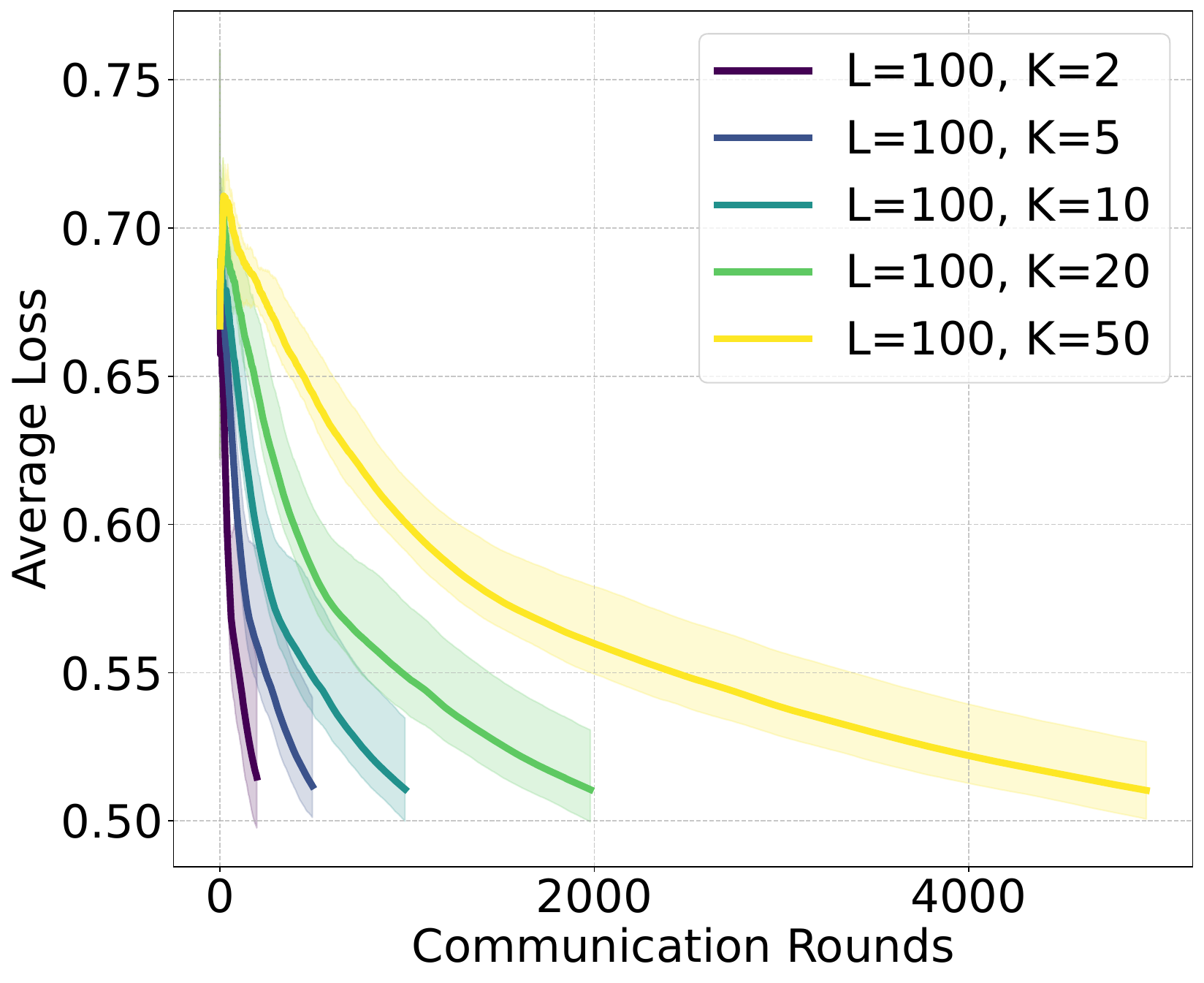}
        \caption{a9a: Loss vs Rounds}
    \end{subfigure}
    \hfill
    \begin{subfigure}[b]{0.24\textwidth}
        \centering
        \includegraphics[width=\textwidth]{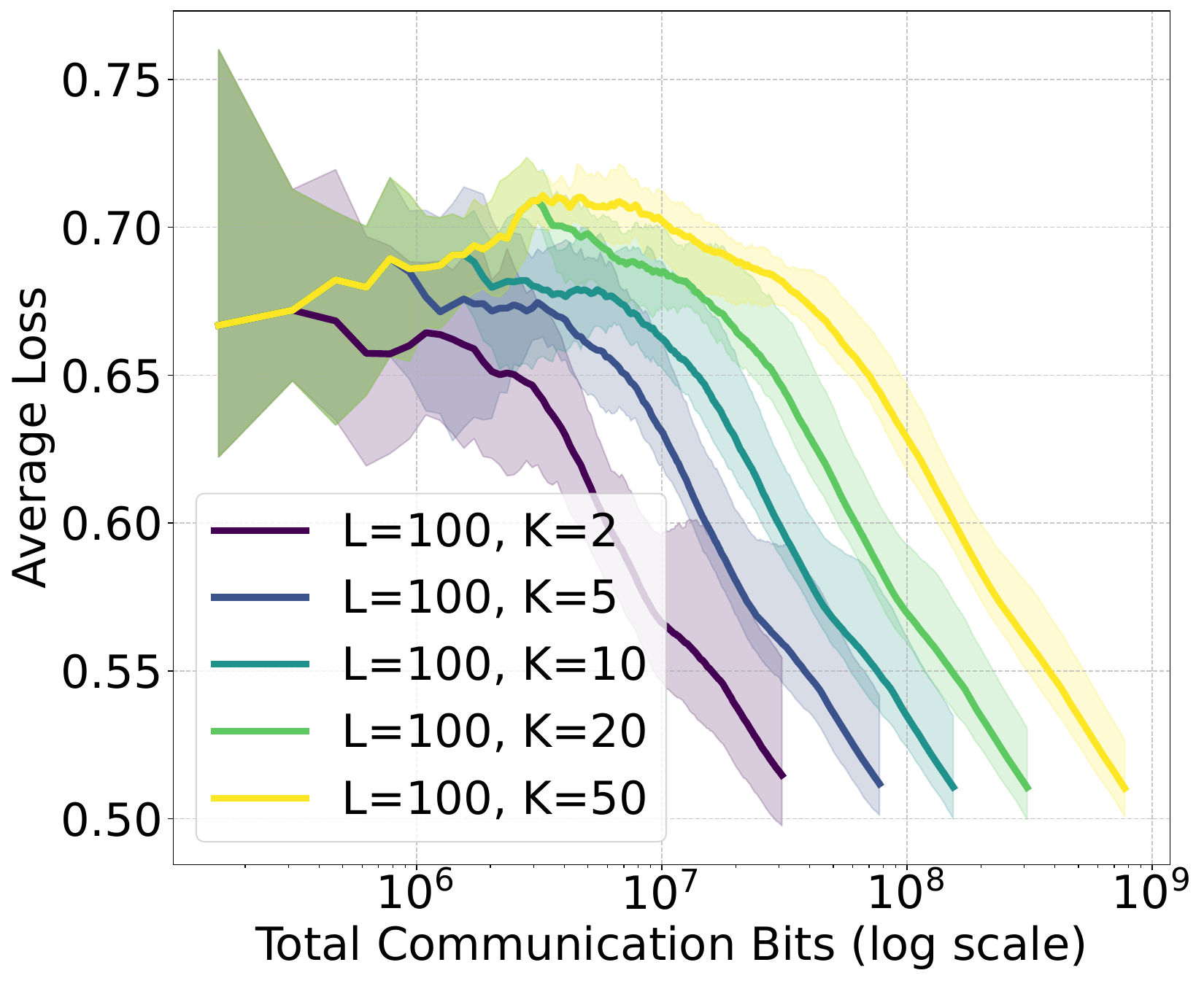}
        \caption{a9a: Loss vs Bits}
    \end{subfigure}
    \caption{Experimental results of CD-FTBL with varying communication rounds $K$ on a small random graph (9 nodes and 18 edges) with $\omega=0.5$ and fixed block size $L=100$.}
    \label{fig:sensitivity_K_w0.5}
\end{figure}

\begin{figure}[htbp]
    \centering
    \begin{subfigure}[b]{0.24\textwidth}
        \centering
        \includegraphics[width=\textwidth]{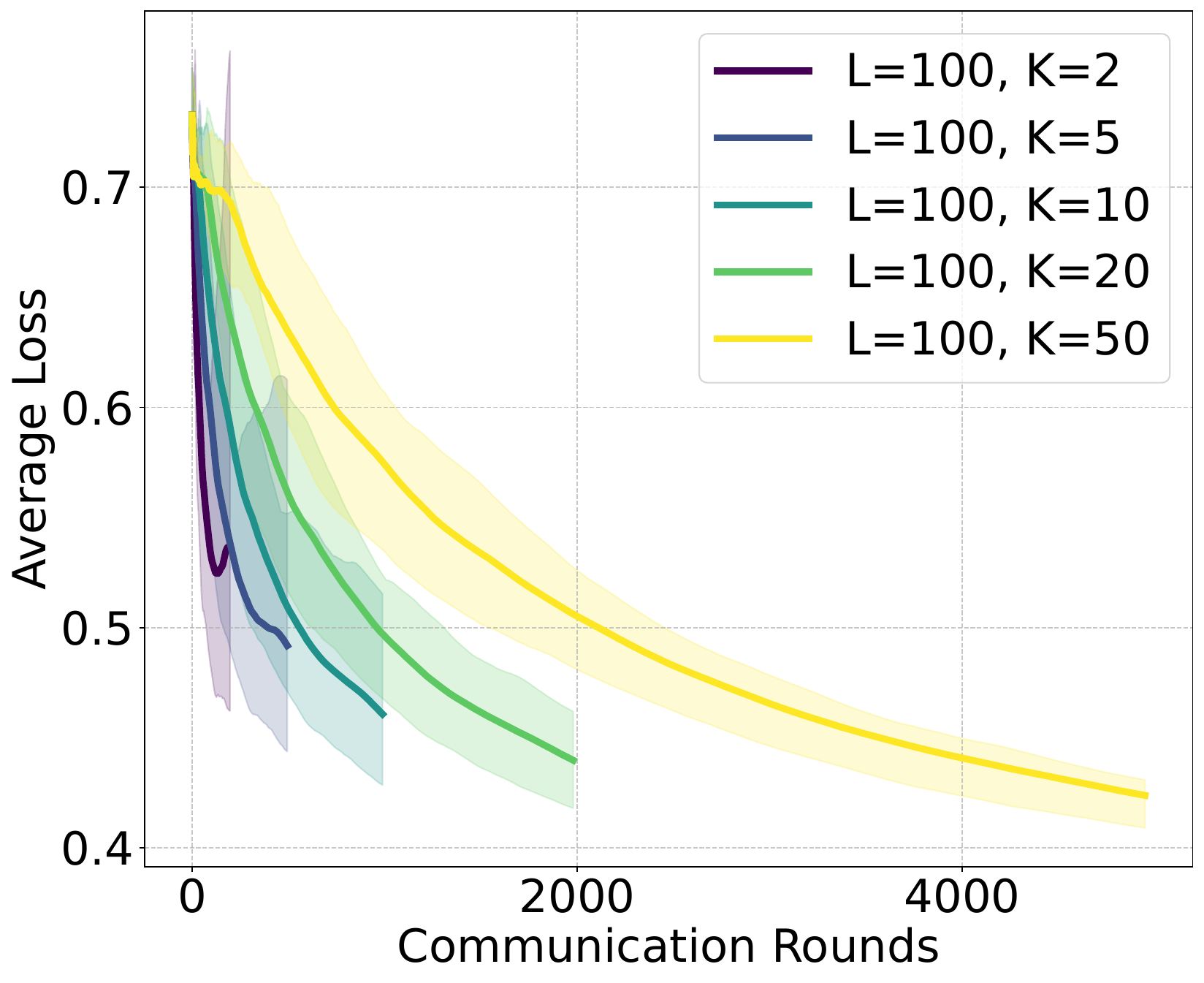}
        \caption{ijcnn1: Loss vs Rounds}
    \end{subfigure}
    \hfill
    \begin{subfigure}[b]{0.24\textwidth}
        \centering
        \includegraphics[width=\textwidth]{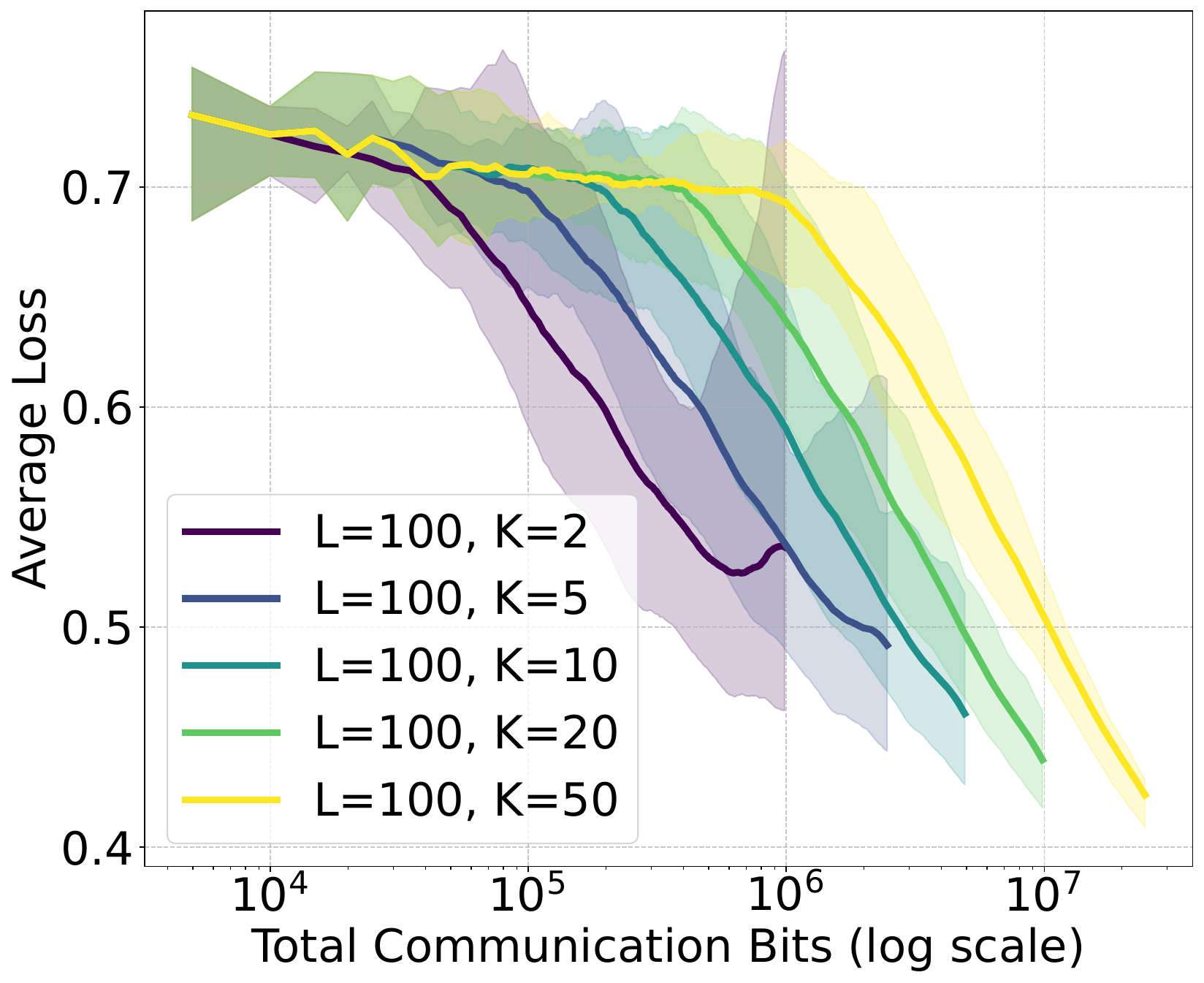}
        \caption{ijcnn1: Loss vs Bits}
    \end{subfigure}
    \hfill
    \begin{subfigure}[b]{0.24\textwidth}
        \centering
        \includegraphics[width=\textwidth]{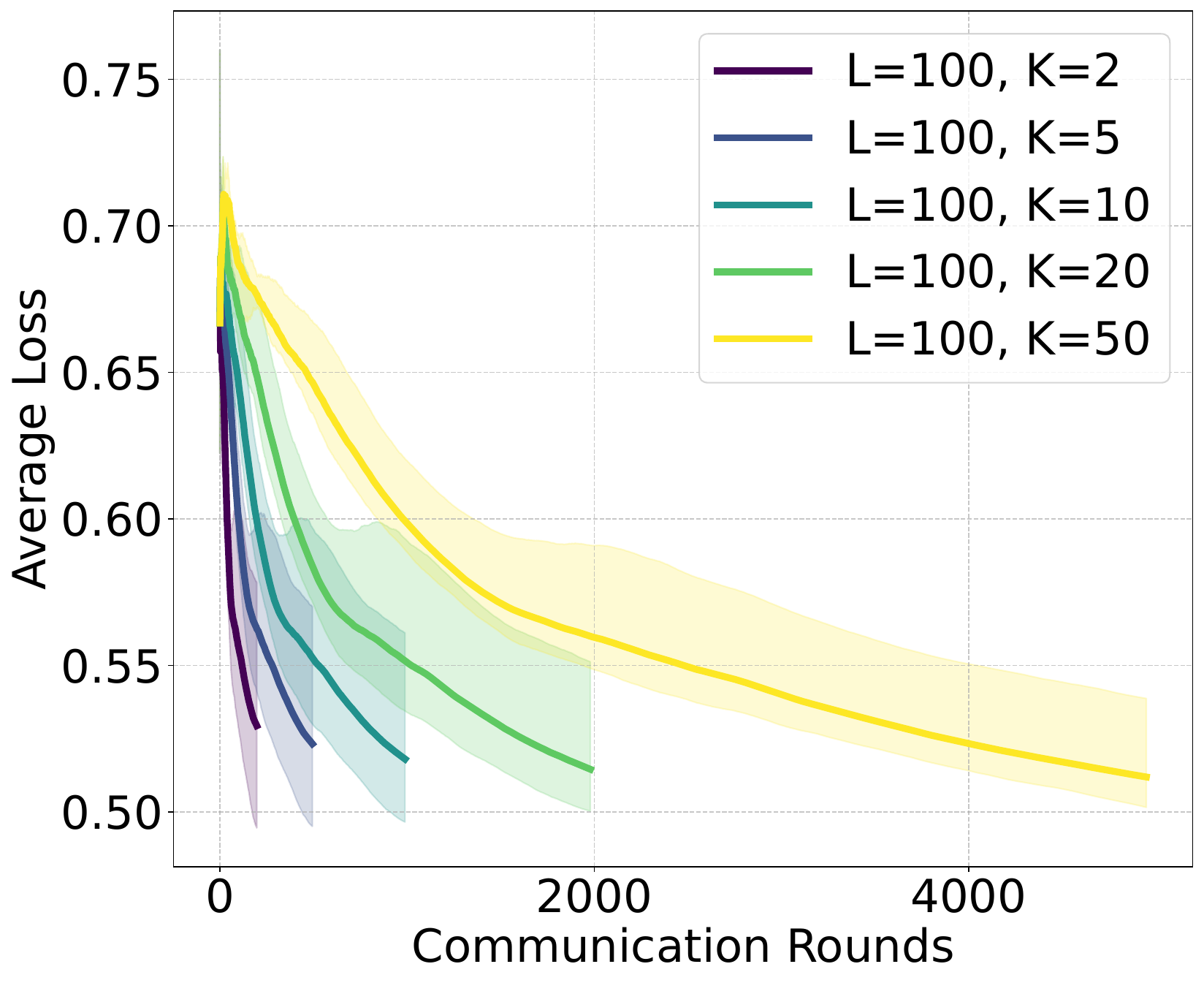}
        \caption{a9a: Loss vs Rounds}
    \end{subfigure}
    \hfill
    \begin{subfigure}[b]{0.24\textwidth}
        \centering
        \includegraphics[width=\textwidth]{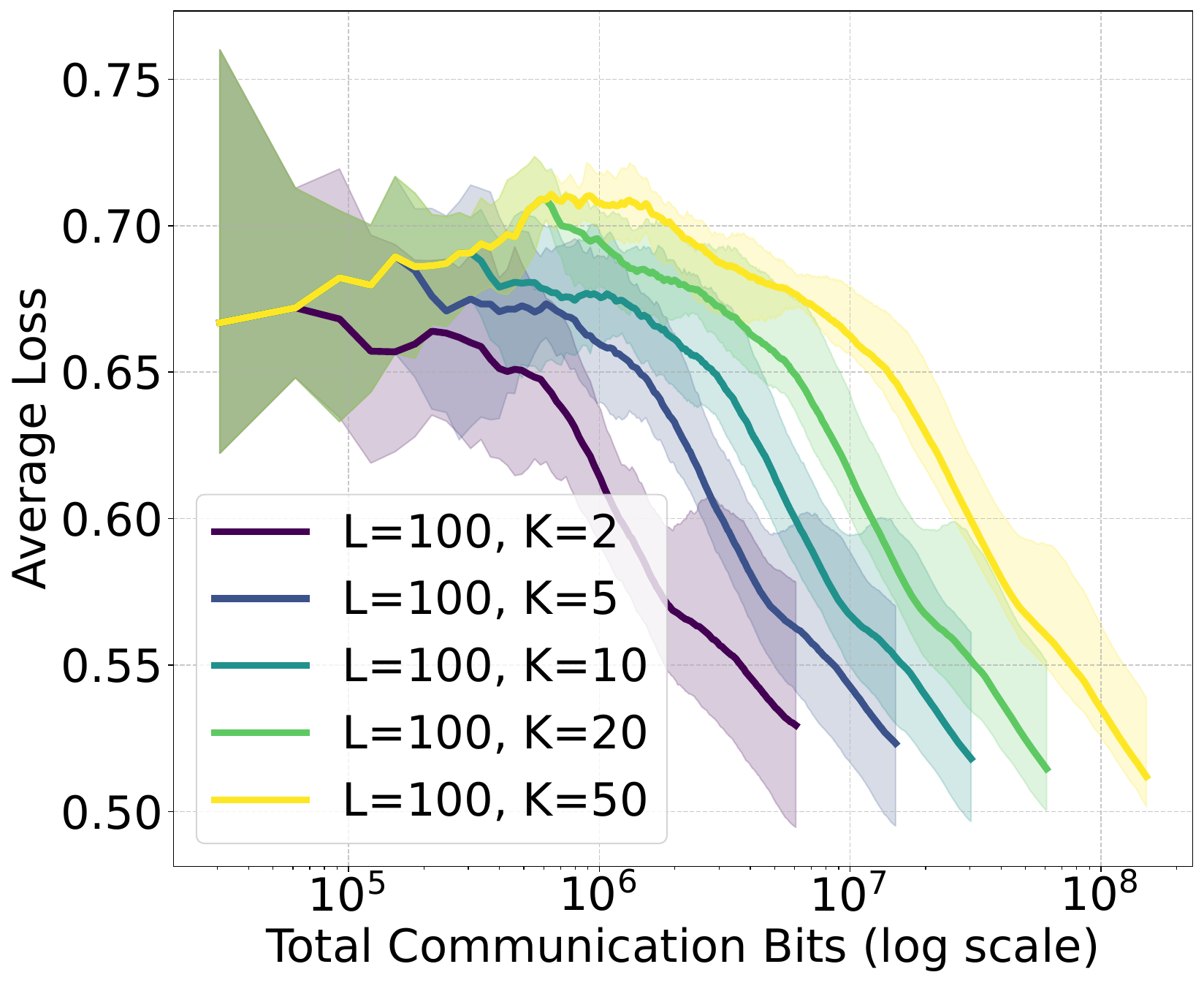}
        \caption{a9a: Loss vs Bits}
    \end{subfigure}
    \caption{Experimental results of CD-FTBL with varying communication rounds $K$ on a small random graph (9 nodes and 18 edges) with $\omega=0.1$ and fixed block size $L=100$.}
    \label{fig:sensitivity_K_w0.1}
\end{figure}

\begin{figure}[htbp]
    \centering
    \begin{subfigure}[b]{0.24\textwidth}
        \centering
        \includegraphics[width=\textwidth]{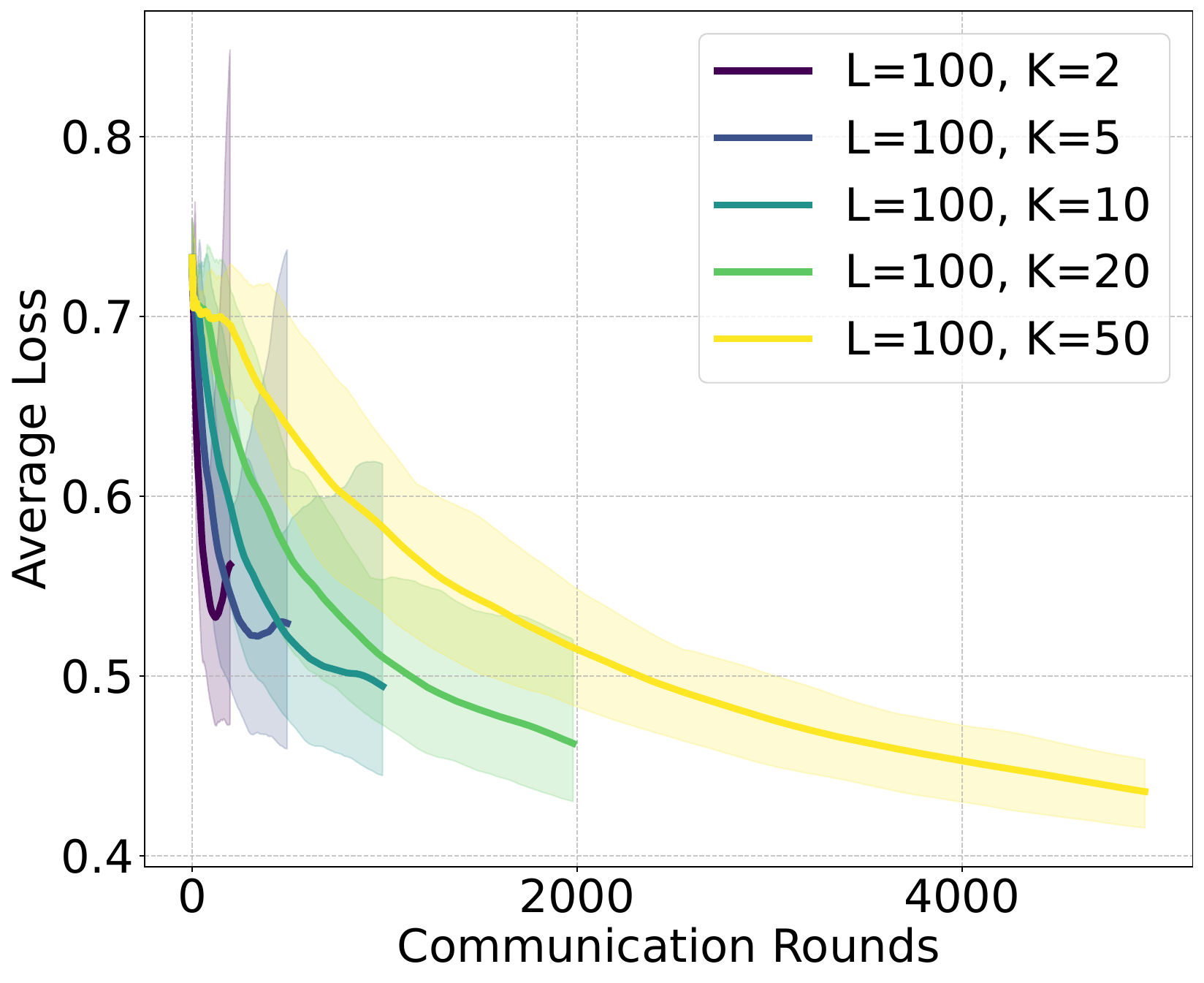}
        \caption{ijcnn1: Loss vs Rounds}
    \end{subfigure}
    \hfill
    \begin{subfigure}[b]{0.24\textwidth}
        \centering
        \includegraphics[width=\textwidth]{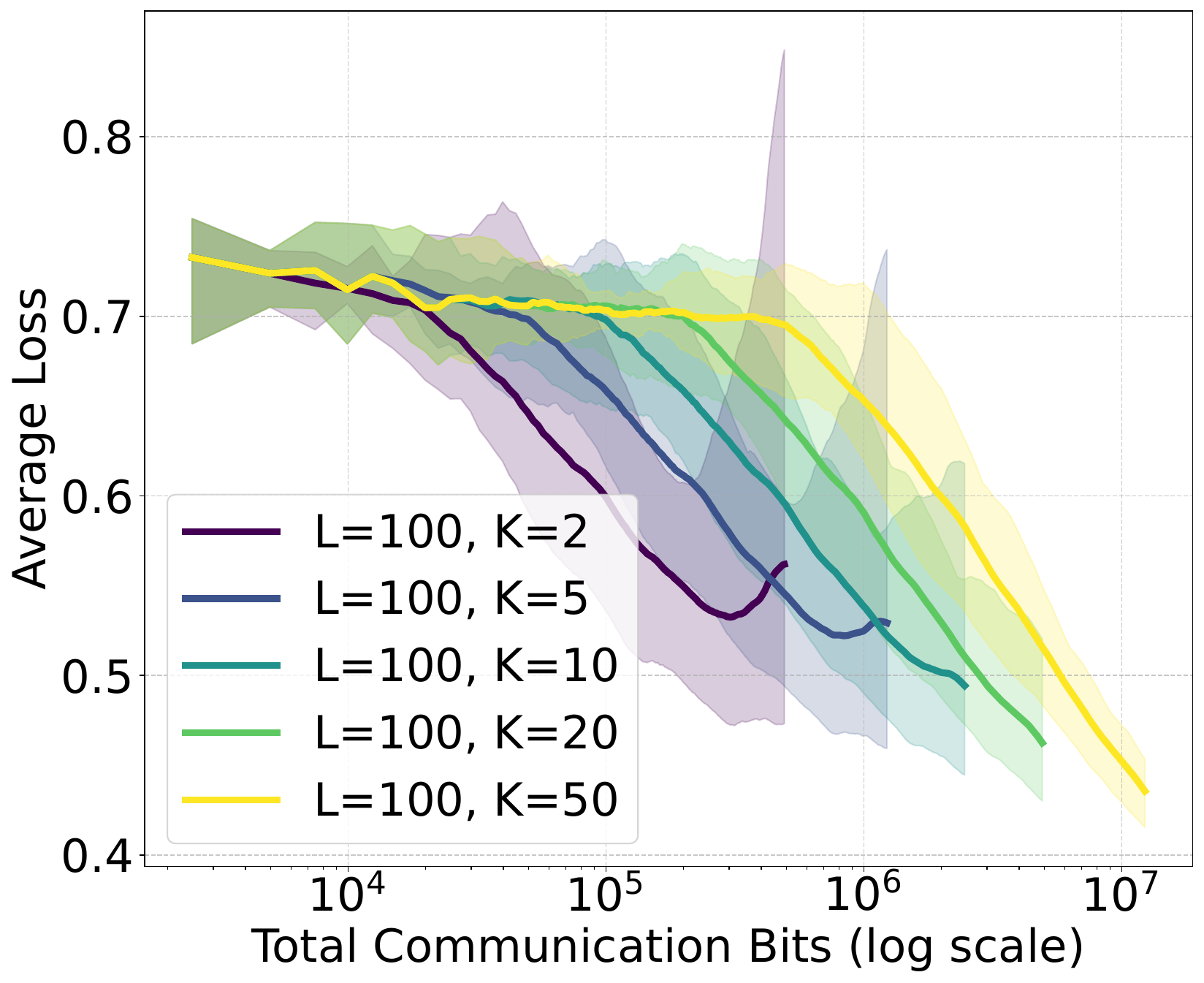}
        \caption{ijcnn1: Loss vs Bits}
    \end{subfigure}
    \hfill
    \begin{subfigure}[b]{0.24\textwidth}
        \centering
        \includegraphics[width=\textwidth]{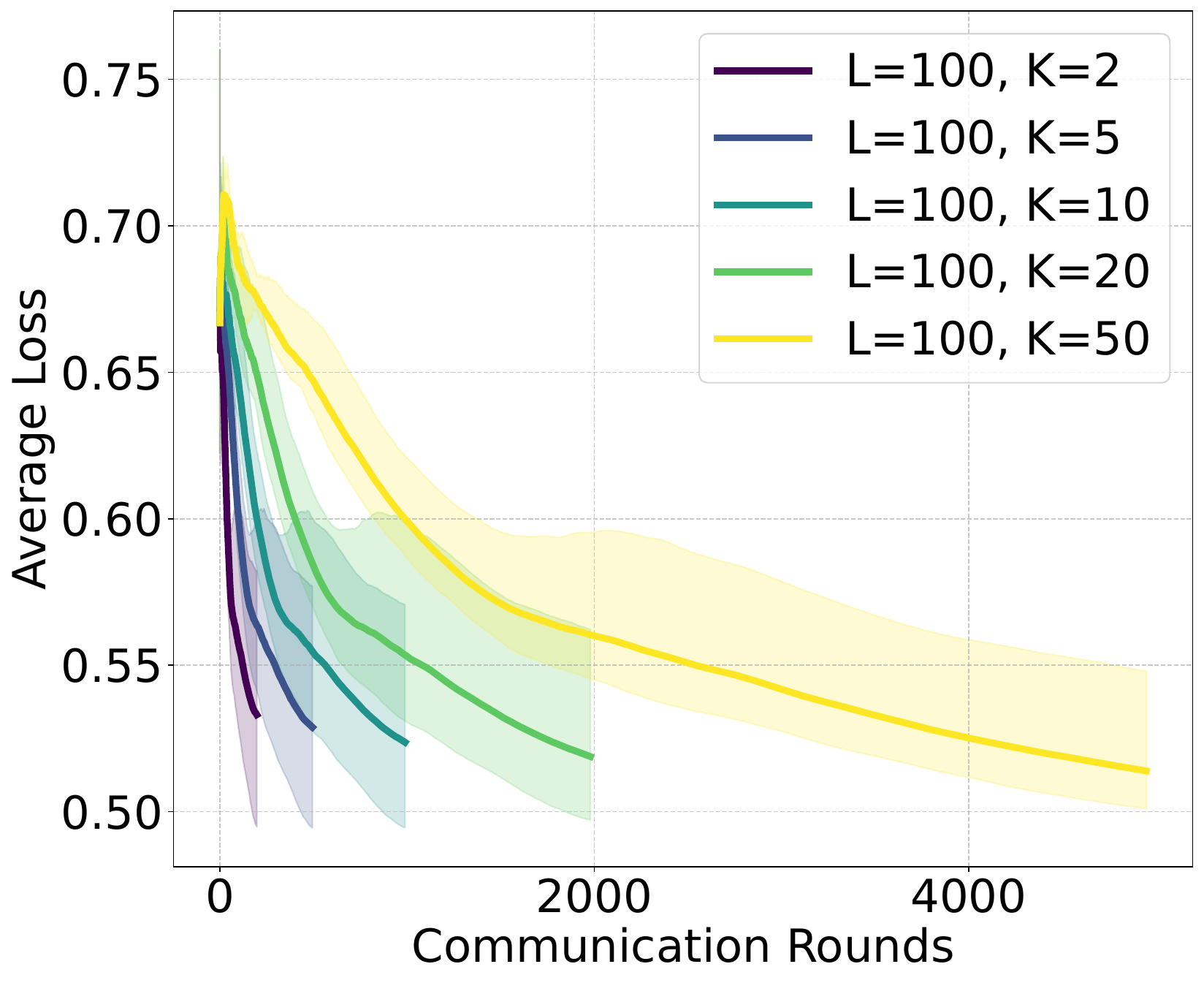}
        \caption{a9a: Loss vs Rounds}
    \end{subfigure}
    \hfill
    \begin{subfigure}[b]{0.24\textwidth}
        \centering
        \includegraphics[width=\textwidth]{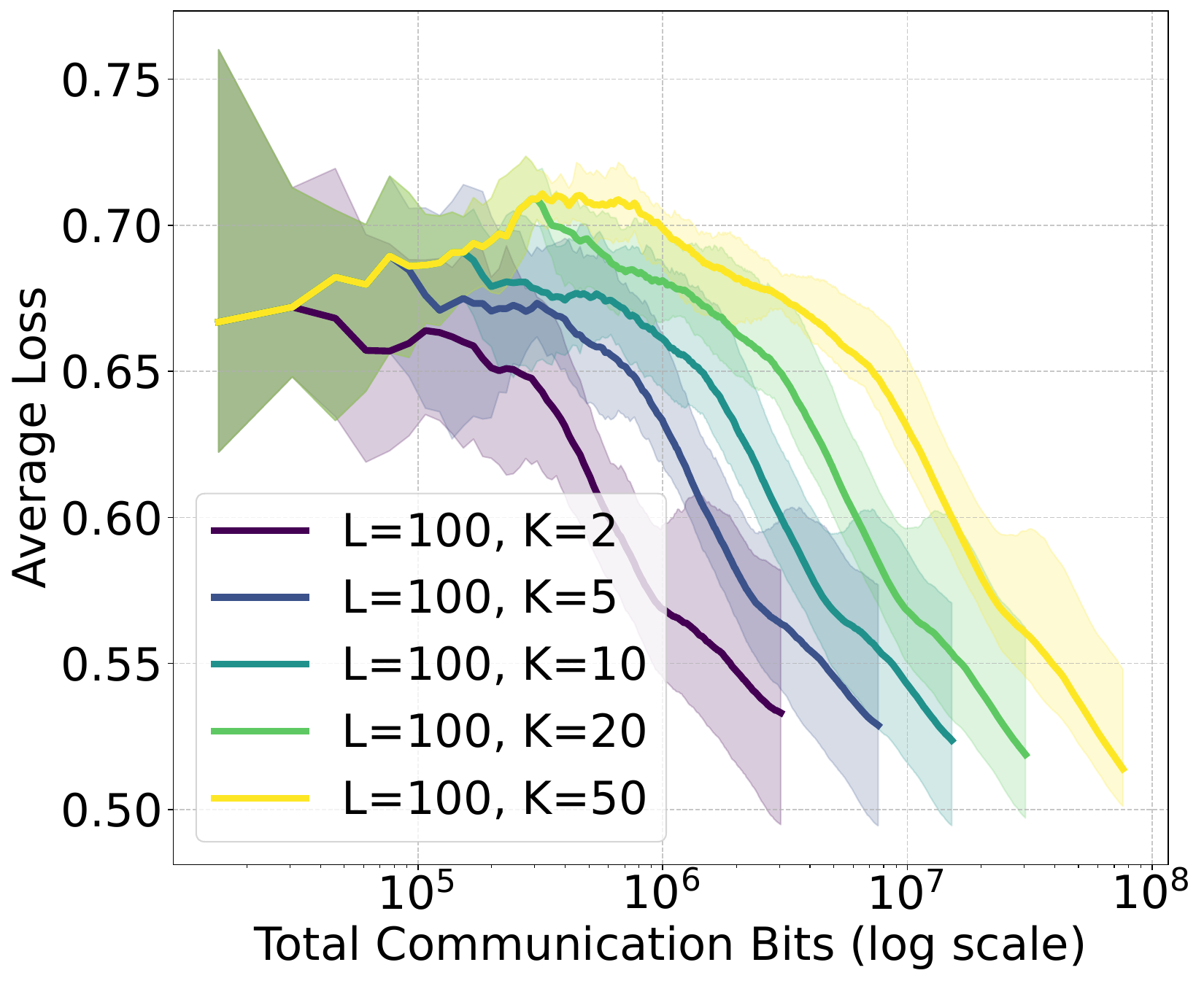}
        \caption{a9a: Loss vs Bits}
    \end{subfigure}
    \caption{Experimental results of CD-FTBL with varying communication rounds $K$ on a small random graph (9 nodes and 18 edges) with $\omega=0.05$ and fixed block size $L=100$.}
    \label{fig:sensitivity_K_w0.05}
\end{figure}

\section{Additional experimental results}
\label{app:experiments}

In this section, we provide additional experimental results with different network sizes, varying compression ratios, and the sensitivity of the communication frequency parameter. 

First, Figures~\ref{fig:n9_omega_0.5_results} and \ref{fig:n9_omega_0.05_results} present the convergence behavior on the small random graph ($9$ nodes and $18$ edges) under compression ratios of $\omega = 0.5$ and $\omega = 0.05$, respectively. To further investigate the scalability of our methods in larger networks, we conduct additional experiments on a larger random graph consisting of $50$ nodes and $100$ edges. The corresponding results for $\omega \in \{0.5, 0.1, 0.05\}$ are illustrated in Figures~\ref{fig:n50_omega_0.5_results}, \ref{fig:n50_omega_0.1_results} and \ref{fig:n50_omega_0.05_results}. Consistent with the main findings, these additional results further validate the superior efficacy and communication efficiency of our proposed algorithms across both full-information and bandit settings, regardless of network scales and compression levels.

Furthermore, to empirically investigate the trade-off between consensus accuracy and communication efficiency, we conduct a sensitivity analysis on the communication frequency $K$ for our CD-FTBL algorithm. Note that in the preceding experiments, the default number of communication rounds per block is set to $K = \lceil L/5 \rceil$. Specifically, we fix the block size at $L=100$ and vary the number of communication rounds per block $K$. The results under compression ratios of $\omega \in \{0.5, 0.1, 0.05\}$ are presented in Figures~\ref{fig:sensitivity_K_w0.5}, \ref{fig:sensitivity_K_w0.1}, and \ref{fig:sensitivity_K_w0.05}. As depicted in the figures, configurations with a smaller $K$ exhibit a significantly faster decay in average loss with respect to both communication rounds and transmitted bits. However, the final loss achieved by smaller $K$ values is slightly higher. This phenomenon highlights a clear trade-off: decreasing $K$ aggressively reduces communication overhead while slightly compromising the final accuracy. Nevertheless, this observation supports our theoretical analysis that a reduced $K$ is sufficient to control the consensus error within a reasonable tolerance.

\section{Proof of Theorem \ref{thm:thm1}}
Recall that $\bar{\d}(z)=(1/n)\sum_{i=1}^{n} \mathbf{\d}_i(z)$. For $z \in [T/L]$, we define the virtual leader sequence as
\begin{equation}
	\label{bar(x)argmin}
	\bar{\mathbf{x}}(z)
	=
	\argmin_{\mathbf{x} \in \mathcal{K}}
	\left\langle \mathbf{x},\bar{\mathbf{z}}(z)\right\rangle
	+\frac{(z-1)L\alpha +2h}{2}\|\mathbf{x}\|_2^2, \notag
\end{equation}
where $\bar{\z}(z)=\sum_{\tau=1}^{z-1}\bar{\d}(\tau)$. Following the analysis of \citet{wan2025optimal}, the total regret can be decomposed into the regret of the virtual leader and a deviation term. To be precise, for $z\in[T/L]$, let $\ell_z(\x)=\langle\bar{\d}(z),\x\rangle +(L\alpha/2)\|\x\|_2^2$ and $\x^\ast \in \min_{\x \in \K} \sum_{t=1}^T f_t(\x)$.
According to (41) and (42) in the proof of Theorem 1 of \citet{wan2025optimal}, we have 
\begin{equation}
\label{eq1-by-wan}
\begin{split}
R_{T,i}\leq& n\sum_{z=1}^{T/L}\left(\ell_z(\bar{\x}(z+1))-\ell_z(\x^\ast)
\right) + nG\sum_{z=1}^{T/L}\left(L\Delta_i(z)+2\sum_{j=1}^n\Delta_j(z)\right),
\end{split}
\end{equation}
where $\Delta_i(z)=\|\x_i(z)-\bar\x(z+1)\|_2$.

To analyze the regret of the virtual leader, we introduce a classical lemma.
\begin{lemma}[Lemma 6.6 in \citet{Garber16}]
	\label{lem:l_t(x)Garber&Hazan}
	Let $\{\ell_t(\mathbf{x})\}_{t=1}^T$ be a sequence of convex functions and $\mathbf{x}_t^* \in \argmin_{\mathbf{x} \in \mathcal{K}} \sum_{\tau=1}^t \ell_\tau(\mathbf{x})$ for any $t \in [T]$. Then, it holds that $\sum_{t=1}^T \ell_t(\mathbf{x}_t^*) - \min_{\mathbf{x} \in \mathcal{K}} \sum_{t=1}^T \ell_t(\mathbf{x}) \leq 0.$
\end{lemma}
By further defining $\ell_0(\x)=h\|\x\|_2^2$ and then applying Lemma \ref{lem:l_t(x)Garber&Hazan}, it is easy to verify that the first term in the right side of \eqref{eq1-by-wan} can be bounded by $nhR^2$. By combining this result and taking expectation over two sides of \eqref{eq1-by-wan}, we have
\begin{equation}
	\label{eq:regret_decomp}
	\begin{aligned}
		\E[R_{T, i}] 
		\leq &nG\sum_{z=1}^{T/L}\left(L\E[\Delta_i(z)]+2\sum_{j=1}^n\E[\Delta_j(z)]\right) +n h R^2.
	\end{aligned}
\end{equation}
Then, we proceed to analyze $\E[\Delta_i(z)]$, which now is affected by compressed communication. However, similar to analysis of \citet{wan2025optimal}, for any $z\geq 2$, we still split $\E[\Delta_i(z)]$ into a consensus  error term and a leader stability term, i.e.,
\begin{equation}
	\label{eq:delta_split}
	\begin{aligned}
		 \E[\Delta_i(z)]
		\le& \E \left[\|\mathbf{x}_i(z) - \bar{\mathbf{x}}(z-1)\|_2\right] + \E[\|\bar{\mathbf{x}}(z-1) - \bar{\mathbf{x}}(z+1)\|_2].
	\end{aligned}
\end{equation}
To bound the consensus  error in \eqref{eq:delta_split}, we introduce two useful lemmas, where the latter one is derived by using Lemma \ref{lem_ezl_ez0_maintext}.

\begin{lemma}[Lemma 5 in \citet{Duchi2011}]
	\label{lem:projection G-lipschitz}
	Let $\Pi_{\mathcal{K}}(\mathbf{u}, \eta) = \argmin_{\mathbf{x} \in \mathcal{K}} \left\langle \mathbf{u}, \mathbf{x} \right\rangle + \frac{1}{\eta} \|\mathbf{x}\|_2^2$. For any $\mathbf{u}, \mathbf{v} \in \mathbb{R}^d$, we have $\|\Pi_{\mathcal{K}}(\mathbf{u}, \eta) - \Pi_{\mathcal{K}}(\mathbf{v}, \eta)\|_2 \leq \frac{\eta}{2} \|\mathbf{u} - \mathbf{v}\|_2$.
\end{lemma}

\begin{lemma}
	\label{lem_z_and_zbar}
	Suppose Assumption \ref{doubly stochastic}, \ref{assum_compress} and \ref{G-lipschitz} hold. Consider Algorithm \ref{CD-FTGL} and \ref{CD-FTBL} with the consensus step size $\gamma$ and the communication rounds $K$ in Theorem \ref{thm:thm1} and \ref{thm:thm2}, then for all $z \in [T/L]$, it holds that
	\begin{equation}
		\begin{aligned}
			\mathbb{E} \left[\left\|\z_i(z)-\bar{\z}(z) \right\|_2\right] < \frac{2\E[\left\|D(z-1)\right\|_F]}{\sqrt{n}}, \notag
		\end{aligned}
	\end{equation}
	where $D(z) = [\d_1(z), \dots, \d_n(z)] \in \mathbb{R}^{d \times n}$  for any $z\geq 1$ and $D(0)=\ze^{d\times n}$.
\end{lemma}

Under Assumptions \ref{r<K<R} and \ref{G-lipschitz}, for any $z \in [T/L]$, we have
\begin{equation}
	\label{eq:bound_of_D_z}
	\begin{aligned}
        \|D(z)\|_F = \sqrt{\sum_{i=1}^n \|\d_i(z)\|_2^2} 
        =\sqrt{\sum_{i=1}^n \left\| \sum_{t \in \mathcal{T}_z}\left(\nabla f_{t,i}(\x_i(t)) - \alpha \x_i(t)\right) \right\|_2^2} 
        \leq \sqrt{n} L (G + \alpha R). 
	\end{aligned}
\end{equation}
By combining Lemma~\ref{lem_z_and_zbar} with \eqref{eq:bound_of_D_z}, we have
\begin{equation}
	\mathbb{E} \left[\left\|\mathbf{z}_i(z) - \bar{\mathbf{z}}(z)\right\|_2\right] < 2 L\left(G + \alpha R\right) \notag
\end{equation}
for any $z \in [T/L]$. Then, by applying Lemma~\ref{lem:projection G-lipschitz} and the above inequality, for any $z\geq 2$, we have
\begin{equation}
	\label{eq:local estimation error bound}
	\begin{aligned}
		\E \left[\left\|\mathbf{x}_i(z) - \bar{\mathbf{x}}(z-1)\right\|_2\right] \le& \frac{\E \left[\left\|\mathbf{z}_i(z-1) - \bar{\mathbf{z}}(z-1)\right\|_2\right]}{(z-2)L\alpha + 2h} 
		\le \frac{2 L(G + \alpha R)}{(z-2)L\alpha + 2h} .
	\end{aligned}
\end{equation}

For the leader stability in \eqref{eq:delta_split}, according to (38) in the proof of Theorem 1 in \citet{wan2025optimal}, for any $z\geq 2$, we have
\begin{equation}
\label{eq2-by-wan}
	\|\bar{\mathbf{x}}(z-1) - \bar{\mathbf{x}}(z+1)\|_2 
	\le \frac{4L(G + 2\alpha R)}{z L \alpha + 2h}. 
\end{equation}

Finally, note that $\Delta_i(1)=\|\x_i(1)-\bar\x(2)\|_2\leq (2L(G + 2\alpha R))/(L\alpha +2h)$, which is also due to (38) in the proof of Theorem 1 in \citet{wan2025optimal}. Therefore, this proof can be completed by substituting \eqref{eq:local estimation error bound} and \eqref{eq2-by-wan} into \eqref{eq:delta_split}, and combining with \eqref{eq:regret_decomp} and the upper bound of $\Delta_i(1)$.

\section{Proof of Theorem \ref{thm:thm2}}
For $ z \in [T/L] $, we define the virtual leader sequence as
\begin{align}
  \bar\x(z+1)
  =
  \argmin_{\x\in\mathcal{K}_{\epsilon}}
   \left\langle \x,\bar\z(z)\right\rangle +\frac{(z-1)L\alpha +2h}{2}\|\x\|_2^2 \notag
\end{align}
where $\bar{\z}(z)=\sum_{\tau=1}^{z-1}\bar{\d}(\tau)$, but $\bar\d(z)$ is defined as
\begin{align}
  \bar\d(z)
  &=\frac{1}{n}\sum_{i=1}^n \d_i(z)
  =\frac{1}{n}\sum_{i=1}^n\sum_{t\in\mathcal{T}_z}
  \left(\g_i(t)-\alpha\x_i(z)\right). \notag
\end{align}
% where $\g_i(t)$ is the unbiased gradient estimate.
Moreover, let $\tilde{\x}^* = (1 - \epsilon / r)\x^* \in \mathcal{K}_\epsilon$. Following the proof of Theorems 5 and 6 in \citet{Wan-22-JMLR}, the total regret can be bounded as
\begin{equation}
\label{eq:bandit_decomp_step1}
\E[R_{T, i}] 
\leq \underbrace{\E\left[\sum_{z=1}^{T/L} \sum_{t \in \mathcal{T}_z} \sum_{j=1}^n \left( \hat{f}_{t,j,\epsilon}(\mathbf{x}_i(z)) - \hat{f}_{t,j,\epsilon}(\tilde{\x}^*) \right)\right]}_{:=\hat{R}_{T,i}}   + 3\epsilon n G T + \frac{\epsilon n G R T}{r},
\end{equation}
where $\hat{f}_{t,j,\epsilon}(\cdot)$ is the $\epsilon$-smoothed version of $f_{t,j}(\cdot)$. 

Note that according to Lemma 2.6 in \citet{Hazan2016}, over the shrunk set $\K_\epsilon$, the $\epsilon$-smoothed function $\hat{f}_{t,j,\epsilon}(\cdot)$ shares the same strong convexity and Lipschitz property as $f_{t,j}(\cdot)$. Therefore, following \eqref{eq1-by-wan} and using Lemma \ref{lem:epsilon-smooth}, it is not hard to verify that
\begin{equation}
\label{eq:bandit_decomp_step2}
\hat{R}_{T,i}
 \leq \underbrace{n \sum_{z=1}^{T/L} \E\left[ \ell_z(\bar{\mathbf{x}}(z+1)) - \ell_z(\tilde{\mathbf{x}}^*) \right]}_{:=\bar{R}_z} + nG\sum_{z=1}^{T/L}\left(L\E[\Delta_i(z)]+2\sum_{j=1}^n\E[\Delta_j(z)]\right),
\end{equation}
where $\ell_z(\mathbf{x}) = \langle \mathbf{x}, \bar{\mathbf{d}}(z) \rangle + (L\alpha/2)\|\mathbf{x}\|_2^2$ and $\Delta_i(z)=\|\x_i(z)-\bar\x(z+1)\|_2$.

Unlike the full-information setting, we need to introduce a new lemma to bound $\mathbb{E}[\|\bar{\d}(z)\|_2^2]$.
\begin{lemma}[Derived from Lemma 6 in \citet{Wan-22-JMLR}]
\label{lem6-wan}
Under Assumptions \ref{r<K<R}, \ref{G-lipschitz}, and \ref{bound:|f|<M}, for any \(i \in [n]\) and \(z \in [T/L]\), let $\bar{\d}(z) = \frac{1}{n}\sum_{i=1}^n\d_i(z)$, Algorithm \ref{CD-FTBL} ensures
\begin{equation}
\mathbb{E}[\|\d_i(z)\|_2^2]
\le
2 L\frac{d^2M^2}{\epsilon^2}
+2L^2G^2
+2\alpha^2L^2R^2.\notag
\end{equation}
\end{lemma}
For brevity, let
\begin{align}
    B_1 := 2\frac{\sqrt{L}dM}{\epsilon}+2 L G+3\alpha LR, 
    \qquad B_2 := \sqrt{2L\frac{d^2M^2}{\epsilon^2} + 2L^2G^2 + 2\alpha^2L^2R^2}. \notag 
\end{align}
Due to Lemma \ref{lem6-wan} and the definition of $\bar{\d}(z)$, we have
\begin{equation*}
\begin{aligned}
\mathbb{E}\left[ \| \bar{\d}(z) \|_2^2 \right] 
\leq \frac{1}{n} \sum_{i=1}^n \mathbb{E}[ \| \d_i(z) \|_2^2 ] \leq B_2.
\end{aligned}
\end{equation*}
By combining the above inequality with Jensen's inequality, we have
\begin{equation}
\E[\|\nabla \ell_z(\mathbf{x})\|_2] = \E[\left\|\bar\d(z) + L\alpha\x\right\|_2]
  \le \sqrt{\E\left[\|\bar{\d}(z)\|_2^2\right]} + L\alpha R
  \le B_1. \notag
\end{equation}
Then, to bound the first term in the right side of \eqref{eq:bandit_decomp_step2}, we notice that
\begin{equation}
\label{eq:bandit_decomp_step3}
\begin{split}
\bar{R}_z
\leq &n\sum_{z=1}^{T/L} \E[ \ell_z(\bar{\mathbf{x}}(z+2)) - \ell_z(\tilde{\mathbf{x}}^*) ] 
+ nB_1\sum_{z=1}^{T/L}\E[\|\bar{\mathbf{x}}(z+2)-\bar{\mathbf{x}}(z+1)\|_2]\\
\leq &nhR^2+ nB_1\sum_{z=1}^{T/L}\E[\|\bar{\mathbf{x}}(z+2)-\bar{\mathbf{x}}(z+1)\|_2],
\end{split}
\end{equation}
where the second inequality is derived by using Lemma \ref{lem:l_t(x)Garber&Hazan}.

% The virtual regret is bounded by $hR^2$. The stability cost is bounded with the gradient bound $B_1$ as follows.
% \begin{align}
% \label{eq:stability_cost}
% & \E[\ell_z(\bar{\mathbf{x}}(z+1)) - \ell_z(\bar{\mathbf{x}}(z+2))] \notag \\
% &\leq \E\left[\|\nabla \ell_z(\bar{\mathbf{x}}(z+1))\|_2\right] \cdot \|\bar{\mathbf{x}}(z+1) - \bar{\mathbf{x}}(z+2)\|_2 \notag \\
% &\le B_1 \E\left[\left\|\bar{\mathbf{x}}(z+1) - \bar{\mathbf{x}}(z+2)\right\|_2\right].
% \end{align}
Following (38) in the proof of Theorem 1 in \citet{wan2025optimal}, for any $z\geq 1$, we have
\begin{equation}
\label{eq:stability_bound}
\E\left[\left\|\bar{\mathbf{x}}(z+1) - \bar{\mathbf{x}}(z+2)\right\|_2\right] \le \frac{2B_1}{zL\alpha + 2h}.
\end{equation}

Moreover, similar to the full-information setting, for any $z\geq 2$, we split $\E[\Delta_i(z)]$ into a consensus error term and a leader stability term as follows
\begin{equation}
\label{deviation_decomp}
\begin{aligned}
\E[\Delta_i(z)]\leq\E\left[\left\|\x_i(z)-\bar\x(z)\right\|_2\right] + \left\|\bar\x(z)-\bar\x(z+1)\right\|_2.
\end{aligned}
\end{equation}
By applying Lemmas~\ref{lem:projection G-lipschitz} and \ref{lem_z_and_zbar}, for any $z\geq 2$, we have
\begin{align}
\label{eq:deviation_error_1}
\E\left[\left\|\x_i(z)-\bar\x(z)\right\|_2\right]
\le \frac{2 B_2}{(z-2)L\alpha+2h}.
\end{align}
By substituting \eqref{eq:deviation_error_1} and \eqref{eq:stability_bound} into \eqref{deviation_decomp}, for any $z\geq 2$, we have
\begin{equation}
\label{deviation_decomp-wan}
\begin{aligned}
\E[\Delta_i(z)]\leq\frac{2 B_2}{(z-2)L\alpha+2h} + \frac{2 B_1}{(z-1) L \alpha + 2h}.
\end{aligned}
\end{equation}
% Similarly to \eqref{eq:stability_bound}, the second term is bounded by
% \begin{equation}
% \label{eq:deviation_error_2}
% \left\|\bar{\mathbf{x}}(z) - \bar{\mathbf{x}}(z+1)\right\|_2 \le \frac{2 B_1}{(z-1) L \alpha + 2h}.
% \end{equation}
Finally, this proof can be completed by  substituting \eqref{eq:stability_bound} into \eqref{eq:bandit_decomp_step3}, and then combining with \eqref{eq:bandit_decomp_step1}, \eqref{eq:bandit_decomp_step2}, \eqref{deviation_decomp-wan} and $\Delta_i(1)=\|\x_i(1)-\bar\x(2)\|_2=0$.
\section{Proof of Lemma \ref{lem_ezl_ez0_maintext}}
\label{Proof of Lemma lem_ezl_ez0_maintext}

To facilitate the analysis, we first rewrite the error definition in matrix notation. Let $Z^k = [\mathbf{z}_1^k, \dots, \mathbf{z}_n^k] \in \mathbb{R}^{d \times n}$ and $\hat{Z}^k = [\hat{\mathbf{z}}_1^k, \dots, \hat{\mathbf{z}}_n^k] \in \mathbb{R}^{d \times n}$ be the concatenated matrices of local variables and their replicas at iteration $k$. The average is denoted by $\bar{Z}^k = Z^k \mathcal{M}$, where $\mathcal{M} = \frac{1}{n} \mathbf{1}_n \mathbf{1}_n^\top$.
Consequently, the total error $e^k$ defined in Lemma \ref{lem_ezl_ez0_maintext} can be equivalently written using the Frobenius norm, which is
\begin{equation}
    e^k = \mathbb{E} [ \| Z^k - \bar{Z}^k \|_F^2 ] + \mathbb{E} [ \| Z^k - \hat{Z}^k \|_F^2 ]. \notag
\end{equation}

The proof relies on establishing a recursive bound for these consensus and compression errors. We first state a lemma that characterizes the one-step error dynamics.

\begin{lemma}
\label{lem:consensus_and_compress_maintext}
Suppose Assumptions \ref{doubly stochastic}, \ref{assum_compress}, and \ref{G-lipschitz} hold, and let $\gamma \in (0, 1]$. 
Let $E_1(k) := \mathbb{E}_Q [\| Z^k - \bar{Z}^k \|_F^2]$ and $E_2(k) := \mathbb{E}_Q [\| Z^k - \hat{Z}^k \|_F^2]$. 
For any iteration $k \geq 0$, the errors satisfy:
\begin{align}
E_1(k+1) 
&\leq \left(1 - \frac{\gamma \rho}{2} \right)^2 E_1(k) + \left(\gamma + \frac{2\gamma}{\rho}\right) (1 - \omega) \beta^2 E_2(k), \label{eq:rec_1_main} \\
E_2(k+1) 
&\leq \left( 1 + \frac{2}{\omega} \right) \gamma^2 \beta^2 E_1(k) + \left( 1 - \frac{\omega}{2} \right)(1 + 8 \gamma) E_2(k). \label{eq:rec_2_main}
\end{align}
\end{lemma}

Recall the total error $e^k = E_1(k) + E_2(k)$. Summing \eqref{eq:rec_1_main} and \eqref{eq:rec_2_main}, we obtain the linear recurrence, i.e.,
\begin{equation}
e^{k+1} \le \eta(\gamma) E_1(k) + \xi(\gamma) E_2(k) \le \max\{\eta(\gamma), \xi(\gamma)\} \cdot e^k, \notag
\end{equation}
where the coefficients are defined as
\begin{align}
\eta(\gamma) &:= \left(1 - \frac{\gamma \rho}{2} \right)^2 + \left(1 + \frac{2}{\omega} \right) \gamma^2 \beta^2,  \notag\\
\xi(\gamma) &:= \left( \gamma + \frac{2\gamma}{\rho} \right) (1 - \omega) \beta^2 + \left(1 - \frac{\omega}{2} \right)(1 + 8\gamma).  \notag
\end{align}
It is not hard to establish a valid upper bound for $\max\{\eta(\gamma), \xi(\gamma)\}$ by following the proof of Theorem 2 in \citet{ICML-Gossip}. For completeness, we provide the detailed proof here.

First, we observe that $\eta(\gamma)$ is a convex quadratic function with respect to $\gamma$. Its minimum occurs at
$\gamma' = \frac{2\rho\omega}{4(\omega+2)\beta^2 + \rho^2\omega}$, with value $\eta(\gamma') = \frac{4(\omega+2)\beta^2}{4(\omega+2)\beta^2 + \rho^2\omega} < 1$.
Thus, by Jensen's inequality, for any $\lambda \in [0, 1]$, we have
\begin{equation}
\label{eq:jensen_general}
\begin{aligned}
\eta(\lambda \gamma') 
&\le (1-\lambda)\eta(0) + \lambda \eta(\gamma') 
= 1 - \lambda (1 - \eta(\gamma')) 
= 1 - \lambda \frac{\rho^2 \omega}{4(\omega+2)\beta^2 + \rho^2\omega}.
\end{aligned}
\end{equation}
For the specific choice
$\lambda' = \frac{4(\omega+2)\beta^2 + \rho^2\omega}{4(1-\omega)(\rho+2)\beta^2 + 32\rho - 16\rho\omega + 2\rho^2}$,
we obtain the step size $\gamma^* = \lambda' \gamma'$ as given in \eqref{gamma-value}
\begin{equation}
\gamma^* = \frac{\rho\omega}{2(1-\omega)(\rho+2)\beta^{2}+16\rho-8\rho\omega+\rho^{2}}. \notag
\end{equation}
Substituting this $\lambda'$ back into the inequality \eqref{eq:jensen_general}, we verify that
\begin{equation}
\eta(\gamma^*) \le 1 - \frac{\rho^2 \omega}{4(1-\omega)(\rho+2)\beta^{2}+32\rho-16\rho\omega+2\rho^{2}} = 1 - \frac{\gamma^* \rho}{2}. \notag
\end{equation}
Similarly, substituting $\gamma^*$ into  $\xi(\gamma)$, it holds that
\begin{equation}
\xi(\gamma^*) = \left(\gamma^* + \frac{2\gamma^*}{\rho}\right)(1-\omega)\beta^2 + \left(1-\frac{\omega}{2}\right)(1+8\gamma^*) = 1 - \frac{\gamma^* \rho}{2}. \notag
\end{equation}
Thus, we conclude that $\max\{\eta(\gamma^*), \xi(\gamma^*)\} \le 1 - \frac{\gamma^* \rho}{2}$ and the contraction factor is bounded by $\theta := 1 - \frac{\gamma^* \rho}{2}$. Applying the recursion iteratively yields
\begin{equation}
e^k \le \theta e^{k-1} \le \dots \le \theta^k e^0. \notag
\end{equation}
This completes the proof of Lemma \ref{lem_ezl_ez0_maintext}.

\section{Proof of Lemma \ref{lem:consensus_and_compress_maintext}}
\label{Proof of Lemma lem:consensus_and_compress_maintext}
We analyze the evolution of the consensus and compression errors using the matrix-form updates.
Since $A$ is doubly stochastic, we have $\bar{Z}^k = Z^k\mathcal{M}$ and $\bar{Z}^kA = \bar{Z}^k$.
The update rules in Algorithm~\ref{alg:choco} imply the following dynamics, i.e.,
\begin{align}
    \hat{Z}^{k+1} &= \hat{Z}^k + Q\left(Z^k - \hat{Z}^k\right), \label{eq:update_Z_hat} \\
    Z^{k+1} &= Z^k + \gamma \hat{Z}^{k+1} (A-I) = Z^k - \gamma \hat{Z}^{k+1}\mathcal{L}. \label{eq:update_Z}
\end{align}
Assumption \ref{assum_compress} combined with \eqref{eq:update_Z_hat} yields the contraction property of
\begin{equation}
\label{eq:compression_contraction}
    \mathbb{E}_Q \left[\left\| Z^k - \hat{Z}^{k+1} \right\|_F^2\right] \leq (1 - \omega) \left\| Z^k - \hat{Z}^k \right\|_F^2.
\end{equation}

\paragraph{Part 1: Consensus error analysis.}
Recall that $\bar{Z}^k\mathcal{L} = 0$ and $\mathcal{M}\mathcal{L} = 0$. Multiplying \eqref{eq:update_Z} by $(I-\mathcal{M})$, we isolate the consensus deviation by
\begin{equation}
\begin{aligned}
    Z^{k+1} - \bar{Z}^{k+1} &=  \left( Z^k - \gamma  \hat{Z}^{k+1}\mathcal{L} \right)(I - \mathcal{M}) \\
    &= (Z^k - \bar{Z}^k)(I - \gamma \mathcal{L}) - \gamma (\hat{Z}^{k+1} - Z^k)\mathcal{L}. \notag
\end{aligned}
\end{equation}
Using Young's inequality $\|X+Y\|_F^2 \le (1+\alpha)\|X\|_F^2 + (1+\alpha^{-1})\|Y\|_F^2$ with $\alpha = \frac{\gamma\rho}{2}$, we have
\begin{equation}
\label{eq:consensus_bound_raw}
\begin{aligned}
    \| Z^{k+1} - \bar{Z}^{k+1} \|_F^2 \le \left( 1 + \frac{\gamma \rho}{2} \right) \underbrace{\| (Z^k - \bar{Z}^k) (I - \gamma \mathcal{L})\|_F^2}_{T_1} + \left( 1 + \frac{2}{\gamma \rho} \right) \underbrace{\| \gamma  (\hat{Z}^{k+1} - Z^k)\mathcal{L} \|_F^2}_{T_2}.
\end{aligned}
\end{equation}
For $T_1$, utilizing the property $(Z^k - \bar{Z}^k)\mathcal{M} = 0$ and the spectral bound $\|I - \gamma\mathcal{L} - \mathcal{M}\|_2 \le 1 - \gamma\rho$ for $\gamma\rho \le 1$, we obtain that
\begin{equation}
    T_1 = \| (Z^k - \bar{Z}^k)(I - \gamma \mathcal{L} - \mathcal{M}) \|_F^2 \le (1 - \gamma\rho)^2 \| Z^k - \bar{Z}^k \|_F^2. \notag
\end{equation}
For $T_2$, using $\|\mathcal{L}\|_2 = \beta$, we have $\mathbb{E}_Q[T_2] \le \gamma^2 \beta^2 (1-\omega) \E_Q [\| Z^k - \hat{Z}^k \|_F^2]$.
Let $E_1(k) := \mathbb{E}_Q [\| Z^k - \bar{Z}^k \|_F^2]$ and $E_2(k) := \mathbb{E}_Q [\| Z^k - \hat{Z}^k \|_F^2]$.
Substituting these bounds into \eqref{eq:consensus_bound_raw} and observing that $(1+\frac{\gamma\rho}{2})(1-\gamma\rho)^2 \le (1-\frac{\gamma\rho}{2})^2$ for $\gamma\rho \le 1$, we derive that
\begin{equation}
    E_1(k+1) \le \left(1 - \frac{\gamma \rho}{2} \right)^2 E_1(k) + \left(\gamma + \frac{2\gamma}{\rho}\right) (1 - \omega) \beta^2 E_2(k). \notag
\end{equation}

\paragraph{Part 2: Compression error analysis.}
Rearranging \eqref{eq:update_Z} allows us to express the compression update as
\begin{equation}
\begin{aligned}
    Z^{k+1} - \hat{Z}^{k+1} &= Z^k - \gamma \hat{Z}^{k+1} \mathcal{L} - \hat{Z}^{k+1} = (Z^k - \hat{Z}^{k+1})(I + \gamma \mathcal{L}) - \gamma (Z^k - \bar{Z}^k)\mathcal{L}.\notag
\end{aligned}
\end{equation}
By using Young's inequality again with $\alpha = \frac{\omega}{2}$, we have
\begin{equation}
    \| Z^{k+1} - \hat{Z}^{k+1} \|_F^2 \le \left( 1 + \frac{\omega}{2} \right) \| (Z^k - \hat{Z}^{k+1})(I + \gamma \mathcal{L}) \|_F^2 + \left( 1 + \frac{2}{\omega} \right) \| \gamma (Z^k - \bar{Z}^k)\mathcal{L} \|_F^2. \notag
\end{equation}
By applying \eqref{eq:compression_contraction} and the above inequality, we obtain
\begin{equation}
\begin{aligned}
    E_2(k+1) &\le \left( 1 + \frac{\omega}{2} \right) \|I+\gamma\mathcal{L}\|_2^2 (1-\omega) E_2(k) + \left( 1 + \frac{2}{\omega} \right) \gamma^2 \beta^2 E_1(k). \notag
\end{aligned}
\end{equation}
Using $\|I+\gamma\mathcal{L}\|_2 \le 1+\gamma\beta$ and the inequality $(1+\frac{\omega}{2})(1-\omega)(1+\gamma\beta)^2 \le (1-\frac{\omega}{2})(1+8\gamma)$ for $\beta \le 2, \gamma \le 1$, we conclude
\begin{equation}
    E_2(k+1) \le \left( 1 + \frac{2}{\omega} \right) \gamma^2 \beta^2 E_1(k) + \left( 1 - \frac{\omega}{2} \right)(1 + 8 \gamma) E_2(k). \notag
\end{equation}
This completes the proof of Lemma~\ref{lem:consensus_and_compress_maintext}.

\section{Proof of Lemma \ref{lem_z_and_zbar}}
\label{Proof of Lemma lem_z_and_zbar}

The proof relies on the recursive error bound established in Lemma \ref{lem_ezl_ez0_maintext}. We proceed the proof in three steps: analyzing the error propagation, determining the number of iterations $K$, and deriving the final bound.

\paragraph{Step 1: Error propagation analysis.}
At the beginning of block $z$ (i.e., $k=0$), the states are initialized based on the output of the previous block and the gradient update
\begin{equation}
    \z_i^0(z) = \z_i^K(z-1) + \d_i(z-1), \quad \hat{\z}_i^0(z) = \hat{\z}_i^K(z-1). \notag
\end{equation}
Consequently, the initial error $e_z^0$ is bounded by the error at the end of the previous block, $e_{z-1}^K$. Applying the inequality $\|a+b\|_2^2 \le 2\|a\|_2^2 + 2\|b\|_2^2$ and the variance decomposition, we derive
\begin{equation}
\label{eq:e_z_0_bound}
\begin{aligned}
    e_z^0 &= \mathbb{E} \sum_{i=1}^n \left( \|\z_i^0(z) - \bar{\z}^0(z)\|_2^2 + \|\z_i^0(z) - \hat{\z}_i^0(z)\|_2^2 \right) \\
    &\le 2 e_{z-1}^K + 2 \sum_{i=1}^n \left( \|\d_i(z-1) - \bar{\d}(z-1)\|_2^2 + \|\d_i(z-1)\|_2^2 \right) \\
    &\le 2 e_{z-1}^K + 4 \|D(z-1)\|_F^2.
\end{aligned}
\end{equation}
Now, we invoke Lemma \ref{lem_ezl_ez0_maintext}. By setting the step size $\gamma$ as specified, the algorithm ensures the contraction property $e_z^K \leq \theta e_z^0$, where the contraction factor is defined as $\theta := (1 - \frac{\gamma \rho}{2})^K$. Let $C_z := 4 \|D(z-1)\|_F^2$. Combining this with \eqref{eq:e_z_0_bound}, we obtain the recursion
\begin{equation}
    e_z^K \le \theta e_z^0 \le 2\theta e_{z-1}^K + \theta C_z. \notag
\end{equation}
Denoting $e_z := e_z^K$ and noting the initialization $e_1 = 0$ (since $\z_i(1) = \mathbf{0}$), we solve this recursion for $z \ge 2$. Assuming $2\theta < 1$, summing the geometric series yields
\begin{equation}
\label{eq:recursion_solved}
    e_z \le \theta C_z \sum_{j=0}^{z-2} (2\theta)^j < \frac{\theta C_z}{1 - 2\theta}.
\end{equation}

\paragraph{Step 2: Choosing number of iterations $K$.}
Next, we determine the condition on $K$ to ensure the error coefficient scales with $1/\sqrt{n}$. Specifically, we require
\begin{equation}
    \sqrt{\frac{\theta}{1 - 2\theta}} \le \frac{1}{\sqrt{n}} \iff \theta \le \frac{1}{n+2}. \notag
\end{equation}
Using the inequality $\ln(1-x) \le -x$ for $x \in (0,1)$, it suffices to choose $K$ such that
\begin{equation}
    K \ge \frac{\ln(n+2)}{-\ln(1 - \frac{\gamma\rho}{2})} \ge \frac{2\ln(n+2)}{\gamma\rho}. \notag
\end{equation}
Under this condition, we have $\theta \le 1 / (n+2) < 1/2$, which validates the convergence of the geometric series in \eqref{eq:recursion_solved}.

\paragraph{Step 3: Final bound.}
Combining the results from Step 1 and Step 2, we obtain the final bound on the consensus error
\begin{equation}
    \mathbb{E} \left[\left\|\z_i(z) - \bar{\z}(z)\right\|_2\right] \le \sqrt{e_z} < \sqrt{\frac{\theta}{1 - 2\theta}} \sqrt{C_z} \le \frac{1}{\sqrt{n}} \cdot 2\|D(z-1)\|_F. \notag
\end{equation}
This concludes the proof of Lemma \ref{lem_z_and_zbar}.

\section{Proof of Corollaries~\ref{cor:thm1-c} and \ref{cor:thm1-sc}}
\paragraph{Convex case.}
In the following, we first consider the case with convex losses, by substituting \( \alpha = 0 \), \( h = 3\sqrt{LT}G/R \) and \( L=\left\lceil2\ln(n+2)/\gamma \rho\right\rceil \) into (\ref{thm1}), we have
\begin{equation}
\begin{aligned}
\E[R_{T,i}] &\leq 3nLG \left( \sum_{z=2}^{T/L} \frac{LG}{h} + \sum_{z=1}^{T/L} \frac{2LG}{h} \right) + nhR^2 
\leq \frac{9 nLG^2T}{h} + nhR^2 \\
&\leq 6nGR\sqrt{LT}
= O \left( n \sqrt{\log n} \sqrt{T} \cdot (\gamma\rho)^{-1/2}\right)
= O \left( n \sqrt{\log n} \cdot \rho^{-1}\omega^{-1/2}\cdot \sqrt{T} \right). \notag
\end{aligned}
\end{equation}
When \( \omega = 1 \), we have $(\gamma \rho)^{-1/2} = \sqrt{(8+\rho)/\rho}=O(\rho^{-1/2})$. Thus, the regret bound can be improved to
\begin{equation}
\begin{aligned}
\E[R_{T,i}]
&= O \left( n \sqrt{\log n} \cdot \rho^{-1/2} \cdot \sqrt{T} \right).  \notag
\end{aligned}
\end{equation}
\paragraph{Strongly convex case.}
We continue to consider the case with the strongly convex losses, by substituting $h = \alpha L,  L=\left\lceil 2\ln(n+2)/\gamma \rho\right\rceil$ into (\ref{thm1}), we have 
\begin{equation}
\begin{aligned}
\E[R_{T,i}] &\leq n \alpha L R^2 + \frac{6nLG}{\alpha} \left( (G + \alpha R)\ln (T/L) + 2(G + 2\alpha R)\ln (T/L) \right) \\
&= n \alpha L R^2 + \frac{6nLG\left( 3G + 5\alpha R \right)\ln (T/L)}{\alpha}\\
& = O\left(n \log n \cdot \gamma^{-1} \rho^{-1}  \cdot \log T\right)
= O\left(n \log n \cdot \rho^{-2} \omega^{-1}  \cdot \log T\right).  \notag
\end{aligned}
\end{equation}
When \( \omega = 1 \), we have $(\gamma \rho)^{-1/2} = \sqrt{(8+\rho)/\rho}=O(\rho^{-1/2})$. Thus, the regret bound can be improved to
\begin{equation}
\begin{aligned}
\E[R_{T,i}]
 = O\left(n \log n \cdot \rho^{-1}  \cdot \log T\right). \notag
\end{aligned}
\end{equation}

\section{Proof of Corollaries~\ref{cor:thm2-c} and \ref{cor:thm2-sc}}
\paragraph{Convex case.}
In the following, we first consider the case with convex losses, in which the parameters of our Algorithm~\ref{CD-FTBL} are set to
$\alpha = 0$, $L = \max\{K,\sqrt{T}\}$, $K = \left\lceil 2\ln(n+2)/\gamma \rho\right\rceil$, $h = \sqrt{d L T} M / R$ and $\epsilon = c \, d^{1/2} T^{-1/4}$.
Because of $\alpha=0$ we have
\begin{equation}
\label{cor7_1}
\begin{aligned}
\E[R_{T,i}]
&\leq 6nLG \sum_{z=2}^{T/L} \frac{\sqrt{2L \frac{d^2M^2}{\epsilon^2} +2L^2G^2}}{2h} + 6 n L G \sum_{z=1}^{T/L} \frac{2\frac{\sqrt{L}dM}{\epsilon}+2 L G}{2 h} \\
&\quad + 2 n \sum_{z=1}^{T/L}\frac{\left(2\frac{\sqrt{L}dM}{\epsilon} + 2 L G\right)^2}{2 h} + n h R^2 + 3\epsilon n G T + \frac{\epsilon n G R T}{r}.
\end{aligned}
\end{equation}
Because $\epsilon=c\, d^{1/2} T^{-1/4}$ and $h = \sqrt{d L T} M / R$, we can bound the first term on the R.H.S of \eqref{cor7_1} by
\begin{equation}
\label{cor7_2}
\begin{aligned}
&6nLG \sum_{z=2}^{T/L} \frac{\sqrt{2 L \frac{d^2M^2}{\epsilon^2} + 2 L^2 G^2}}{2 h} \leq 6 n L G \cdot \frac{T}{L} \left( \frac{\sqrt{L}dM}{h\epsilon}+\frac{ L G}{h} \right) 
=O(nT^{3/4})+O(n\sqrt{LT}).
\end{aligned}
\end{equation}
Similarly, the second term can be bounded by
\begin{equation}
\label{cor7_3}
6 n L G \sum_{z=1}^{T/L} \frac{2\frac{\sqrt{L}dM}{\epsilon}+2 L G}{2 h}
= 6nLG \cdot \frac{T}{L} \left(\frac{\sqrt{L}dM}{h \epsilon}+\frac{LG}{h} \right)
= O(nT^{3/4})+O(n \sqrt{LT}).
\end{equation}
And the third term can be bounded by
\begin{equation}
\label{cor7_4}
\begin{aligned}
&2 n \sum_{z=1}^{T/L}\frac{\left(2\frac{\sqrt{L}dM}{\epsilon}+2 L G\right)^2}{2 h}
\leq 4n \cdot \frac{T}{L} \left(\frac{Ld^2M^2}{h \epsilon^2}+\frac{L^2G^2}{h} \right)
= O(n T^{3/4}) + O(n \sqrt{LT}).
\end{aligned}
\end{equation}
Moreover, the remaining terms on the R.H.S of \eqref{cor7_1} have the bound of
\begin{equation}
\label{cor7_5}
\begin{aligned}
& n h R^2 + \left(3+\frac{R}{r}\right)\epsilon n G T
= n M\sqrt{d L T} R + \left(3+\frac{R}{r}\right)c d^{1/2} n G T^{3/4}\\
&= O(n\sqrt{LT})+O(nT^{3/4}). 
\end{aligned}
\end{equation}
Finally, by combining (\ref{cor7_1}), (\ref{cor7_2}), (\ref{cor7_3}), (\ref{cor7_4}) and (\ref{cor7_5}), we can obtain
\begin{equation}
\begin{aligned}
\E[ R_{T,i}] = O(nT^{3/4})+O(n\sqrt{LT})
\leq O(nT^{3/4} + n\sqrt{KT}). \notag
\end{aligned}
\end{equation}
 
\paragraph{Strongly convex case.}
We continue to consider the case with the strongly convex losses, in which the parameters of our Algorithm \ref{CD-FTBL} are set to $\alpha > 0$, $L =\max \{K, T^{2/3}(\ln T)^{-2/3}\}$, $K = \left\lceil 2\ln(n+2)/\gamma \rho\right\rceil$, $\epsilon = c\, d^{2/3} T^{-1/3}(\ln T)^{1/3}$ and $h = \alpha L$.\newline
Because $\alpha > 0$, $\epsilon = c\, d^{2/3} T^{-1/3}(\ln T)^{1/3}$ and $h = \alpha L$, we can bound the first term on the R.H.S of \eqref{cor7_1} by
\begin{equation}
\label{cor8_1}
\begin{aligned}
&6nLG \sum_{z=2}^{T/L} \frac{\sqrt{2L \frac{d^2M^2}{\epsilon^2} +2L^2G^2  +2\alpha^2L^2R^2}}{(z-2)L\alpha + 2h} 
= 6nLG \sum_{z=2}^{T/L} \frac{\sqrt{2L \frac{d^2M^2}{\epsilon^2} + 2L^2G^2 + 2\alpha^2L^2R^2}}{\alpha L z} \\
&\leq 6 n G (1+\ln (T/L)) \left(\frac{2\sqrt{L}dM}{\alpha \epsilon}+ \frac{2L\sqrt{G^2+\alpha^2 R^2}}{\alpha} \right)\\
&= O\left(n \sqrt{L}\, T^{1/3} (\log T)^{2/3}\right)  + O\left(n L \log T\right). 
\end{aligned}
\end{equation}
Then, we can bound the second term by 
\begin{equation}
\label{cor8_2}
\begin{aligned}
& 6 n L G \sum_{z=1}^{T/L} \frac{2 \frac{\sqrt{L} d M}{\epsilon} + 2 L G + 3 \alpha L R}{(z-1) L \alpha + 2 h}
= 6 n L G \sum_{z=1}^{T/L} \frac{2 \frac{\sqrt{L} d M}{\epsilon} + 2 L G + 3 \alpha L R}{\alpha L (z+1)} \\
&\leq 6 n G (1+\ln T/L) \left(\frac{2\sqrt{L}dM}{\alpha\epsilon}+\frac{L(2G+3\alpha R)}{\alpha} \right) \\
&= O\left(n \sqrt{L} T^{1/3} (\log T)^{2/3}\right)  + O\left( n L\log T \right).
\end{aligned}
\end{equation}
Similarly, the third term can be bounded by
\begin{equation}
\label{cor8_3}
\begin{aligned}
& 2 n \sum_{z=1}^{T/L} \frac{\left( 2 \frac{\sqrt{L} d M}{\epsilon} + 2 L G + 3 \alpha L R \right)^2}{z L \alpha + 2 h}
= 2 n \sum_{z=1}^{T/L} \frac{\left( 2 \frac{\sqrt{L} d M}{\epsilon} + 2 L G + 3 \alpha L R \right)^2}{\alpha L (z+2)} \\
&\leq 4n(1+\ln(T/L))\left(\frac{4d^2M^2}{\alpha \epsilon^2}+\frac{L(2G+3\alpha R)^2}{\alpha} \right)
= O\left( n T^{2/3} (\log T)^{1/3} \right) + O\left( n L\log T \right).
\end{aligned}
\end{equation}
The remaining terms on the R.H.S of \eqref{cor7_1} can be bounded by
\begin{equation}
\label{cor8_4}
\begin{aligned}
& n h R^2 +3\epsilon n G T + \frac{\epsilon n G R T}{r}
= n\alpha L R^2 + \left(3+\frac{R}{r}\right) c n G d^{2/3} T^{2/3}(\ln T)^{1/3} \\
&=O(n T^{2/3}(\log T)^{1/3}).
\end{aligned}
\end{equation}

Finally, by combining (\ref{thm2}),  (\ref{cor8_1}), (\ref{cor8_2}), (\ref{cor8_3}) and (\ref{cor8_4}), we can obtain
\begin{equation}
\begin{aligned}
\E[ R_{T,i}]
&= O\left( n T^{2/3} (\log T)^{1/3} \right)+ O\left( n L\log T \right)
\leq O\left( n T^{2/3} (\log T)^{1/3} + n K\log T \right).\notag
\end{aligned}
\end{equation}

\section{Compatibility with both biased and unbiased compressors}
\label{app:unified_compressors}

At first glance, Assumption \ref{assum_compress} might seem designed specifically for biased compressors. However, it is actually a general framework that can easily capture both biased and unbiased compressors. We explain this below.

\textbf{Biased compressors.} Compressors like Top-$k$ sparsification are inherently biased ($\mathbb{E}[\mathcal{Q}(x)] \neq x$). They directly fit Assumption \ref{assum_compress} because they are defined by bounding the compression error. For example, the Top-$k$ operator preserves the $k$ components with the largest magnitudes. Since the discarded $d-k$ components are the smallest, their sum of squares is at most $\frac{d-k}{d}\|x\|_2^2$. This guarantees $\|\mathcal{Q}(x) - x\|_2^2 \leq (1 - \frac{k}{d})\|x\|_2^2$. Therefore, it naturally satisfies Assumption \ref{assum_compress} with $\omega = k/d$.

\textbf{Unbiased compressors.} Unbiased compressors, such as QSGD or Rand-$k$, are usually defined by a bounded variance as follow
\begin{align}
    \mathbb{E}[\mathcal{C}(x)] &= x, \notag \\
    \mathbb{E}[\|\mathcal{C}(x)\|_2^2] &\leq \zeta \|x\|_2^2, \notag
\end{align}
where $\zeta \geq 1$. Directly calculating the compression error gives $\mathbb{E}[\|\mathcal{C}(x) - x\|_2^2] \leq (\zeta - 1)\|x\|_2^2$. Since $\zeta - 1$ can be greater than $1$, this does not immediately match the $(1-\omega)\|x\|_2^2$ format in Assumption \ref{assum_compress}.

However, we can make unbiased compressor fit our framework perfectly by applying a simple scaling factor \citep{Beznosikov2020OnBC}. We define a scaled compressor $\mathcal{Q}'(x) = \frac{1}{\zeta} \mathcal{C}(x)$. By calculating its expected squared error, we have
\begin{align}
    \mathbb{E}[\|\mathcal{Q}'(x) - x\|_2^2] &= \mathbb{E}\left[\left\|\frac{1}{\zeta}\mathcal{C}(x) - x\right\|_2^2\right] \notag \\
    &= \frac{1}{\zeta^2}\mathbb{E}[\|\mathcal{C}(x)\|_2^2] - \frac{2}{\zeta}\|x\|_2^2 + \|x\|_2^2 \notag \\
    &\leq \left(\frac{1}{\zeta} + 1 - \frac{2}{\zeta}\right)\|x\|_2^2 = \left(1 - \frac{1}{\zeta}\right)\|x\|_2^2. \notag
\end{align}

This shows that the scaled compressor $\mathcal{Q}'(x)$ perfectly satisfies Assumption \ref{assum_compress} with a compression ratio of $\omega = 1/\zeta$. Therefore, all the theoretical results in this paper apply to unbiased compressors simply by replacing $\omega$ with $1/\zeta$.

% \newpage
% \input{checklist.tex}
\end{document}